\documentclass[11pt,a4paper]{article}
\usepackage{paper-layout}
\renewcommand{\paperabstractfont}{\fontsize{9.2pt}{11.5pt}\selectfont}
\usepackage{multirow}
\usepackage{wrapfig}
\usepackage{algorithm}
\usepackage{algpseudocode}
\SetSymbolFont{operators}{normal}{OT1}{cmr}{m}{n}
\SetSymbolFont{operators}{bold}{OT1}{cmr}{bx}{n}
\SetMathAlphabet{\mathit}{normal}{OT1}{cmr}{m}{it}
\SetMathAlphabet{\mathit}{bold}{OT1}{cmr}{bx}{it}
\usepackage{fontawesome5}
\usepackage{placeins}

\newtcolorbox{insightbox}[1]{
  enhanced jigsaw,
  unbreakable,
  colback=themecolor!4!white,
  colframe=themecolor!82!black,
  boxrule=0.8pt,
  arc=4pt,
  left=7pt,
  right=7pt,
  top=6pt,
  bottom=4pt,
  before skip=10pt plus 2pt minus 1pt,
  after skip=10pt plus 2pt minus 1pt,
  fontupper=\small,
  title={#1},
  colbacktitle=themecolor,
  coltitle=white,
  fonttitle=\scriptsize\bfseries,
  attach boxed title to top left={xshift=12pt,yshift=-2.6mm},
  boxed title style={
    colback=themecolor,
    colframe=themecolor,
    boxrule=0pt,
    arc=2pt,
    left=4pt,
    right=4pt,
    top=0.5pt,
    bottom=0.5pt
  }
}

\newcommand{\promptchip}[1]{%
  \node[anchor=west, fill=themecolor, text=white, rounded corners=2pt,
        inner xsep=4pt, inner ysep=1.6pt, font=\scriptsize\sffamily\bfseries]
    at ([xshift=10pt]frame.north west) {#1};}
\newcommand{\prompttag}[1]{%
  \node[anchor=south east, inner sep=3pt, text=themecolor!80!black,
        font=\sffamily\fontsize{5.6}{6}\selectfont\itshape]
    at (frame.south east) {#1};}
\newtcolorbox{promptbox}[3][]{
  enhanced,
  breakable,
  frame hidden,
  interior hidden,
  underlay={
    \path[draw=themecolor!75!black, line width=0.6pt, fill=themecolor!3!white,
          rounded corners=3pt]
      (frame.south west) rectangle (frame.north east);
  },
  overlay unbroken={\promptchip{#2}\prompttag{#3}},
  overlay first={\promptchip{#2}},
  overlay last={\prompttag{#3}},
  pad before break=4pt,
  pad after break=4pt,
  left=6pt,
  right=6pt,
  top=9pt,
  bottom=11pt,
  before skip=12pt,
  after skip=8pt,
  fontupper=\ttfamily\fontsize{8.3}{10.7}\selectfont\raggedright,
  parbox=false,
  #1
}
\newlist{plist}{itemize}{1}
\setlist[plist]{label={-},leftmargin=1.1em,labelsep=0.4em,nosep,before=\vspace{1pt},after=\vspace{2pt}}
\newlist{penum}{enumerate}{1}
\setlist[penum]{label={\arabic*.},leftmargin=1.4em,labelsep=0.4em,nosep,before=\vspace{1pt},after=\vspace{2pt}}

\usepackage[capitalize,nameinlink,noabbrev]{cleveref}
\crefformat{section}{\mbox{Section}~#2#1#3}
\Crefformat{section}{\mbox{Section}~#2#1#3}
\crefformat{subsection}{\mbox{Section}~#2#1#3}
\Crefformat{subsection}{\mbox{Section}~#2#1#3}
\crefname{appendix}{Appendix}{Appendices}
\Crefname{appendix}{Appendix}{Appendices}
\crefname{subappendix}{Appendix}{Appendices}
\Crefname{subappendix}{Appendix}{Appendices}
\crefname{subsubappendix}{Appendix}{Appendices}
\Crefname{subsubappendix}{Appendix}{Appendices}

\newcommand{\method}{\mbox{\textsc{S-Mind}}}
\newcommand{\tokens}{spatial tokens}
\newcommand{\contribmark}{\tikz[baseline=-0.3em]\draw[themecolor, line width=0.08em, line join=round, rounded corners=0.12em] (0,-0.24em) -- (0,0.24em) -- (0.4157em,0) -- cycle;}
\DeclareRobustCommand{\badge}[1]{\tikz[baseline=(b.base)]\node[circle, fill=themecolor, text=white, inner sep=0pt, minimum size=1.15em, font=\footnotesize\bfseries](b){#1};}
\providecommand{\tbd}{\textcolor{red!70}{\scriptsize TBD}}

\newcommand{\projectlink}{\url{https://kanaboon1.github.io/S-Mind/}}
\newcommand{\ghlink}{\url{https://github.com/KANABOON1/S-Mind}}
\newcommand{\hflink}{\url{https://huggingface.co/KANABOON1/S-Mind}}
\newcommand{\projectlinks}{\par\vspace{0.35em}\begingroup
  \raggedright\fontsize{8pt}{9.6pt}\selectfont
  \renewcommand{\arraystretch}{1.0}\setlength{\tabcolsep}{0pt}
  \begin{tabular}{@{}l@{\hspace{0.55em}}>{\bfseries\color{themecolor}}l@{\hspace{1.1em}}l@{}}
  \faGlobe & Project Page & \projectlink \\
  \faGithub & GitHub & \ghlink \\
  \faLink & Hugging Face & \hflink
  \end{tabular}\par\endgroup}

\newcommand{\institutionlogoheight}{0.94cm}
\newcommand{\institutionlogogap}{0.3cm}
\newcommand{\PlaceFirstPageHeader}{\AddToShipoutPictureFG*{%
  \put(\LenToUnit{\dimexpr1in+\oddsidemargin\relax},\LenToUnit{\dimexpr\paperheight-0.813333cm\relax}){%
    \makebox(0,0)[lt]{\makebox[\textwidth][l]{%
      \includegraphics[height=\institutionlogoheight]{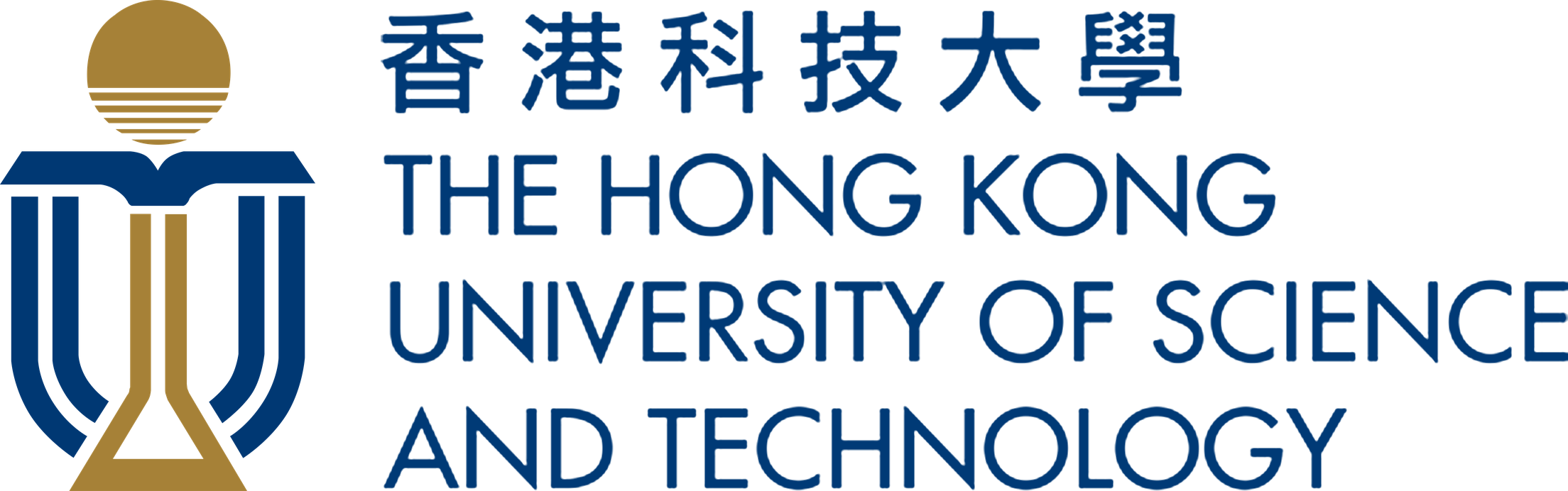}%
      \hspace{\institutionlogogap}%
      \includegraphics[height=\institutionlogoheight]{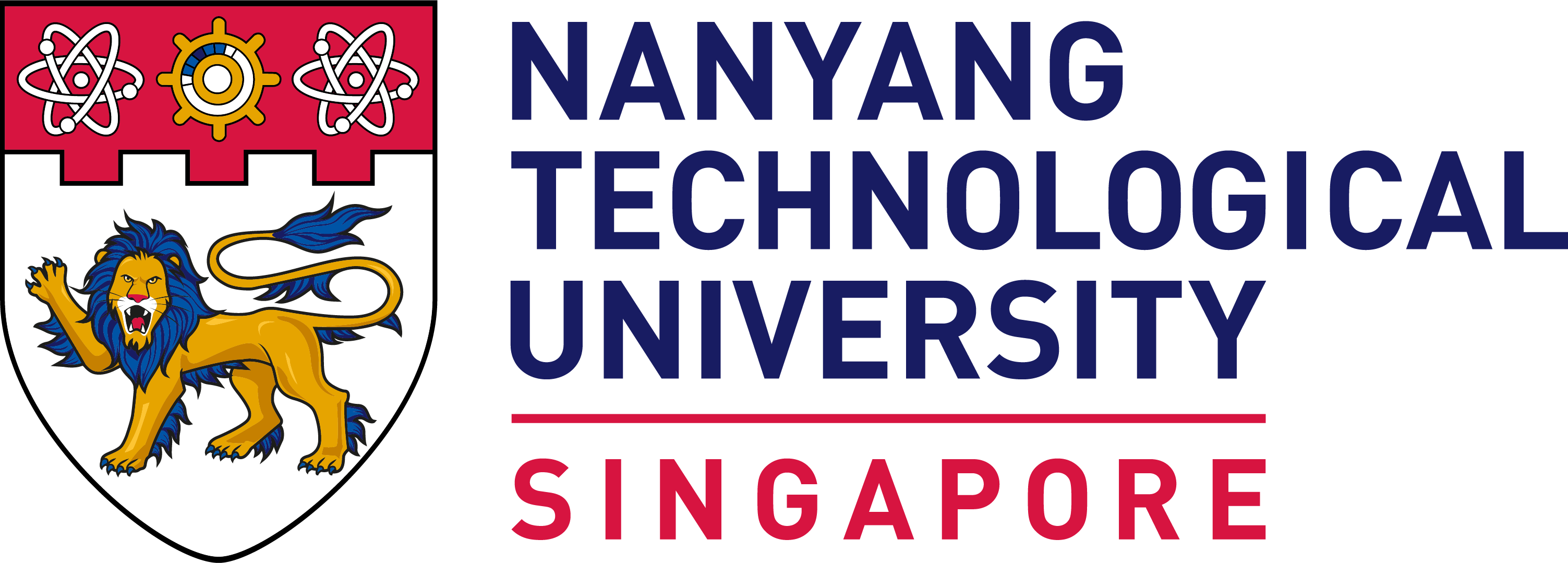}%
      \hspace{\institutionlogogap}%
      \includegraphics[height=\institutionlogoheight]{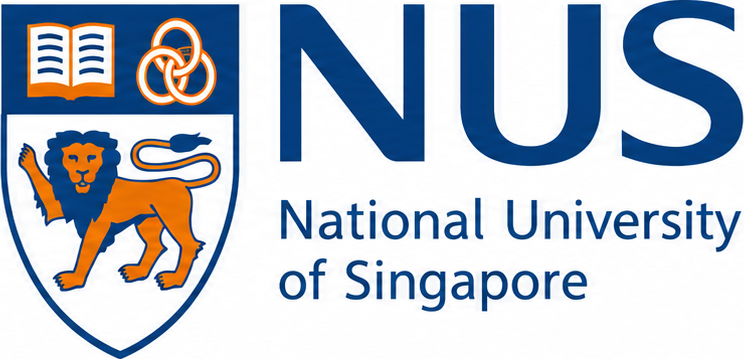}\hfill}}}%
  \put(\LenToUnit{\dimexpr1in+\oddsidemargin\relax},\LenToUnit{\dimexpr\paperheight-1.90cm\relax}){%
    \makebox(0,0)[lt]{\rule{\textwidth}{0.4pt}}}%
}}
\posttitle{\par\endgroup\vspace{0.1em}}
\postauthor{\par\endgroup\vspace{0.05em}}
\title{\vspace*{-0.5cm}%
  \includegraphics[height=1.45cm]{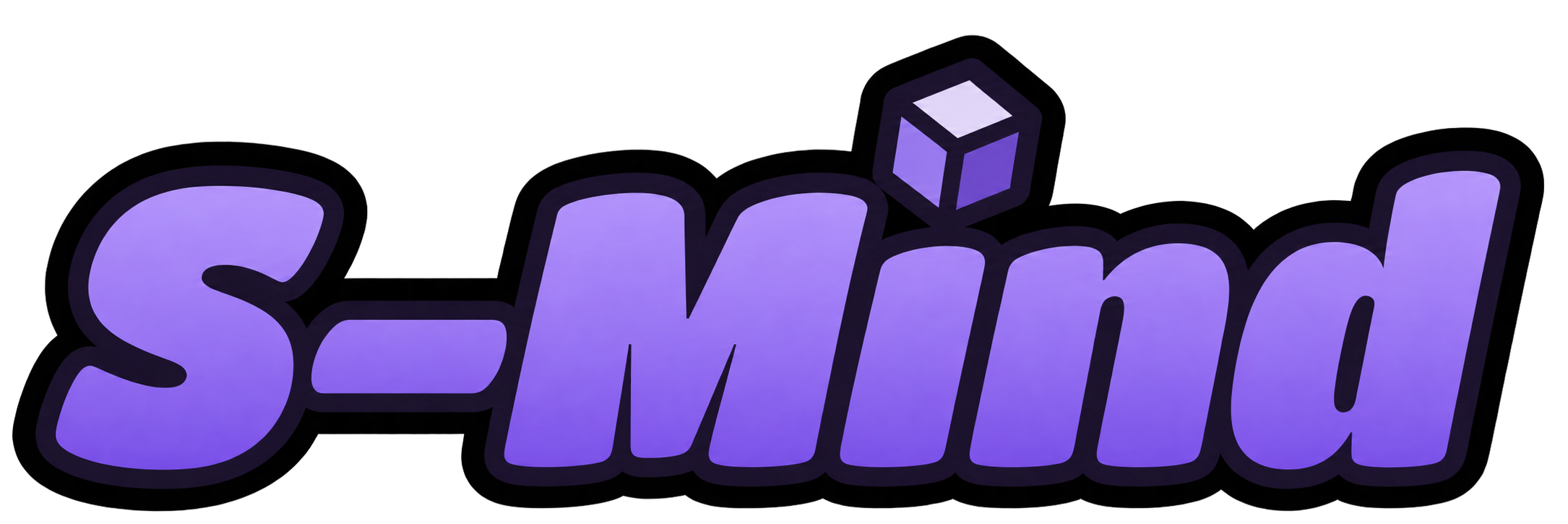}\\[0.1em]
  {\fontsize{15pt}{18pt}\selectfont\textbf{Internalizing Multi-View Geometry for Spatial Intelligence}}}
\author{{\fontsize{10pt}{12.2pt}\selectfont
\textbf{Muxin Fu}$^{1,2,*}$,\; \textbf{Peike Yu}$^{2,*}$,\; \textbf{Guibin Zhang}$^{3,\ddagger}$,\; \textbf{Wenrui Zhou}$^{1}$,\; \textbf{Runze Liu}$^{1}$,\\
\textbf{Lang Feng}$^{2}$,\; \textbf{Fuchao Yang}$^{2}$,\; \textbf{Feng Chen}$^{2}$,\; \textbf{Yixuan Li}$^{2}$,\; \textbf{Mengao Zhang}$^{2}$,\; \textbf{Shiying Zhang}$^{4}$,\\
\textbf{Wen Shen}$^{4}$,\; \textbf{Ling Pan}$^{1}$,\; \textbf{Shuicheng Yan}$^{3}$,\; \textbf{Bo An}$^{2,\dagger}$}\\[0.2em]
{\fontsize{9.2pt}{11.4pt}\selectfont $^1$The Hong Kong University of Science and Technology\quad $^2$Nanyang Technological University}\\[-0.1em]
{\fontsize{9.2pt}{11.4pt}\selectfont $^3$National University of Singapore\quad $^4$Tongji University}\\[0.05em]
{\fontsize{8.2pt}{10pt}\selectfont $^*$Equal contribution\quad $^\ddagger$Project lead\quad $^\dagger$Corresponding author}}
\date{}
\hypersetup{pdftitle={S-Mind: Internalizing Multi-View Geometry for Spatial Intelligence},pdfauthor={},pdfsubject={},pdfkeywords={}}

\begin{document}
\PlaceFirstPageHeader
\maketitle
\newskip\bodyfootins \bodyfootins=\skip\footins
\setlength{\skip\footins}{5pt}
\begingroup\renewcommand{\thefootnote}{}\footnotetext{Work done while at NTU.}\endgroup
\vspace{-2.0em}

\begin{abstract}
Spatial intelligence, the ability to reason about 3D space from visual observations, becomes essential as multimodal large language models (MLLMs) move into the physical world.
Beyond perceiving each view, it rests on two cornerstones of human cognition: fusing multiple views into one coherent 3D scene and mentally transforming that scene to new viewpoints as reasoning unfolds.
Yet our spatial probes reveal that MLLMs largely lack both capabilities, even after spatial fine-tuning.
We therefore present \method{}, which internalizes multi-view geometry within the MLLM through a compact set of \tokens{}.
It supervises these tokens to hold the fused 3D scene and re-emits them to re-express this scene from every viewpoint the reasoning adopts.
Across four spatial benchmarks, \method{} improves its Qwen3.5 backbones by up to 22.26 points and surpasses frontier general models such as \mbox{GPT-6-Astra} ($+$0.45), \mbox{Kimi-K3.1-Pro} ($+$1.97), and \mbox{GLM-5.3-Flash} ($+$7.80) in average accuracy.
Moreover, the same probes confirm that it acquires both capabilities, reaching fusion and transformation accuracies of 65.11\% and 82.17\%, against at most 16.02\% and 37.28\% for the baselines.
At inference, \method{} also runs about 100$\times$ faster than a \mbox{tool-based} agent on the same backbone.
Instilled through these \tokens{} during training, spatial capability becomes intrinsic to the backbone and persists even when the tokens are withheld, offering a promising path toward spatial intelligence in MLLMs.
\projectlinks
\end{abstract}

\noindent\begin{minipage}{\linewidth}
\centering
\vspace{0.15em}
\includegraphics[width=0.483003\linewidth]{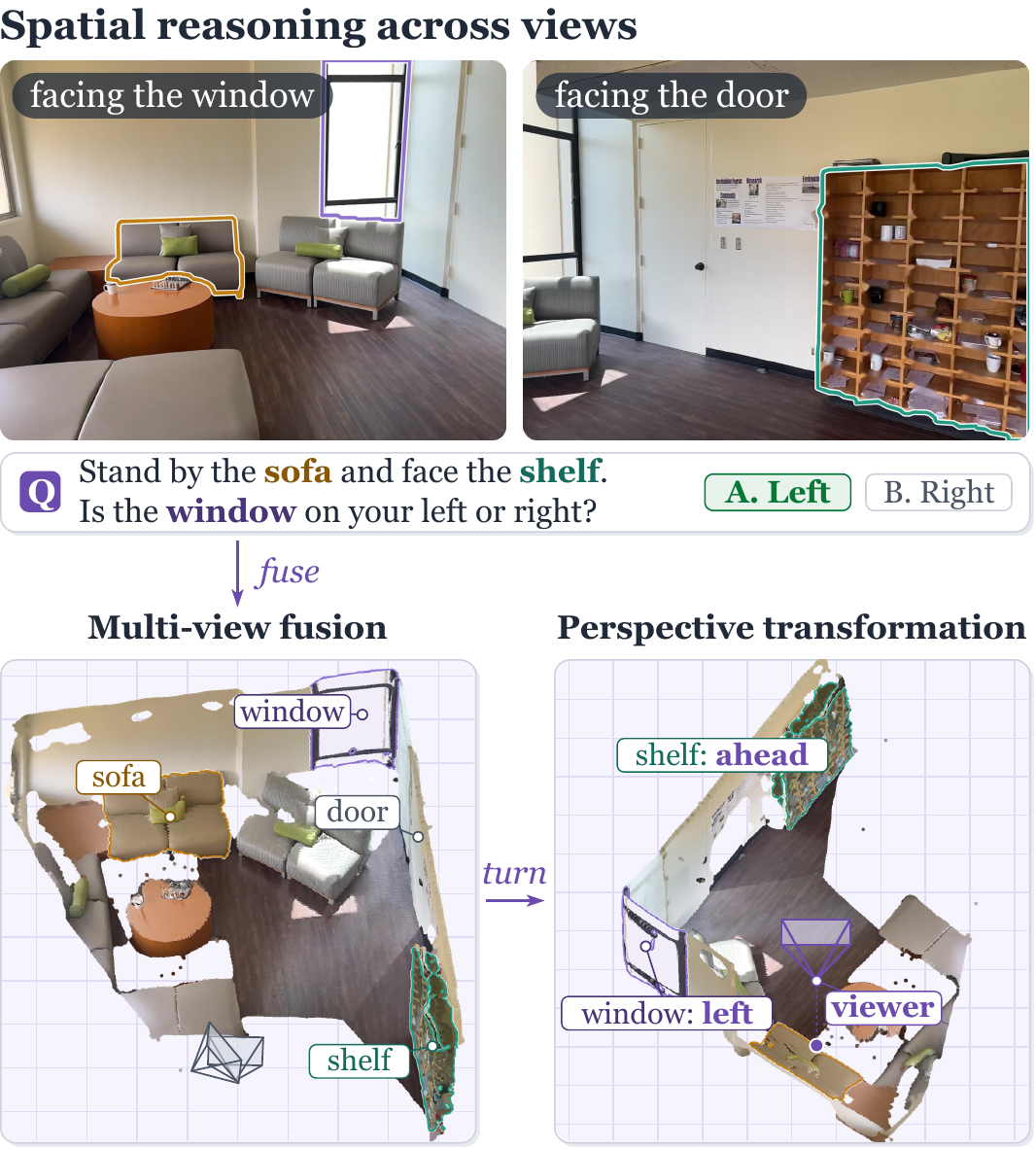}%
\includegraphics[width=0.516997\linewidth]{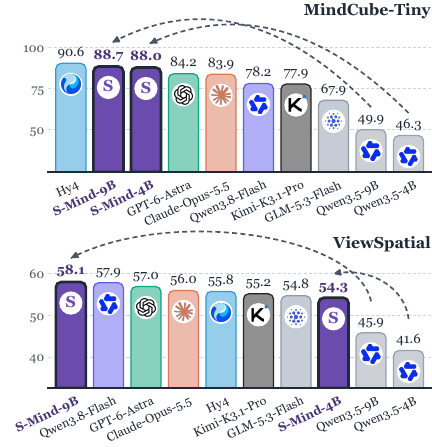}
\captionof{figure}{\textbf{Spatial reasoning across views, and accuracy on representative benchmarks.} Left: answering the question requires fusing the two snapshots into one scene and turning that scene to an observer standing by the sofa. Right: accuracy (\%) on MindCube-Tiny and ViewSpatial (\cref{tab:main}). By internalizing spatial capability through spatial tokens, \method{} substantially improves its Qwen3.5 backbones at both scales and rivals frontier general models such as GPT-6-Astra and Claude-Opus-5.5.}
\label{fig:teaser}
\end{minipage}
\clearpage
\skip\footins=\bodyfootins
\section{Introduction}
\label{sec:intro}

\afterpage{\begin{figure}[t]
\centering
\includegraphics[width=\linewidth]{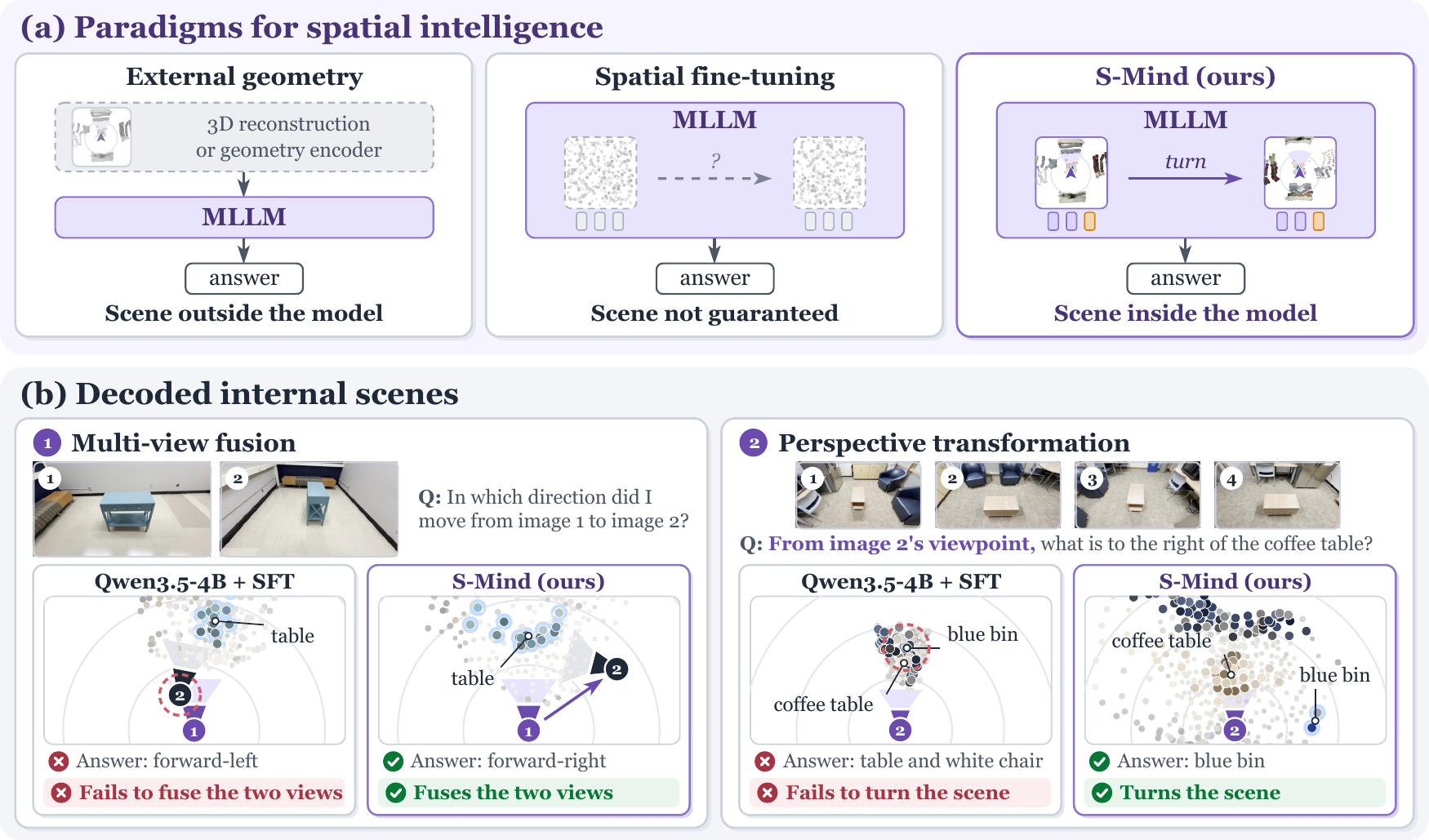}
\caption{\textbf{Internalizing spatial capability in MLLMs.} The top panel (a) contrasts the two existing paradigms for spatial intelligence with \method{}: external-geometry methods build the scene outside the model via 3D reconstruction; spatial fine-tuning improves the answers without forming a coherent internal scene; and \method{} explicitly fuses the views into an internal scene and rotates it as the reasoning proceeds. The bottom panel (b) shows top-down views of the internal scenes decoded on two questions: the spatially fine-tuned baseline (Qwen3.5-4B\,+\,SFT, read through the spatial probes) neither fuses the two views nor turns the scene to the stated viewpoint, whereas \method{} does both. Dashed circles mark the baseline's errors.}
\label{fig:intro}
\end{figure}}

Two snapshots of a room, one facing the window and one facing the door, are enough to walk through it in the mind: stand by the sofa, face the shelf, and the window is on the left (\cref{fig:teaser}).
Behind this lie two operations: assembling the snapshots into one model of the room, and mentally rotating that model to the new standpoint~\citep{shepard1971mental}.
As multimodal large language models (MLLMs)~\citep{step3vl,phi4reasoningvision,qwen35} advance rapidly in visual understanding and reasoning, real-world tasks such as embodied AI~\citep{pragmaticvla,threedthinkvla}, autonomous driving~\citep{lastvla,nureasoning}, and extended reality~\citep{egocopilot,openglass} increasingly hinge on such spatial intelligence.

Despite its importance, spatial intelligence remains a major bottleneck for MLLMs~\citep{mindcube,easi,theoryofspace}, requiring the model to mentally construct a coherent 3D scene rather than merely recognize 2D visual patterns. As \cref{fig:intro}a illustrates, existing approaches generally fall into two paradigms:
\textbf{(i)~External-geometry methods} construct the scene outside the model, via 3D reconstruction and rendering~\citep{think3d,mindjourney,spatialclaw} or an attached geometry encoder~\citep{spatialmllm,vlm3r,spacemind}. While providing explicit geometry, they fail to internalize spatial capability and incur high inference latency~\citep{spatialopsd,gpd}.
\textbf{(ii)~Spatial fine-tuning methods} aim to train the MLLM to construct the scene internally, with supervision progressing from the output text alone~\citep{spatialladder,vst,prosr} to latent states aligned with 3D geometry~\citep{threedthinker,threedthinkvla,geovr,hatch,geoanchor}. This paradigm has drawn growing attention for its promise of internalizing spatial capability within the model itself.

However, the evidence for this promise rests largely on end-task accuracy. We therefore introduce spatial probes, extending recent analyses of the spatial representations inside MLLMs~\citep{sspace}, to decode the scene a model holds internally. Our probes reveal that, despite substantial gains in accuracy, spatially fine-tuned models retain two persistent deficiencies in spatial capability, as illustrated in \cref{fig:intro}b.
\textbf{(i)~Limited multi-view fusion:} these models encode each view well but in isolation, and the views are barely fused into a coherent scene.
\textbf{(ii)~Incomplete perspective transformation:} as in base models~\citep{sspace}, their internal representation still largely fails to follow when the reasoning shifts to a new viewpoint.
This mismatch raises a natural question:

\nopagebreak
\begin{insightbox}{Research Question}
\centering
\emph{How can an MLLM truly internalize spatial capability, assembling its views\\ into one scene and rotating that scene to a new viewpoint, as humans do?}
\end{insightbox}

To address this question, we introduce \method{}, a framework that endows an MLLM with an internal scene representation, built from its input views and transformed as its reasoning unfolds.
\textbf{(I) Multi-view fusion.} \method{} introduces a compact set of \tokens{} whose hidden states consolidate the views into a unified 3D scene, encoding both the geometry each image observes and the pose of each camera in a shared coordinate frame.
\textbf{(II) Perspective transformation.} With the views fused into one scene, adopting a new viewpoint reduces to re-expressing that scene in a different frame of reference. Whenever the reasoning shifts to a new viewpoint, the model therefore re-emits the \tokens{}, which now encode the fused scene as observed from that viewpoint, and conditions its subsequent reasoning on them. 
By training the model to construct and manipulate an entire 3D scene, \method{} internalizes genuine spatial capability rather than superficial heuristics.

Extensive evaluations across four spatial reasoning benchmarks and two model scales demonstrate the effectiveness of \method{}.
\textbf{(I)~Strong performance:} \method{} reaches an average accuracy of up to 74.03\%, surpassing the strongest fine-tuned baseline by 6.76 points and the best released spatial reasoning model by 19.11 points, and rivaling frontier general models such as GPT-6-Astra ($+$0.45) and Kimi-K3.1-Pro ($+$1.97).
\textbf{(II)~Internalized spatial capability:} At the representation level, \method{} overcomes both deficiencies. Multi-view fusion raises the recovery rate of cross-view 3D points from at most 16.02\% for the baselines to 65.11\%, and perspective transformation lifts the accuracy of locating points relative to a new viewpoint from at most 37.28\% to 82.17\%.
\textbf{(III)~Efficient and internalized reasoning:} Requiring no tool calls, \method{} runs roughly 100$\times$ faster than an external-geometry agent on the same backbone. Moreover, spatial capability is instilled into the backbone through the \tokens{} during training: even with all of them withheld at inference, \method{} loses only 2.37 points and still leads the strongest baseline on its backbone by 5.28 points.

\paragraph{Contributions.} Our contributions are summarized as follows:
\begin{itemize}[label=\contribmark, leftmargin=11pt, labelwidth=10pt, labelsep=5pt, topsep=-2.2pt]
\item \textbf{Representation-Level Diagnosis.} Through our spatial probes, we identify two deficiencies in the spatial capability of current MLLMs, limited multi-view fusion and incomplete perspective transformation, which persist even after spatial fine-tuning substantially improves their accuracy.
\item \textbf{Practical Solution.} We propose \method{}, which addresses both deficiencies through \tokens{} that consolidate the views into a unified 3D scene and are re-emitted at each viewpoint shift to re-express that scene in the new frame.
\item \textbf{Experimental Validation.} We validate \method{} across four benchmarks and two model scales, showing that it internalizes spatial capability, achieves state-of-the-art accuracy among spatial reasoning models, rivals frontier general models, and attains high inference efficiency.
\end{itemize}

\section{Related Work}
\label{sec:related}

\paragraph{External-Geometry Methods.}
External-geometry methods compute the scene outside the MLLM and supply it as input.
Tool-based designs invoke specialist modules at inference: some build and operate on an explicit scene representation, such as a 3D reconstruction, an object-level abstraction, or a vector sketch~\citep{think3d,spatialclaw,mentalimagery,mentalthink}; others synthesize views from new standpoints using generative models~\citep{mindjourney,camcue} or warp visual tokens toward them~\citep{tokenwarping}.
Encoder-based designs instead integrate geometry into the architecture, pairing the visual encoder with a feed-forward reconstruction model such as VGGT~\citep{vggt} and fusing their features~\citep{spatialmllm,vlm3r,spacemind}; subsequent work refines where and how these features are injected~\citep{guide,spatialstack,geothinker}.
However, both designs offload scene construction to external modules, failing to internalize genuine spatial capability while incurring extra inference overhead~\citep{spatialopsd,gpd}.
\method{} keeps the geometry but moves it from inference into training, where it teaches the model itself to fuse the views and transform the scene.

\paragraph{Spatial Fine-Tuning Methods.}
Spatial fine-tuning methods train the MLLM to form the scene internally, with no geometry at inference.
Most methods shape only what the model says: they scale up spatial instruction data~\citep{spatialvlm,spar,cambrians,vst}, train with progressive curricula, process-level rewards, or interaction trajectories~\citep{spatialladder,prosr,spatialinteractor}, elicit textual cognitive maps or cross-view alignments before the answer~\citep{mindcube,trace,viewfusion}, or distill tool-using agents and privileged geometry~\citep{spatialopsd,gpd}.
A growing body of work, closer in spirit to latent reasoning~\citep{coconut,mirage,hive}, shapes what the model holds instead: it aligns latent states with 3D foundation models or across views~\citep{threedthinker,threedthinkvla,geovr,hatch}, or adds dedicated tokens for positions, directions, and an observer's orientation~\citep{geoanchor,perspectivetokens}.
Yet these targets largely encode the geometry of the input views, leaving viewpoint changes unsupervised.
\method{} instead supervises its \tokens{} to fuse the views into one frame and to re-express the scene at each viewpoint the reasoning adopts.

\section{Does Fine-Tuning Really Internalize Spatial Capability in MLLMs?}
\label{sec:diagnosis}

Higher answer accuracy after fine-tuning does not by itself establish that a model has internalized spatial capability.
We therefore propose \emph{spatial probes}, which extend recent analyses of spatial representations in MLLMs~\citep{sspace}, to examine what a model encodes about a scene before and after spatial fine-tuning.
They target two capabilities fundamental to spatial intelligence~\citep{easi,mindcube}: a \emph{geometry probe} tests whether multiple views are fused into a unified internal scene, and a \emph{viewpoint probe} tests whether that scene is transformed to the viewpoints adopted during reasoning.
More details are in \cref{app:diagnosis}.

\begin{figure}[t]
\centering
\includegraphics[width=\linewidth]{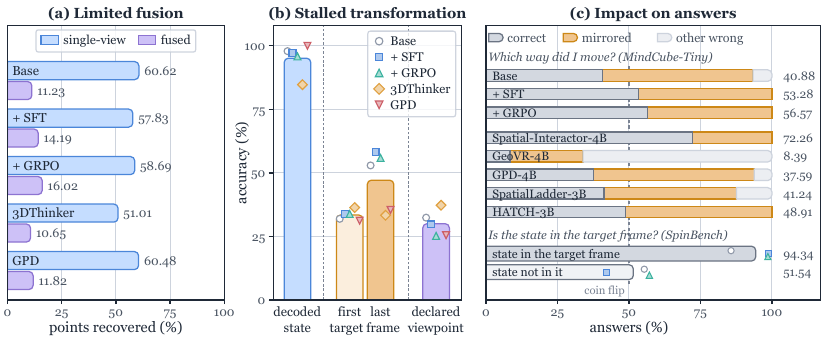}
\caption{\textbf{Diagnosis of multi-view fusion and perspective transformation in MLLMs.} (a)~The geometry probe recovers the 3D points of the first image (\emph{single-view}) and of the other images in its frame (\emph{fused}). (b)~The viewpoint probe reads the relation in the \emph{decoded state} and directly in the \emph{target frame} at the \emph{first} and \emph{last} mention of the objects, and the scene at the \emph{declared viewpoint}. (c)~The same deficiencies in the answers; \emph{mirrored} answers place the second camera on the opposite side. Bars are means over models; markers are individual models.}
\label{fig:diagnosis}
\end{figure}

\paragraph{Finding 1: Limited multi-view fusion.}
To isolate what the model itself encodes, the geometry probe appends learnable tokens to the frozen model and decodes from their hidden states the 3D point of every visual token, expressed in the first image's frame.
As \cref{fig:diagnosis}a shows, on Qwen3.5-4B (\emph{Base}) it recovers 60.62\% of the first image's own points (\emph{single-view}) but only 11.23\% of the other images' points (\emph{fused}), which can be placed only by relating the views.
Fine-tuning barely closes this gap: the \emph{+\,SFT} and \emph{+\,GRPO}~\citep{grpo} variants lift the fused points only to 14.19\% and 16.02\%, and the released 3DThinker and GPD reach 10.65\% and 11.82\%.
Crucially, the probe's verdict is borne out by behavior.
Questions on how the camera moved between two views hinge on exactly this fusion, yet as the top of \cref{fig:diagnosis}c shows, SFT and GRPO are right on only 53.28\% and 56.57\% of such questions in MindCube-Tiny, every error placing the second camera on the mirrored side.
Overall, multi-view fusion remains a persistent challenge for MLLMs, one that current spatial fine-tuning does little to overcome.

\paragraph{Finding 2: Incomplete perspective transformation.}
\citet{sspace} find that base MLLMs fall short in transforming spatial representations across viewpoints.
We further use the viewpoint probe to test whether spatial fine-tuning closes this gap: the probe reads the relation between two objects directly in the queried \emph{target frame} and after an external rotation into it, separating transformation from perception.
As \cref{fig:diagnosis}b shows, the model perceives the scene but largely fails to transform it.
On SpinBench's single-image relation questions~\citep{spinbench}, the \emph{decoded state} gives the correct relation for 95.14\% of them once rotated to the queried viewpoint, yet for only 33.44\% when read directly in the target frame at the first mention of the objects, and 47.15\% at the last.
The same holds when the reasoning itself declares the viewpoint: the scene read in the frame of each \emph{declared viewpoint} is recovered with only 30.02\% accuracy on average, barely above the chance level of 25\%.
The failure is reflected in the answers: as the bottom of \cref{fig:diagnosis}c shows, the Qwen3.5-4B models reach 85.71--98.73\% accuracy on questions whose final state lies in the target frame, but only 42.11--57.14\%, near chance, on the rest.
Spatial fine-tuning thus does not close the gap: the scene stays largely anchored to the camera instead of following the reasoning to new viewpoints.

\section{\method{}: Fusing Views, Transforming Perspectives}
\label{sec:method}

\begin{figure*}[t]
\centering
\includegraphics[width=\textwidth]{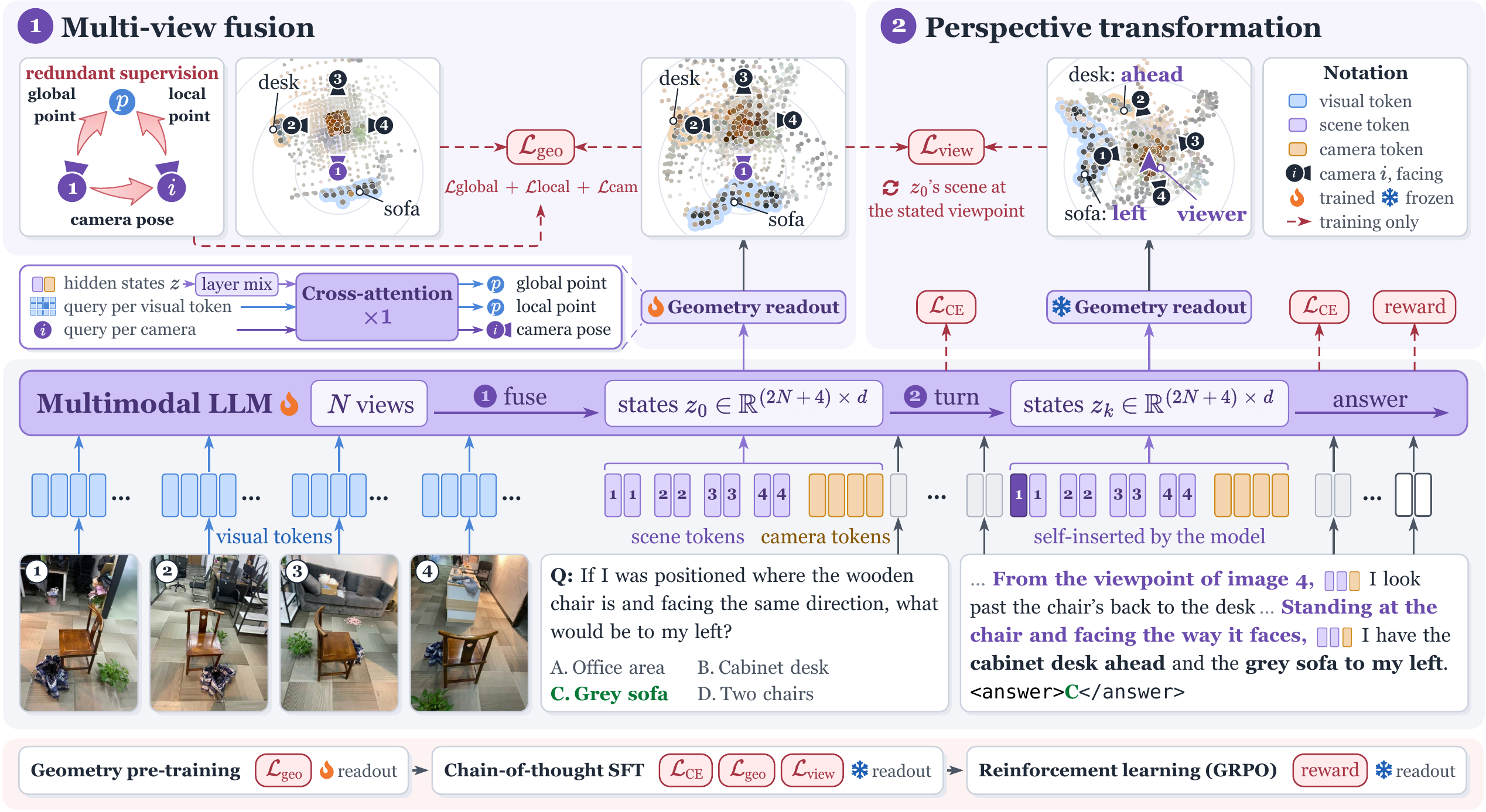}
\caption{\textbf{Overview of \method{}.} \badge{1}~\emph{Multi-view fusion}: the \tokens{} appended after the $N$ views hold the fused scene $\mathbf{z}_0$, from which a geometry readout decodes, under $\mathcal{L}_{\mathrm{geo}}$, the 3D point of every visual token and the pose of every camera. \badge{2}~\emph{Perspective transformation}: whenever the reasoning adopts a viewpoint, the model re-emits the \tokens{}, and $\mathcal{L}_{\mathrm{view}}$ trains their states $\mathbf{z}_k$ to hold the scene of $\mathbf{z}_0$ as seen from that viewpoint. The bottom row shows the three training stages; neither the readout nor the losses are used at inference.}
\label{fig:method}
\end{figure*}

In this section, we detail the two components of \method{}, each together with its training: multi-view fusion ($\triangleright$~Section~\ref{sec:method-fusion}), which consolidates the input views into one scene held by a compact set of \tokens{}, and perspective transformation ($\triangleright$~Section~\ref{sec:method-turning}), which re-expresses that scene in the frame of every viewpoint the reasoning adopts.
The overall pipeline is illustrated in \cref{fig:method}.

\subsection{Multi-View Fusion}
\label{sec:method-fusion}
Multi-view fusion integrates all views into a single scene, which hardly emerges spontaneously (\cref{sec:diagnosis}).
\method{} therefore reserves dedicated positions in the model's own sequence for the fused scene and supervises their hidden states directly: since these states are computed by the model itself, the supervision compels it to fuse the views and internalize the observed spatial structure.

\paragraph{Spatial tokens.}
Given $N$ views $\{I_i\}_{i=1}^{N}$, we append $K_sN{+}K_c$ \tokens{} after the images, factorizing the scene into what each view sees and where each camera stands: $K_s$ \emph{scene tokens} per view and $K_c$ \emph{camera tokens} shared by all views.
Their hidden states serve as the scene representation,
\begin{equation}
\mathbf{z}_0=\big[\mathbf{z}^{(1)};\,\dots;\,\mathbf{z}^{(N)};\,\mathbf{z}^{\mathrm{cam}}\big]\in\mathbb{R}^{(K_sN+K_c)\times d},
\end{equation}
where $\mathbf{z}^{(i)}$ belongs to view $i$ and $\mathbf{z}^{\mathrm{cam}}$ to the cameras.
Residing in the sequence itself, these states are visible to all subsequent reasoning through ordinary attention and require no architectural change.

\paragraph{Token-level geometry.}
We ground the scene at the granularity the MLLM perceives: every visual token is assigned the 3D point it observes.
Whereas M-RoPE~\citep{qwen2vl} encodes where each visual token lies in time and in the image plane, token-level geometry endows it with 3D coordinates.
A single cross-attention readout $\mathcal{R}_\phi$ mixes the hidden states of $\mathbf{z}_0$ across layers with learned weights and decodes geometry and camera poses from them:
\begin{equation}
\begin{aligned}
\big(\hat{\mathbf{p}}^{\mathrm{g}}_{i,u},\,\hat{\mathbf{p}}^{\mathrm{l}}_{i,u}\big)&=\mathcal{R}_\phi\big(\mathbf{q}_u;\,[\mathbf{z}^{(i)};\mathbf{z}^{\mathrm{cam}}]\big),\\
\hat{\mathbf{T}}_i&=\mathcal{R}_\phi\big(\mathbf{q}^{\mathrm{cam}}_i;\,\mathbf{z}_0\big),
\end{aligned}
\end{equation}
where the query $\mathbf{q}_u$ encodes only the grid position $u$ of a visual token in view $i$, and $\mathbf{q}^{\mathrm{cam}}_i$ is derived from $\mathbf{z}^{(i)}$.
$\hat{\mathbf{p}}^{\mathrm{g}}_{i,u}$ and $\hat{\mathbf{p}}^{\mathrm{l}}_{i,u}$ denote the observed point in the frame of the first view (\emph{global}) and of its own camera (\emph{local}), and $\hat{\mathbf{T}}_i$ the pose of camera $i$.
Since the queries carry no image content, the readout can decode only the geometry that $\mathbf{z}_0$ already holds. More details are provided in \cref{app:readout}.

\paragraph{Geometry objective.}
To instill the fused scene in $\mathbf{z}_0$, we pre-train on posed image sets and regress the decoded points and poses onto their targets under $\mathcal{L}_{\mathrm{geo}}=\mathcal{L}_{\mathrm{global}}+\mathcal{L}_{\mathrm{local}}+\mathcal{L}_{\mathrm{cam}}$, with
\begin{equation}
\begin{aligned}
&\mathcal{L}_{\mathrm{global}}=\frac{1}{|\Omega|}\sum_{(i,u)\in\Omega}\big[c^{\mathrm{g}}_{i,u}\,\rho(\hat{\mathbf{p}}^{\mathrm{g}}_{i,u}-\mathbf{p}^{\mathrm{g}}_{i,u})-\alpha\log c^{\mathrm{g}}_{i,u}\big],\\
&\mathcal{L}_{\mathrm{cam}}=\frac{1}{N-1}\sum_{i=2}^{N}\rho(\hat{\mathbf{T}}_i-\mathbf{T}_i),
\end{aligned}
\end{equation}
where $\Omega$ is the set of visual tokens with a valid target, $\rho$ is the Huber loss, $c^{\mathrm{g}}_{i,u}\geq 1$ is a per-point confidence predicted by the readout, and $\alpha$ weights the confidence regularizer~\citep{vggt}.
Through $\mathcal{L}_{\mathrm{local}}$, the counterpart of $\mathcal{L}_{\mathrm{global}}$ on the local points, we deliberately make the supervision redundant, as global points, local points, and camera poses each follow from the other two, so that the points and poses decoded from $\mathbf{z}_0$ are driven toward one coherent scene (\cref{app:design}).

\subsection{Perspective Transformation}
\label{sec:method-turning}
A fused scene is tied to the frame of the first view, whereas spatial reasoning routinely leaves that frame for another viewpoint.
\method{} therefore carries the scene along with the reasoning, re-expressing it in the model's own sequence at each new viewpoint.

\paragraph{Re-emitting spatial tokens.}
A reasoning chain traverses viewpoints $v_1,\dots,v_M$.
Upon declaring $v_k$, the model emits the \tokens{} anew, whose hidden states form $\mathbf{z}_k$, the counterpart of $\mathbf{z}_0$ at $v_k$.
Emitted right after the declaration, these tokens attend to the fused scene in $\mathbf{z}_0$ and the viewpoint just declared, so the model steers $\mathbf{z}_0$ with its own words, turning the scene by merely naming a viewpoint.

\paragraph{Viewpoint objective.}
We supervise $\mathbf{z}_k$ to hold the very scene of $\mathbf{z}_0$, re-expressed in the frame of $v_k$.
Concretely, the readout $\mathcal{R}_\phi$, now frozen, decodes from $\mathbf{z}_k$ a point $\hat{\mathbf{p}}^{k}_{i,u}$ for every visual token, which is regressed onto its counterpart in $\mathbf{z}_0$ under the rigid transformation to $v_k$:
\begin{equation}
\mathcal{L}_{\mathrm{view}}=\frac{1}{M|\Omega|}\sum_{k=1}^{M}\sum_{(i,u)\in\Omega}\!\!\rho\big(\hat{\mathbf{p}}^{k}_{i,u}-\mathrm{sg}\big[\hat{\mathbf{T}}_{v_k}^{-1}\hat{\mathbf{p}}^{\mathrm{g}}_{i,u}\big]\big),
\end{equation}
where $\hat{\mathbf{p}}^{\mathrm{g}}_{i,u}$ is the global point decoded from $\mathbf{z}_0$, $\hat{\mathbf{T}}_{v_k}$ is the pose of $v_k$ in the first view's frame, decoded from $\mathbf{z}_0$ when $v_k$ is an input camera and otherwise built from the cameras and objects named in its declaration (\cref{app:data}), and $\mathrm{sg}[\cdot]$ denotes the stop-gradient operator.
With $\mathbf{z}_0$ supplying the scene, $\mathcal{L}_{\mathrm{view}}$ supervises only how the scene transforms, separating manipulation from perception.

\paragraph{Training with reasoning.}
Perspective transformation is learned in concert with reasoning.
Starting from the geometry pre-trained model, SFT on reasoning chains minimizes
\begin{equation}
\mathcal{L}_{\mathrm{SFT}}=\mathcal{L}_{\mathrm{CE}}+\lambda\,\big(\mathcal{L}_{\mathrm{geo}}+\mathcal{L}_{\mathrm{view}}\big),
\end{equation}
where the cross-entropy $\mathcal{L}_{\mathrm{CE}}$ over the chain also teaches the model when to emit the \tokens{}, $\mathcal{L}_{\mathrm{view}}$ trains the re-emitted tokens to hold the scene as seen from each declared viewpoint, and $\mathcal{L}_{\mathrm{geo}}$ anchors $\mathbf{z}_0$ to the fused scene, with $\lambda$ weighting the two spatial terms.
GRPO~\citep{grpo} then refines the model with the correctness of the final answer as the reward (\cref{app:dynamics}), letting the end task govern how the reasoning draws on the transformed scene.
\section{Experiments}
\label{sec:experiments}
\afterpage{
\providecommand{\tbd}{\textcolor{red!70}{\scriptsize TBD}}

\begin{table*}[t]
\centering
\caption{\small Accuracy (\%) on spatial reasoning benchmarks. \textbf{Bold} and \underline{underline} mark the best and the second-best results among the spatial reasoning models and the Qwen3.5-4B-based models, and separately among the Qwen3.5-9B-based models; the general models (in gray) are listed for reference and not ranked. $^\dagger$Evaluated by us through the provider's API. $^\ddagger$Released weights evaluated by us. $^\S$Taken from the EASI leaderboard~\citep{easi}. More details are in \cref{app:baselines}.}
\label{tab:main}
\setlength{\tabcolsep}{4pt}
\resizebox{\textwidth}{!}{%
\begin{tabular}{@{}lcccccccccc@{}}
\toprule
& \textbf{Average} & \multicolumn{4}{c}{\textbf{MindCube-Tiny}} & \textbf{SPAR-Bench} & \textbf{BLINK} & \multicolumn{3}{c}{\textbf{ViewSpatial}} \\
\cmidrule(lr){2-2} \cmidrule(lr){3-6} \cmidrule(lr){7-7} \cmidrule(lr){8-8} \cmidrule(l){9-11}
Model & Avg. & Among & Around & Rotation & Overall & R\&V & Spatial & Camera & Person & Overall \\
\midrule
\multicolumn{11}{c}{\textit{General Models}} \\
\midrule
\rowcolor{gray!12} GPT-6-Astra$^\dagger$               & 73.58 & 78.33 & 86.80 & 98.50 & 84.19 & 70.17 & 83.00 & 53.16 & 60.55 & 56.97 \\
\rowcolor{gray!12} Claude-Opus-5.5$^\dagger$           & 74.15 & 79.83 & 82.40 & 98.00 & 83.90 & 71.95 & 84.75 & 50.49 & 61.19 & 56.00 \\
\rowcolor{gray!12} Hy4$^\dagger$                       & 77.39 & 89.83 & 86.80 & 97.50 & 90.57 & 71.92 & 91.25 & 53.88 & 57.63 & 55.81 \\
\rowcolor{gray!12} Qwen3.8-Flash$^\dagger$             & 73.98 & 69.83 & 84.40 & 95.50 & 78.19 & 69.84 & 90.00 & 55.87 & 59.77 & 57.88 \\
\rowcolor{gray!12} GLM-5.3-Flash$^\dagger$            & 66.23 & 56.67 & 73.20 & 95.00 & 67.90 & 67.49 & 74.75 & 54.86 & 54.67 & 54.76 \\
\rowcolor{gray!12} Kimi-K3.1-Pro$^\dagger$            & 72.06 & 69.33 & 84.00 & 96.00 & 77.90 & 66.85 & 88.25 & 54.35 & 56.07 & 55.23 \\
\midrule
\multicolumn{11}{c}{\textit{Spatial Reasoning Models}} \\
\midrule
SpatialLadder-3B$^\ddagger$        & 45.25 & 44.33 & 58.40 & 29.50 & 44.86 & 36.68 & 60.50 & 36.29 & 41.49 & 38.97 \\
Cambrian-S-7B$^\S$          & 47.12 & 38.98 & 39.20 & 33.00 & 37.88 & 45.31 & 64.00 & 40.45 & 42.07 & 41.28 \\
HATCH-3B$^\ddagger$               & 54.89 & 52.33 & 64.00 & 34.00 & 51.62 & 57.99 & 71.75 & 40.92 & 35.64 & 38.20 \\
GeoVR-4B$^\ddagger$               & 44.16 & 39.50 & 46.80 & 31.00 & 39.62 & 43.60 & 52.50 & 32.86 & \underline{48.49} & 40.91 \\
Spatial-Interactor-4B$^\ddagger$   & 54.92 & 83.00 & 85.20 & 61.00 & 79.33 & 39.71 & 58.25 & 44.56 & 40.30 & 42.37 \\
GPD-4B$^\ddagger$                  & 53.59 & 42.33 & 63.20 & 27.50 & 44.48 & 54.93 & 69.50 & 46.55 & 44.44 & 45.47 \\
\midrule
\multicolumn{11}{c}{\textit{Qwen3.5-4B Based Models}} \\
\midrule
Qwen3.5-4B                          & 50.57 & 39.83 & 57.20 & 52.00 & 46.29 & 46.17 & 68.25 & 43.48 & 39.79 & 41.58 \\
\quad + SFT                         & 62.74 & 73.67 & 76.80 & 65.00 & 72.76 & 64.85 & 68.75 & 46.84 & 42.51 & 44.61 \\
\quad + GRPO                        & 65.08 & 78.00 & 79.20 & \underline{70.00} & 76.76 & 67.47 & 68.25 & 50.09 & 45.70 & \underline{47.83} \\
\quad + ProSR                       & 65.18 & 78.00 & 80.40 & 66.50 & 76.38 & \underline{67.93} & 68.75 & \underline{50.24} & 45.26 & 47.67 \\
\quad + 3DThinker                   & 62.86 & 75.00 & 76.00 & 67.00 & 73.71 & 64.69 & 66.75 & 47.71 & 44.95 & 46.29 \\
\rowcolor[HTML]{EAE8FD} \quad + \textbf{\method{}} (SFT) & \underline{66.92} & \underline{88.33} & \textbf{90.00} & 54.50 & \underline{82.29} & 64.24 & \underline{75.00} & 47.45 & 44.95 & 46.17 \\
\rowcolor[HTML]{EAE8FD} \quad + \textbf{\method{}} (RL) & \textbf{72.83} & \textbf{92.00} & \underline{88.40} & \textbf{75.50} & \textbf{88.00} & \textbf{71.03} & \textbf{78.00} & \textbf{55.07} & \textbf{53.52} & \textbf{54.27} \\
\midrule
\multicolumn{11}{c}{\textit{Qwen3.5-9B Based Models}} \\
\midrule
Qwen3.5-9B                          & 54.23 & 43.67 & 55.20 & 62.00 & 49.90 & 47.68 & \underline{73.50} & 47.74 & 44.07 & 45.85 \\
\quad + SFT                         & 62.27 & 71.33 & 76.40 & 56.00 & 69.62 & 65.00 & 67.50 & 48.61 & 45.40 & 46.95 \\
\quad + GRPO                        & \underline{67.27} & 79.83 & 78.80 & \textbf{83.00} & 80.19 & \underline{70.67} & 66.00 & \underline{52.22} & \underline{52.19} & \underline{52.21} \\
\rowcolor[HTML]{EAE8FD} \quad + \textbf{\method{}} (SFT) & 66.94 & \underline{88.33} & \underline{86.00} & 64.00 & \underline{83.14} & 66.14 & 70.75 & 48.47 & 47.03 & 47.72 \\
\rowcolor[HTML]{EAE8FD} \quad + \textbf{\method{}} (RL) & \textbf{74.03} & \textbf{91.83} & \textbf{86.80} & \underline{81.50} & \textbf{88.67} & \textbf{71.64} & \textbf{77.75} & \textbf{58.58} & \textbf{57.56} & \textbf{58.05} \\
\bottomrule
\end{tabular}%
}
\end{table*}
}

\subsection{Setup}
\label{sec:setup}

\paragraph{Evaluation Benchmarks.}
We evaluate \method{} on four benchmarks, each targeting a distinct aspect of spatial intelligence: \emph{mental modeling} with MindCube-Tiny~\citep{mindcube}, \emph{relational reasoning} with SPAR-Bench~\citep{spar}, \emph{spatial perception} with BLINK~\citep{blink}, and \emph{perspective taking} with ViewSpatial-Bench~\citep{viewspatial}.
Together, these benchmarks span spatial intelligence from perceiving a scene to reasoning about it from viewpoints that no image shows.
Following EASI~\citep{easi}, we report the accuracy on each benchmark and the average over the four, with the evaluated subsets and the protocol detailed in \cref{app:benchmarks}.

\paragraph{Baselines.}
We compare \method{} against three groups of baselines: \textbf{(I) general models}, including GPT-6-Astra~\citep{gpt6astra}, Claude-Opus-5.5~\citep{claudeopus55}, Hy4~\citep{hy4}, Qwen3.8-Flash~\citep{qwen38flash}, GLM-5.3-Flash~\citep{glm53flash}, and Kimi-K3.1-Pro~\citep{kimik31pro}; \textbf{(II) spatial reasoning models} with released weights, including SpatialLadder~\citep{spatialladder}, Cambrian-S~\citep{cambrians}, HATCH~\citep{hatch}, GeoVR~\citep{geovr}, Spatial-Interactor~\citep{spatialinteractor}, and GPD~\citep{gpd}; and \textbf{(III) methods trained in our setting}, including SFT, GRPO~\citep{grpo}, ProSR~\citep{prosr}, and 3DThinker~\citep{threedthinker}.
Further details of the baselines and their configurations are provided in \cref{app:baselines}.

\paragraph{Training Data.}
All methods trained on our backbone draw on the same three sources, MindCube~\citep{mindcube}, SpatialLadder~\citep{spatialladder}, and SPAR~\citep{spar}.
\method{} uses them in three stages: geometry pre-training on posed image sets from ScanNet~\citep{scannet} and MindCube, SFT on reasoning chains written by a teacher model for questions of the three sources, and RL on multiple-choice question pools built from the same sources and from ScanNet training scenes.
Details of the SFT data are provided in \cref{app:data}.

\paragraph{Implementation Details.}
We build \method{} on Qwen3.5-4B~\citep{qwen35}, and additionally on Qwen3.5-9B to examine its scalability, with $K_s{=}2$ scene tokens per view and $K_c{=}4$ camera tokens.
We set $\lambda{=}0.1$ and train LoRA~\citep{lora} adapters in all stages.
We report \method{} after its SFT stage and after its RL stage, denoted \method{}~(SFT) and \method{}~(RL).
Detailed training setups and parameter configurations are listed in \cref{app:training}.

\begin{table}[t]
\begin{minipage}[t]{0.6\linewidth}
\vspace{0pt}
\caption{Inference cost per question on Qwen3.5-4B, with eight concurrent questions on one H100 GPU: the calls
to the backbone and to external tools, the wall-clock time, the speed-up over \mbox{SpatialClaw}~\citep{spatialclaw}, and the
accuracy (\%). The timing protocol and the full setup are in \cref{app:efficiency}.}
\label{tab:efficiency}
\footnotesize
\setlength{\tabcolsep}{4pt}
\providecommand{\twolines}[2]{\begin{tabular}[b]{@{}r@{}}#1\\#2\end{tabular}}
\providecommand{\ourband}{\kern-\tabcolsep\rlap{\color[HTML]{EAE8FD}\rule[-\arraystretch\dp\strutbox]{\dimexpr\linewidth+2\tabcolsep\relax}{\arraystretch\dimexpr\ht\strutbox+\dp\strutbox\relax}}\kern\tabcolsep}
\begin{tabular*}{\linewidth}{@{\extracolsep{\fill}}lrrrrr@{}}
\toprule
Model & \twolines{Model}{calls} & \twolines{Tool}{calls} & \twolines{Time (s)}{$\downarrow$} & \twolines{Speed-up}{$\uparrow$} & \twolines{Acc.}{$\uparrow$} \\
\midrule
\multicolumn{6}{l}{\textit{MindCube-Tiny (1,050 questions)}} \\
SpatialClaw & 7.74 & 4.68 & 111.74 & 1$\times$ & 56.95 \\
\ourband\method{} & 1 & 0 & \textbf{1.38} & \textbf{81$\times$} & \textbf{88.00} \\
\midrule
\multicolumn{6}{l}{\textit{ViewSpatial (850 questions)}} \\
SpatialClaw & 6.32 & 3.74 & 100.75 & 1$\times$ & 46.71 \\
\ourband\method{} & 1 & 0 & \textbf{0.88} & \textbf{115$\times$} & \textbf{55.88} \\
\bottomrule
\end{tabular*}

\end{minipage}\hfill
\begin{minipage}[t]{0.365\linewidth}
\vspace{0pt}
\centering
\includegraphics[width=\linewidth]{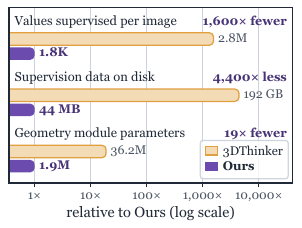}
\captionof{figure}{\textbf{Training \method{} is far lighter than training 3DThinker.} Each row compares one cost of training geometry into the model, as measured in \cref{app:cost}.}
\label{fig:train-cost}
\end{minipage}
\end{table}

\subsection{Main Results}
\label{sec:main-results}

\paragraph{\method{} Achieves the Strongest Performance among Spatial Reasoning Models.}
\Cref{tab:main} shows that \method{} built on Qwen3.5-4B reaches an average accuracy of 72.83\%, surpassing the strongest method trained in our setting, ProSR, by 7.65 points and the best released spatial reasoning model, Spatial-Interactor-4B, by 17.91 points.
The gains are consistent: \method{} ranks first on all four benchmarks, ahead of the strongest baseline on each by 8.67 points on MindCube-Tiny, 3.10 on SPAR-Bench, 6.25 on BLINK, and 6.44 on ViewSpatial.

\paragraph{\method{} Scales to a Larger Backbone and Rivals Frontier General Models.}
Trained with the same recipe on Qwen3.5-9B, \method{} reaches 74.03\%, 6.76 points above GRPO and 11.76 above SFT, and again ranks first on all four benchmarks, by up to 8.48 points on MindCube-Tiny.
As \cref{tab:main} shows, a compact open model stands on par with frontier general models: at 9B, \method{} matches GPT-6-Astra, Qwen3.8-Flash, and Claude-Opus-5.5 (73.58\%, 73.98\%, and 74.15\%), and even at 4B it surpasses GLM-5.3-Flash and Kimi-K3.1-Pro (66.23\% and 72.06\%).

\paragraph{\method{} Is Efficient in Both Inference and Training.}
\label{sec:efficiency}
\Cref{tab:efficiency} compares the inference cost of \method{} with that of SpatialClaw~\citep{spatialclaw}, an agent on the same backbone that reconstructs the scene and segments objects with external tools, calling the backbone 7.74 times and the tools 4.68 times per MindCube-Tiny question.
Holding the scene in its \tokens{}, \method{} needs no tools and answers in a single generation: it is 81$\times$ faster and 31.05 points more accurate on MindCube-Tiny, and 115$\times$ faster and 9.17 points more accurate on ViewSpatial.
This efficiency extends from inference to training: as \cref{fig:train-cost} shows, compared with 3DThinker, which also trains geometry into the model's states, \method{} supervises 1{,}600$\times$ fewer values per image, stores 4{,}400$\times$ less supervision data, and trains a 19$\times$ smaller geometry module, yet it scores 9.97 points higher on average in \cref{tab:main}.

\subsection{Framework Analysis}
\label{sec:analysis}

Beyond accuracy, we further examine whether the gains of \method{} stem from the two capabilities that \cref{sec:diagnosis} identifies as lacking, namely multi-view fusion and perspective transformation.
We first probe whether the model acquires both capabilities and where its backbone fuses the views, then intervene on the \tokens{} to test whether these capabilities causally affect the answers.

\begin{figure}[t]
\centering
\includegraphics[width=\linewidth]{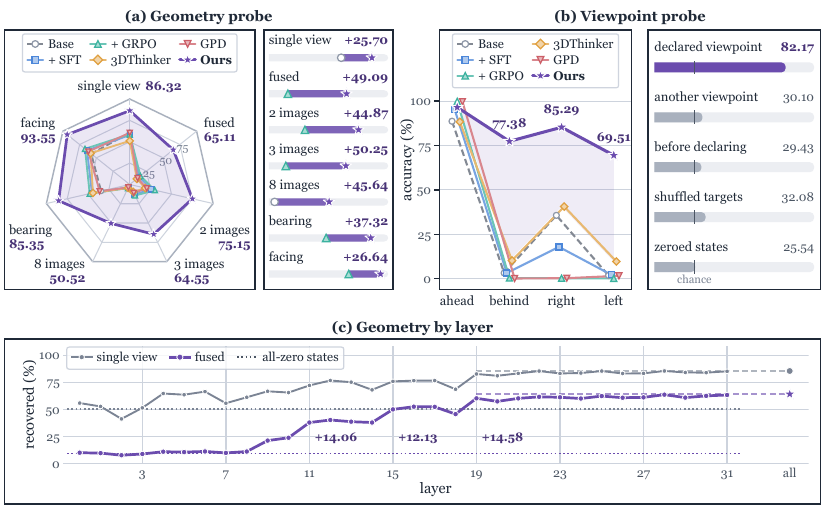}
\caption{\textbf{Analysis of multi-view fusion and perspective transformation in \method{}.} (a)~shows the geometry probe on the baselines and \method{} after SFT (left) and the lead of \method{} over the strongest baseline on each measure (right). (b)~shows the viewpoint probe by the direction of a point from the declared viewer (left) and against four controls (right). (c)~refits the readout of \method{} after geometry pre-training at each layer and on the learned mix of all layers (\emph{all}); vertical lines mark the full-attention layers and dashed lines the level of \emph{all}.}
\label{fig:analysis}
\end{figure}

\begin{figure}[t]
\centering
\includegraphics[width=0.8364\linewidth]{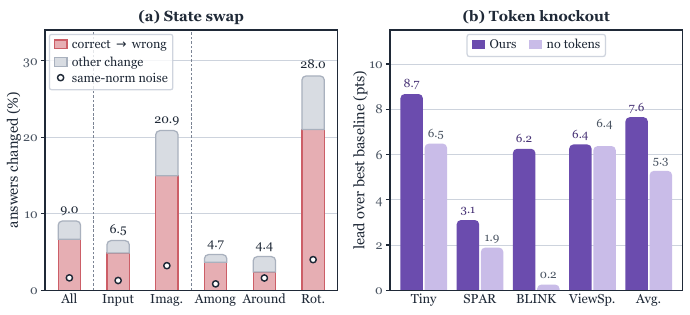}
\caption{\textbf{Interventions on the \tokens{} of \method{}.} Both panels show \method{} after RL. (a)~shows the share of MindCube-Tiny answers that change when the states of the last run of re-emitted \tokens{} are replaced with those of $\mathbf{z}_0$, the scene before any transformation, against a random perturbation of the same norm; questions are grouped by whether the last viewpoint is an input view (\emph{Input}) or an imagined standpoint (\emph{Imag.}), and by setting (\emph{Rot.}: rotation). (b)~shows the lead over the best baseline on each benchmark with and without the \tokens{} at inference. Full results are in \cref{tab:app-knockout,tab:app-swap}.}
\label{fig:analysis-swap}
\end{figure}

\paragraph{The Views Are Fused into One Scene.}
As in \cref{sec:diagnosis}, we retrain the geometry probe on the frozen backbone of \method{} for a fair comparison with the baselines.
\Cref{fig:analysis}a shows that \method{} recovers 65.11\% of the fused points and 86.32\% of the single-view points, whereas the baselines recover at most 16.02\% and 60.62\%.
Cross-view bearing and camera facing improve accordingly, reaching 85.35\% and 93.55\%.
This lead persists as views accumulate: with two images the baselines recover at most 30.28\% of the fused points and with eight fewer than 5\%, while \method{} still recovers 50.52\% with eight.
More importantly, \cref{fig:ablation}b shows that this fusion carries over to the model's answers, raising the accuracy on the 274 two-view questions of \cref{sec:diagnosis}, which hinge on relating the two views, from only 53.28\% and 56.57\% for SFT and GRPO to 96.35\% for \method{}.

\paragraph{The Scene Turns with the Reasoning.}
As in \cref{sec:diagnosis}, we retrain the viewpoint probe on the frozen backbone of \method{} for a fair comparison with the baselines.
\method{} reaches 82.17\% accuracy, whereas the baselines remain between 25.22\% and 37.28\%, close to the probe's controls.
\Cref{fig:analysis}b breaks this gap down by direction: all models recover the points ahead of the declared viewer, but the baselines recover at most 10.29\%, 9.78\%, and 40.69\% of those behind it, to its left, and to its right, compared with 77.38\%, 69.51\%, and 85.29\% for \method{}.
The decoded scene is bound to the declaration: as the controls in \cref{fig:analysis}b show, the accuracy falls to 30.10\% when another viewpoint of the chain is the target and to 29.43\% when the probe reads tokens before the declaration.

\paragraph{The Backbone Fuses the Views in Its Middle Layers.}
The readout leaves the fusion to the backbone by design: its queries carry no image content, and each point query attends only to the scene tokens of its own view and to the camera tokens (\cref{app:readout}).
To locate where the backbone fuses the views, we take \method{} on Qwen3.5-4B after geometry pre-training and refit its readout on the states of $\mathbf{z}_0$ at one layer at a time, scoring it on the 416 ScanNet image sets of the geometry probe.
\Cref{fig:analysis}c shows that the fused points stay at the level of a readout fit on all-zero states through layer 8 (7.64--11.12\% against 9.21\%) and then rise through the middle layers, with their three largest gains, of 12.13 to 14.58 points, at the full-attention layers 11, 15, and 19 of the hybrid backbone.
From layer 19 on, every single layer recovers nearly as many fused points as the learned mix of all layers (57.33--63.54\% against 63.99\%), so the fused scene, once formed, is carried through to the last layer.

\paragraph{The Answers Depend on the Scene.}
Decodability, however, does not imply use.
We therefore intervene on the RL model at inference, replacing the states of the last re-emitted \tokens{} in its own generation with those of $\mathbf{z}_0$.
As \cref{fig:analysis-swap}a shows, this state swap changes 9.05\% of the answers, mostly from correct to wrong, against only 1.62\% for a random perturbation of the same norm (\cref{app:swap}).
The effect is strongest where no input image can substitute for the transformed scene: 20.86\% of the answers change at imagined standpoints and 28.00\% in the rotation setting, whose views barely overlap.
Withholding all \tokens{} at inference likewise lowers the four-benchmark average by 2.37 points, from 72.83\% to 70.46\%.
Despite this drop, \cref{fig:analysis-swap}b shows that \method{} without the tokens outperforms the best baseline on every benchmark, by 5.28 points on average, indicating that the capability is internalized in the backbone, not merely carried by the \tokens{}.

\subsection{Ablation Study and Sensitivity Analysis}
\label{sec:ablation}

\begin{figure}[t]
\centering
\includegraphics[width=\linewidth]{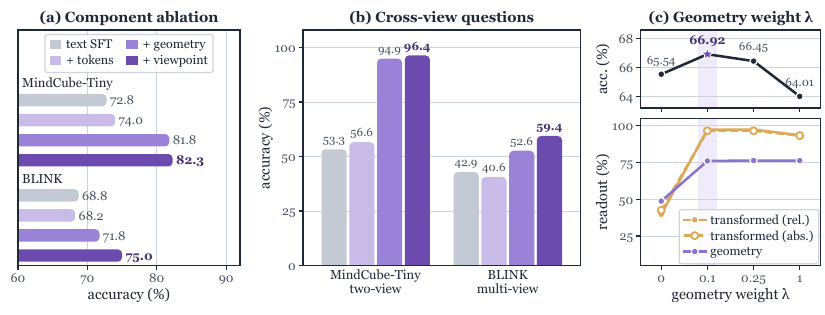}
\caption{\textbf{Ablation and sensitivity analysis.} All variants are built on Qwen3.5-4B and evaluated after SFT on the same data. (a)~adds components to text SFT one at a time: \tokens{} trained only through the language loss (\emph{+\,tokens}), geometry pre-training and loss on $\mathbf{z}_0$ (\emph{+\,geometry}), and the view loss on every re-emitted $\mathbf{z}_k$ (\emph{+\,viewpoint}). (b)~repeats (a) on the questions that relate the views. (c)~varies the geometry weight $\lambda$, with the average accuracy at the top and the geometry and transformation readouts below; the shaded band marks $\lambda{=}0.1$.}
\label{fig:ablation}
\end{figure}

\paragraph{Geometry and Viewpoint Supervision Each Contribute.}
\Cref{fig:ablation}a builds \method{} from text SFT one component at a time.
\textbf{(i)~Unsupervised \tokens{}}, trained solely through the language loss, alter the accuracy by at most 1.24 points, indicating that extra tokens and computation alone contribute little.
\textbf{(ii)~Geometry supervision}, in contrast, yields the largest gain, raising MindCube-Tiny from 74.00\% to 81.81\% and BLINK from 68.25\% to 71.75\%.
\textbf{(iii)~Viewpoint supervision} of every re-emitted $\mathbf{z}_k$ improves both benchmarks further and is most beneficial on BLINK, whose multi-view questions hinge on how the cameras relate, lifting it to 75.00\%.
\Cref{fig:ablation}b further shows the gains on the questions that relate the views: geometry supervision alone lifts accuracy on the 274 two-view questions of MindCube-Tiny from 56.57\% to 94.89\% and on the 133 multi-view questions of BLINK from 40.60\% to 52.63\%, which viewpoint supervision then raises to 59.40\%.

\paragraph{Preserving the Scene Requires Only a Small Geometry Weight.}
\Cref{fig:ablation}c varies the weight $\lambda$ on the two spatial losses during SFT.
Without these losses ($\lambda{=}0$), SFT erodes the geometry acquired in pre-training: the 3D points recovered from $\mathbf{z}_0$ fall to 48.79\%, against 76.13\% at $\lambda{=}0.1$.
The re-emitted \tokens{} are not transformed either: only 39.85\% decode nearest the declared camera's view of the scene in $\mathbf{z}_0$, against 97.49\%, and 42.55\% nearest its view of the true scene, against 96.67\%.
The average accuracy drops accordingly, to 65.54\%, 1.38 points below that at $\lambda{=}0.1$.
Any positive weight preserves the geometry (76.13--76.41\%), but larger weights steadily lower accuracy, from 66.92\% at $\lambda{=}0.1$ to 64.01\% at $\lambda{=}1$.
We adopt $\lambda{=}0.1$, the smallest weight that preserves the scene.

\par\penalty0
\FloatBarrier
\section{Conclusion and Future Work}
\label{sec:conclusion}

\paragraph{Conclusion.}
We introduced \method{}, a framework that internalizes multi-view geometry in MLLMs through a compact set of \tokens{}.
Our spatial probes first showed that the accuracy gains of spatially fine-tuned models conceal two persistent deficiencies: the views are barely fused into a unified internal scene, and this scene rarely follows the reasoning to new viewpoints.
\method{} addresses both by supervising the \tokens{} to hold the fused 3D scene and re-emitting them to re-express this scene from every viewpoint the reasoning adopts.
Across four spatial benchmarks and two model scales, \method{} achieves the strongest performance among spatial reasoning models, rivals frontier general models, and remains highly efficient at inference.
The same probes and causal interventions trace these gains to their source: both capabilities take shape within the model and directly steer its predictions.
These results establish internalized multi-view geometry as a learnable, efficient, and inspectable source of spatial intelligence in MLLMs.

\paragraph{Future Work.}
We view internalized geometry as a cornerstone of spatial intelligence, and \method{} as one step toward models that think in space.
The present work supervises only the geometry that the input views observe.
A natural next step is an internal scene that reaches beyond these views, completing unseen regions, persisting across interactions, and evolving with dynamic environments, so that the same tokens can support embodied agents over long horizons.
As this paradigm matures, spatial intelligence may shift from geometry computed outside the model to a property that models carry within themselves, opening a research agenda around internal world models, inspectable spatial states, and the joint scaling of perception and reasoning.

\section*{Acknowledgments}
We express our deepest gratitude to Jianing Ju and Wentao Zhang for the many extensive and insightful discussions throughout this project. Their thoughtful feedback and constructive suggestions were invaluable, sharpening our ideas and substantially improving this work.

\bibliographystyle{plainnat}
\bibliography{references}

\clearpage
\appendix
\crefalias{section}{appendix}
\crefalias{subsection}{appendix}
\section{Diagnosis Details}
\label{app:diagnosis}

This appendix details the probes of \cref{sec:diagnosis} and reports their full results.

\subsection{Probe Setup}
\label{app:probe-setup}

In this section, we describe the models we diagnose and the probes we apply to them:

\begin{itemize}
    \item \textbf{Models.} \emph{Base} is Qwen3.5-4B~\citep{qwen35}. \emph{+\,SFT} is Base fine-tuned with LoRA~\citep{lora} on reasoning chains written by a teacher model, and \emph{+\,GRPO} is trained for 375 steps of GRPO~\citep{grpo} with a reward on the final answer. They stand for the two common ways of teaching an MLLM spatial reasoning from text and answers alone. We also probe two released spatial reasoning models: 3DThinker~\citep{threedthinker} (its released MindCube checkpoint), built on Qwen2.5-VL-3B, and GPD~\citep{gpd}, built on Qwen3-VL-4B. For them, the S-Space axes are those of each model's backbone, and each model answers in its own reasoning format, decoded greedily with at most 4{,}096 new tokens and a strict extraction of the answer.

    \item \textbf{Geometry probe.} We freeze the whole model and append probe tokens after the images, two per image and four for the cameras. Only the input embeddings of these tokens are trained, together with a readout of a single cross-attention layer, on posed image sets, with the 3D point behind every visual token and the pose of every camera in the frame of the first image as targets. Whatever is read from the states of the probe tokens was therefore computed by the frozen layers from the images. We then refit a fresh readout on the frozen states and score it on 416 held-out ScanNet~\citep{scannet} image sets, 364 of them with more than one view. We report four quantities. \emph{Single-view points} are the points of the first image, counted as recovered when they fall within a quarter of the set's mean point distance. \emph{Fused points} are the points of the later images under the same criterion, which can be recovered only by relating those images to the first. \emph{Cross-view bearing} is the floor-plane direction, in four classes, between the points seen at the same grid position in the first image and in another one. \emph{Camera facing} records whether the optical axis of a later camera is recovered within $45^\circ$. Three controls bound what a readout achieves without the geometry, namely shuffling the targets among sets that share source, task, and number of views, setting the states to zero, and replacing the images with gray ones.

    \item \textbf{Viewpoint probe on SpinBench.} S-Space~\citep{sspace} finds three linear axes in a middle layer of an MLLM along which the positions of objects are encoded. We take the axes of Base at layer 15 and use them unchanged for all three models, so that a difference between models is not a difference between axes. SpinBench~\citep{spinbench} asks for the relation between two objects as seen from a viewpoint other than the camera's, with or without a stated premise (144 and 146 questions). On it we compare two answers to the same question: the \emph{readout}, which decodes the coordinates of the two objects from the state and rotates them to the queried viewpoint outside the model, and the model's own \emph{answer}. We then replay each model's own reasoning and read the state at every mention of the two objects. A \emph{camera-frame} readout is right when the state gives the answer after the external rotation, whereas a \emph{target-frame} readout is right when the state gives the answer with no rotation, that is, when the model has already transformed it to the queried viewpoint. This part uses the questions on which the reasoning mentions both objects at least twice, so that the first and the last mention can be compared.

    \item \textbf{Viewpoint probe on reasoning chains.} Reasoning chains are teacher-forced through the frozen model, and the probe tokens of the geometry probe, which were never fit to a transformed target, are inserted after every viewpoint a chain declares. A fresh readout is fit to the scene in the frame of the declared viewpoint and scored on 296 held-out chains with 653 declared viewpoints. We score the points that come from the images other than the viewer's own by the egocentric quadrant they fall in, as a balanced accuracy over the four quadrants. The controls are the targets of the other viewpoints declared in the same chain, the tokens before any declaration, zero states, and targets shuffled among chains. \Cref{tab:analysis-probe} gives each model's strongest control in parentheses.
\end{itemize}

\subsection{Results on Multi-View Fusion}
\label{app:diag-fusion}

\Cref{tab:analysis-probe} gives the full results of the geometry probe behind Finding 1, for the Qwen3.5-4B models and for the released 3DThinker and GPD, with \method{}, discussed in \cref{sec:analysis}, in its last column.
From Base it recovers 60.62\% of the points of the first image, against 53.79\% for its strongest control, but only 11.23\% of the points of the later images, against 9.06\%, while cross-view bearing (34.41\% against 33.43\%) and camera facing (57.59\% against 56.73\%) stay within a point of their controls.
Since the probe tokens were free to gather anything the frozen layers offer, the frozen model computes hardly any fused geometry that can be read out.
SFT and GRPO raise the fused points to 14.19\% and 16.02\% and the bearing to 45.31\% and 48.03\%.
The gain sits in the sets of two images (25.94\% and 30.28\%, against controls of 12.19\% and 13.22\%).
With three images it is within five points of the control, and with eight images all three models are at control.
The held-out ScanNet sets number 100 with two images, 250 with three, and 14 with eight. Fusion is expected to get harder as images are added, and the question is how much of it a model keeps.
For the released models, the fused points recovered from 3DThinker and GPD, 10.65\% and 11.82\%, likewise stay close to their strongest controls, 9.57\% and 9.59\%.

\begin{table}[!ht]
\centering
\caption{The probes of \cref{sec:diagnosis} on the Qwen3.5-4B models, released spatial reasoning models, and
\method{} after its SFT stage (\%), each followed in parentheses by the strongest of its controls
(\cref{app:probe-setup}). Top: the frozen-backbone geometry probe; for \method{}, its own \tokens{}
are discarded and fresh probe tokens trained in their place. Bottom: the viewpoint probe, which reads the points of
the other images in the frame of the viewpoint just declared (balanced accuracy over the four
egocentric quadrants).}
\label{tab:analysis-probe}
\footnotesize
\setlength{\tabcolsep}{3.5pt}
\begin{tabular}{lcccccc}
\toprule
 & Base & + SFT & + GRPO & 3DThinker & GPD & \textbf{Ours} \\
\midrule
Single-view points & 60.62 {\scriptsize\color{gray}(53.79)} & 57.83 {\scriptsize\color{gray}(53.79)} & 58.69 {\scriptsize\color{gray}(53.79)} & 51.01 {\scriptsize\color{gray}(53.43)} & 60.48 {\scriptsize\color{gray}(53.79)} & 86.32 {\scriptsize\color{gray}(53.79)} \\
Fused points & 11.23 {\scriptsize\color{gray}(9.06)} & 14.19 {\scriptsize\color{gray}(9.16)} & 16.02 {\scriptsize\color{gray}(9.46)} & 10.65 {\scriptsize\color{gray}(9.57)} & 11.82 {\scriptsize\color{gray}(9.59)} & 65.11 {\scriptsize\color{gray}(9.03)} \\
\quad 2 images & 19.21 {\scriptsize\color{gray}(13.38)} & 25.94 {\scriptsize\color{gray}(12.19)} & 30.28 {\scriptsize\color{gray}(13.22)} & 18.54 {\scriptsize\color{gray}(14.03)} & 20.36 {\scriptsize\color{gray}(14.91)} & 75.15 {\scriptsize\color{gray}(12.01)} \\
\quad 3 images & 10.26 {\scriptsize\color{gray}(8.99)} & 12.88 {\scriptsize\color{gray}(9.21)} & 14.30 {\scriptsize\color{gray}(9.36)} & 9.68 {\scriptsize\color{gray}(9.36)} & 10.92 {\scriptsize\color{gray}(9.19)} & 64.55 {\scriptsize\color{gray}(8.99)} \\
\quad 8 images & 4.88 {\scriptsize\color{gray}(3.47)} & 3.73 {\scriptsize\color{gray}(3.47)} & 4.63 {\scriptsize\color{gray}(4.42)} & 4.46 {\scriptsize\color{gray}(3.70)} & 3.69 {\scriptsize\color{gray}(3.47)} & 50.52 {\scriptsize\color{gray}(4.12)} \\
Cross-view bearing & 34.41 {\scriptsize\color{gray}(33.43)} & 45.31 {\scriptsize\color{gray}(34.25)} & 48.03 {\scriptsize\color{gray}(34.49)} & 44.17 {\scriptsize\color{gray}(36.44)} & 36.08 {\scriptsize\color{gray}(33.37)} & 85.35 {\scriptsize\color{gray}(30.47)} \\
Camera facing & 57.59 {\scriptsize\color{gray}(56.73)} & 63.90 {\scriptsize\color{gray}(54.30)} & 66.91 {\scriptsize\color{gray}(56.73)} & 58.60 {\scriptsize\color{gray}(54.01)} & 61.89 {\scriptsize\color{gray}(55.16)} & 93.55 {\scriptsize\color{gray}(56.45)} \\
\midrule
Declared viewpoint & 32.40 {\scriptsize\color{gray}(29.13)} & 29.72 {\scriptsize\color{gray}(26.92)} & 25.22 {\scriptsize\color{gray}(29.48)} & 37.28 {\scriptsize\color{gray}(28.78)} & 25.49 {\scriptsize\color{gray}(29.74)} & 82.17 {\scriptsize\color{gray}(32.08)} \\
\quad camera viewpoints & 31.76 {\scriptsize\color{gray}(30.46)} & 28.91 {\scriptsize\color{gray}(27.44)} & 25.07 {\scriptsize\color{gray}(30.71)} & 35.03 {\scriptsize\color{gray}(30.17)} & 25.10 {\scriptsize\color{gray}(31.13)} & 85.46 {\scriptsize\color{gray}(33.38)} \\
\quad imagined standpoints & 33.61 {\scriptsize\color{gray}(25.68)} & 31.37 {\scriptsize\color{gray}(25.50)} & 25.57 {\scriptsize\color{gray}(26.10)} & 41.92 {\scriptsize\color{gray}(25.97)} & 26.26 {\scriptsize\color{gray}(25.90)} & 73.90 {\scriptsize\color{gray}(28.69)} \\
\bottomrule
\end{tabular}
\end{table}

\Cref{tab:diag-sspace}a shows that the single-image readout of S-Space is likewise unchanged by fine-tuning: it gives the relation between two objects of one image on 86.50\% of the 1{,}200 EmbSpatial~\citep{embspatial} questions for Base, 86.75\% after SFT, and 87.17\% after GRPO, neither significantly different from Base.
That the state can be read view by view does not contradict the missing fusion.
On a two-view set of the S-Space suite whose questions name one of the views, the readout is right on 82.50\% for Base (78.17\% and 76.83\% after SFT and GRPO, both significantly lower), but it reads each object in the frame of its own image and needs no relation between the two.
The readouts of 3DThinker and GPD give 78.42\% and 86.83\% on EmbSpatial and 72.33\% and 71.00\% on the two-view set.

\FloatBarrier
\subsection{Results on Perspective Transformation}
\label{app:diag-transform}

\cref{tab:diag-sspace} gives the full results behind Finding 2 for the same models.
The main text uses the questions with a premise, which state the relation as the camera sees it. The questions without one, the setting of the original experiment~\citep{sspace}, show the same pattern in \cref{tab:diag-sspace}c.
On both variants the readout with the external rotation is right on 89.73--97.92\% of the questions, while the models' own answers are right on 67.12--79.86\%, a gap of 15.97--25.70 points for every model.
SFT and GRPO do not close it: their readouts stay within 2.08 points of Base, and their answers differ from those of Base by $-7.53$ to $+7.64$ points, every interval including zero.

From the first to the last mention of the objects the target-frame readout rises significantly for every model (exact McNemar test), and the camera-frame readout falls significantly after SFT and GRPO but not for Base.
On the questions a model answers wrong, the camera-frame readout at the first mention is right in 96.00--100.00\% of the cases with premise and in 81.08--89.36\% without, so the scene was perceived correctly.
For Base the target-frame readout at the last mention is still right in 26.47\% of them (48.15\% without premise): its answer often does not follow its state, and the two agree on only 64.71\% of all questions.
SFT and GRPO make the answer follow the state (81.62\% and 74.02\% agreement), and their wrong answers are then the questions on which the state was not transformed (2.94\% and 4.00\%, and 8.51\% and 13.51\% without premise).
The columns of (b) and (c) cover different questions for each model, those on which its own reasoning mentions both objects at least twice, so they describe each model separately and are not paired across models.

The released models show the same pattern.
Their own answers on SpinBench remain 19.18--36.80 points below what the state gives after the external rotation, a gap at least as large as for the Qwen3.5-4B models.
Replaying their reasoning on the questions with premise, the share of states in the target frame goes from 36.36\% at the first mention to 33.33\% at the last for 3DThinker (33 questions) and from 31.21\% to 35.46\% for GPD (141).
The viewpoint probe on reasoning chains in \cref{tab:analysis-probe} agrees: GPD stays below its strongest control (25.49\% against 29.74\%), and 3DThinker exceeds its own by 8.50 points (37.28\% against 28.78\%), far short of the 82.17\% of \method{}.

\begin{table}[!ht]
\centering
\caption{Perspective transformation in the Qwen3.5-4B models and in released spatial reasoning models: the viewpoint
probe on SpinBench (accuracy, \%), each model read along the S-Space axes of its backbone (layer 15 of Qwen3.5-4B, layer
22 of Qwen2.5-VL-3B for 3DThinker, layer 17 of Qwen3-VL-4B for GPD). \emph{Readout} decodes the two objects'
coordinates from the state; on SpinBench it then rotates them to the queried viewpoint outside the model.
\emph{Answer} is the model's own answer after reasoning. (b) and (c) read the state at the first and the last
mention of the objects in the model's own reasoning: \emph{target frame} is the share of states that give the
answer with no rotation (already transformed to the queried viewpoint), \emph{camera frame} the share that give it
after the external rotation (still in the camera's frame).}
\label{tab:diag-sspace}
\footnotesize
\setlength{\tabcolsep}{5pt}
\begin{tabular}{lccccc}
\toprule
 & \multicolumn{3}{c}{Qwen3.5-4B} & \multicolumn{2}{c}{Released models} \\
\cmidrule(lr){2-4} \cmidrule(lr){5-6}
 & Base & + SFT & + GRPO & 3DThinker & GPD \\
\midrule
\multicolumn{6}{l}{\textit{(a) Readout, and the model's own answer}} \\
Single image (EmbSpatial): readout & 86.50 & 86.75 & 87.17 & 78.42 & 86.83 \\
Named view of two: readout & 82.50 & 78.17 & 76.83 & 72.33 & 71.00 \\
SpinBench with premise: readout & 97.92 & 97.22 & 95.83 & 84.72 & 100.00 \\
SpinBench with premise: answer & 72.22 & 76.39 & 79.86 & 47.92 & 69.44 \\
SpinBench without premise: readout & 91.10 & 89.73 & 91.78 & 70.55 & 78.77 \\
SpinBench without premise: answer & 74.66 & 67.12 & 71.23 & 51.37 & 55.48 \\
\midrule
\multicolumn{6}{l}{\textit{(b) Along the model's own reasoning, SpinBench with premise}} \\
Questions & 119 & 136 & 127 & 33 & 141 \\
Target frame, first mention & 31.93 & 33.82 & 33.86 & 36.36 & 31.21 \\
Target frame, last mention & 52.94 & 58.09 & 55.91 & 33.33 & 35.46 \\
Camera frame, first mention & 94.96 & 99.26 & 99.21 & 78.79 & 100.00 \\
Camera frame, last mention & 93.28 & 87.50 & 88.98 & 81.82 & 100.00 \\
Last state agrees with answer & 64.71 & 81.62 & 74.02 & 60.61 & 45.39 \\
Wrongly answered questions & 34 & 34 & 25 & 19 & 42 \\
\quad camera frame, first mention & 100.00 & 97.06 & 96.00 & 73.68 & 100.00 \\
\quad target frame, last mention & 26.47 & 2.94 & 4.00 & 26.32 & 33.33 \\
\midrule
\multicolumn{6}{l}{\textit{(c) Along the model's own reasoning, SpinBench without premise}} \\
Questions & 105 & 140 & 132 & 15 & 126 \\
Target frame, first mention & 40.95 & 37.86 & 44.70 & 33.33 & 38.89 \\
Target frame, last mention & 59.05 & 52.86 & 55.30 & 53.33 & 40.48 \\
Camera frame, first mention & 88.57 & 89.29 & 88.64 & 86.67 & 90.48 \\
Camera frame, last mention & 81.90 & 69.29 & 75.76 & 73.33 & 87.30 \\
Last state agrees with answer & 59.05 & 80.71 & 75.76 & 60.00 & 48.41 \\
Wrongly answered questions & 27 & 47 & 37 & 5 & 60 \\
\quad camera frame, first mention & 85.19 & 89.36 & 81.08 & 80.00 & 86.67 \\
\quad target frame, last mention & 48.15 & 8.51 & 13.51 & 40.00 & 41.67 \\
\bottomrule
\end{tabular}
\end{table}

\FloatBarrier
\subsection{Results across Model Scale}
\label{app:diag-scale}

\begin{figure*}[!ht]
\centering
\includegraphics[width=0.89\textwidth]{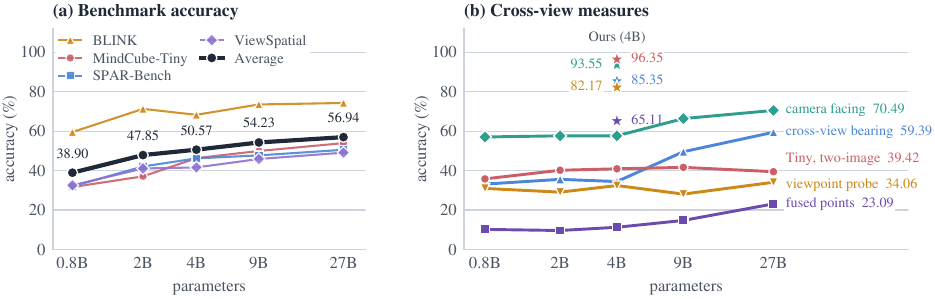}
\caption{Qwen3.5 Base models from 0.8B to 27B parameters, without any training. (a) Accuracy on the four benchmarks and their average (bold, with its values). (b) On the same scale, what the model holds across views: the two-image questions of MindCube-Tiny, the cross-view measures of the geometry probe (fused points, cross-view bearing, and camera facing), and the viewpoint probe. Each line is labeled with its value at 27B. Stars: \method{} on Qwen3.5-4B after its SFT stage, on the same five measures.}
\label{fig:app-scale}
\end{figure*}

We ask whether the two deficiencies of \cref{sec:diagnosis} recede as the model grows.
We take the Base models of the Qwen3.5 family~\citep{qwen35} at 0.8B, 2B, 4B, 9B, and 27B parameters, train nothing, and evaluate each on the four benchmarks under the protocol of \cref{tab:main} and with the geometry and viewpoint probes of \cref{app:probe-setup}, run with the arguments of the 4B probe.
\Cref{fig:app-scale} sets the benchmark accuracy beside the measures of what the model holds across views, on one scale.

\paragraph{Benchmarks.}
As \cref{fig:app-scale}a shows, the average accuracy rises steadily with size, from 38.90\% at 0.8B to 47.85\%, 50.57\%, 54.23\%, and 56.94\%.
The rise is uneven within a benchmark, as \cref{fig:app-scale}b shows: on the 274 questions of MindCube-Tiny that show only two images, the accuracy stays between 35.77\% and 41.61\% and is 39.42\% at 27B, while the whole benchmark rises from 31.62\% to 53.90\%.

\paragraph{Geometry probe.}
Fusion stays near the control up to 4B, where the fused points are 11.23\% against 9.06\%, and emerges only from 9B on: the fused points reach 14.70\% at 9B and 23.09\% at 27B, with cross-view bearing at 49.55\% and 59.39\% and camera facing at 66.33\% and 70.49\%, all far below the 65.11\%, 85.35\%, and 93.55\% of \method{} at 4B.

\paragraph{Viewpoint probe.}
At no size does the state follow the declared viewpoint: the balanced accuracy is 30.93\%, 29.04\%, 32.40\%, 28.05\%, and 34.06\% from 0.8B to 27B, far below the 82.17\% of \method{}.

\paragraph{Summary.}
Scale alone thus does not resolve the two deficiencies.
While the average accuracy gains at every step in size, 18.04 points from 0.8B to 27B, the two-image questions and the viewpoint probe move little over the whole range ($+3.65$ and $+3.13$ points), and the cross-view measures of the geometry probe rise only from 9B on ($+11.86$, $+24.98$, and $+12.90$ points from 4B to 27B for the fused points, cross-view bearing, and camera facing).
For reference, the stars in \cref{fig:app-scale}b mark our 4B model after SFT, well above the Base model of every size, 27B included, on all five measures.

\FloatBarrier
\section{Geometry Readout}
\label{app:readout}

\begin{figure*}[!ht]
\centering
\includegraphics[width=\textwidth]{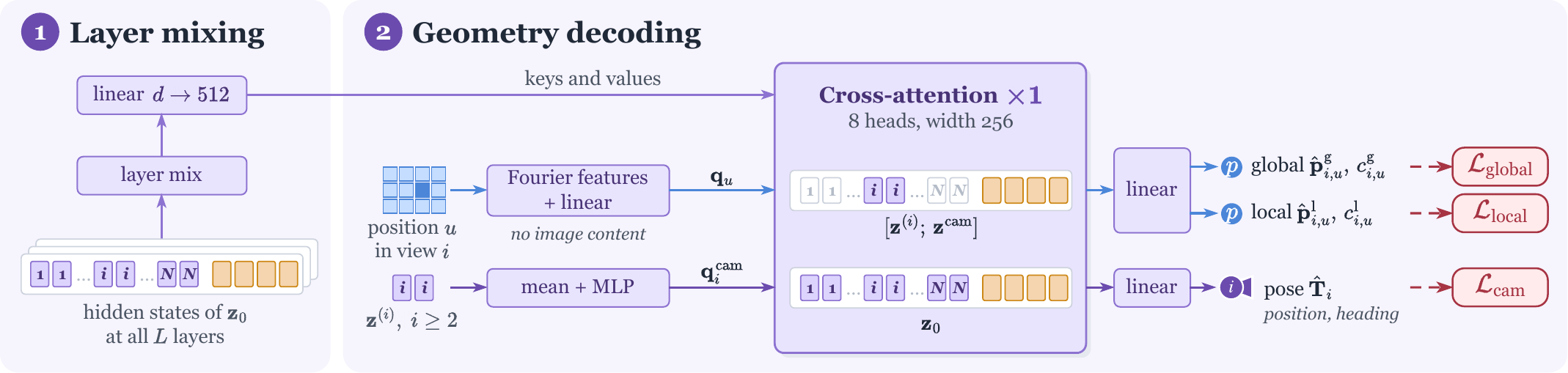}
\caption{Architecture of the geometry readout $\mathcal{R}_\phi$. \badge{1}~\emph{Layer mixing}: the hidden states of $\mathbf{z}_0$ at all $L$ layers are mixed and projected into the keys and values of a single cross-attention layer. \badge{2}~\emph{Geometry decoding}: point queries built from the grid position of a visual token and camera queries built from the scene tokens of a view share this layer. In each row of the layer, the colored tokens are the keys and values that the corresponding query attends to. Linear heads decode the points and poses supervised by $\mathcal{L}_{\mathrm{global}}$, $\mathcal{L}_{\mathrm{local}}$, and $\mathcal{L}_{\mathrm{cam}}$.}
\label{fig:app-readout}
\end{figure*}

\paragraph{Architecture.}
As \cref{fig:app-readout} shows, the readout $\mathcal{R}_\phi$ of \cref{sec:method-fusion} first mixes the hidden states of $\mathbf{z}_0$ across all $L$ layers of the backbone with learned softmax weights and projects them from the hidden width $d$ to 512.
The projected states supply the keys and values of a single cross-attention layer with 8 heads and width 256, shared by two kinds of queries.
The point query $\mathbf{q}_u$ encodes only the grid position $u$ of a visual token, through sinusoidal features at six frequencies and a linear layer, and attends to $[\mathbf{z}^{(i)};\mathbf{z}^{\mathrm{cam}}]$, the scene tokens of its own view together with the camera tokens.
The camera query $\mathbf{q}^{\mathrm{cam}}_i$ is computed by an MLP from the mean of $\mathbf{z}^{(i)}$ and attends to all of $\mathbf{z}_0$, for every camera except the first, which defines the reference frame.
No parameter is tied to a view index, so the same readout serves any number of views.
Linear heads map the attention outputs to the global and local points of every visual token, each with a confidence $c=1+e^{s}$ from a predicted logit $s$, and to the floor position and heading of every decoded camera.
All coordinates lie in floor-aligned frames, with $x$ pointing right, $y$ forward, and $z$ up, and are normalized by the mean point distance of the image set, as in VGGT~\citep{vggt}.
The readout is deliberately weak, with queries that carry no image content, a single attention layer without a residual connection, and 1.9M parameters on \mbox{Qwen3.5-4B} and 2.7M on \mbox{Qwen3.5-9B}.
The intent is to leave the composition of per-view depth with camera poses to the \tokens{} rather than to the readout.

\FloatBarrier
\section{Experimental Details}
\label{app:exp-details}

\subsection{Benchmark Descriptions}
\label{app:benchmarks}

In this section, we describe the four benchmarks used in our experiments:

\begin{itemize}
    \item \textbf{MindCube-Tiny}~\citep{mindcube} has 1,050 questions over two to four views of a scene, in three settings: \emph{among} (600), \emph{around} (250), and \emph{rotation} (200).

    \item \textbf{SPAR-Bench}~\citep{spar} is evaluated on its relational and viewpoint-reasoning tasks (R\&V): 11 of its 20 tasks, 3,969 questions, covering spatial relation, spatial imagination, position matching, camera motion, and distance comparison. As in its official overall score, we report the mean of the per-task accuracies.

    \item \textbf{BLINK}~\citep{blink} is evaluated on its three spatial tasks, multi-view reasoning, relative depth, and spatial relation, on the validation split (400 questions), whose answers are public.

    \item \textbf{ViewSpatial-Bench}~\citep{viewspatial} has 5,712 questions that ask for spatial relations from the perspective of the camera (2,769) and of a person in the scene (2,943). Each group is scored by the accuracy over its questions, as in the benchmark's own results.
\end{itemize}

\paragraph{Evaluation Protocol.}
Every model in \cref{tab:main} that we run from its weights answers without thinking mode, with greedy decoding and at most 4,096 new tokens, and \cref{app:baselines} gives the settings of the general models.
The answer is read by strict matching of the option letter in the requested answer format, without an LLM judge, and an answer that cannot be read is counted as incorrect.
This follows the protocol of EASI~\citep{easi}, and the cited numbers are taken from its public leaderboard\footnote{\url{https://easi.lmms-lab.com/leaderboard/}, as of July 1, 2026.}.
The average in \cref{tab:main} is the mean of the four overall scores.

\subsection{Baseline Setup}
\label{app:baselines}

In this section, we provide details on the three groups of baselines in \cref{tab:main}:

\begin{itemize}
    \item \textbf{General models}, marked $^\dagger$ in \cref{tab:main}, are evaluated through their providers' APIs with our prompts and answer extraction, where a reply that consists of an option letter alone is also accepted. GPT-6-Astra and Claude-Opus-5.5 think at low effort, GLM-5.3-Flash at its default reasoning effort (max), Hy4 and Qwen3.8-Flash run in their default modes via OpenRouter, served by Tencent and Alibaba, and Kimi-K3.1-Pro thinks at low effort and samples at its fixed temperature of 1. The output cap is 32,768 tokens for GLM-5.3-Flash and Qwen3.8-Flash, 16,384 for Kimi-K3.1-Pro, and 65,536 for Hy4; a reply cut at the cap is counted as incorrect.

    \item \textbf{Spatial reasoning models} are listed only if their reported numbers are complete or their final weights are released. In the latter case we evaluate the weights ($^\ddagger$) under the decoding above, with their official prompts and formats, except that SpatialLadder-3B and Spatial-Interactor-4B answer our unified prompt. Cambrian-S's numbers are from the EASI leaderboard.

    \item \textbf{Methods trained in our setting} are reproduced on Qwen3.5-4B from their papers and, where available, their official code, on the same training sources as \method{}. Among them, SFT and GRPO are the key controls: they are trained on the same reasoning chains as \method{} and differ from \method{} in the absence of the \tokens{} and their supervision. On Qwen3.5-9B we repeat the comparison with the base model, SFT, and GRPO.
\end{itemize}

\subsection{SFT Data Construction}
\label{app:data}

In this section, we describe how the data of the SFT stage are constructed: the questions, the reasoning chains, and the viewpoints that the chains declare, which give the targets of the view loss:

\begin{itemize}
    \item \textbf{Question bank.} All methods trained on our backbones draw their questions from one bank of 16,808 multiple-choice questions, namely 8,808 questions from the training set of MindCube~\citep{mindcube} over two to four views, 4,000 questions of SpatialLadder~\citep{spatialladder} on relative direction and relative distance over a single image or eight views, and 4,000 multi-view questions of SPAR-7M~\citep{spar} in seven task types spanning object relation, spatial imagination, camera motion, position matching, and distance comparison. The questions of SpatialLadder and SPAR come from ScanNet~\citep{scannet} training scenes. SpatialLadder's are sampled round-robin over scenes, with the multi-view direction questions stratified so that no option holds more than 35\% of the answers, and SPAR's with at most five questions per scene and task type. The boxes and points that a SPAR question refers to are drawn onto its images with SPAR's official script. All questions share one format: the wording of the source, one option per line, and a single instruction to reason first and then answer with an option letter, with every image resized to at most $640\times480$ pixels. For development we hold out 560 MindCube, 200 SpatialLadder, and 350 SPAR questions by scene, which no training stage uses.

    \item \textbf{Reasoning chains.} The reasoning chains are written by GPT-6-Astra. For each question the teacher receives the images, the question as the student sees it, and a top-down map of the scene in text: the official layout of MindCube or, on ScanNet, the floor positions and extents of the annotated objects visible in the images together with the camera poses. The student never sees the map. The teacher writes one sentence per viewpoint: each sentence opens with a phrase, free of commas, that states where the viewer stands and which way it faces, followed after a comma by what is seen from there, and a final sentence starting with ``So'' draws the conclusion and gives the answer. A chain may mention neither the map nor any coordinate, and may use only the numbers that the question and its options contain. A chain is kept if its answer is correct and it obeys these rules, and each question is sampled at most three times. Of the 7,824 questions sent to the teacher, 7,500 chains are kept, 2,500 per source. For \method{}, a run of \tokens{} is placed after the first comma of every sentence but the last, i.e., right after each declared viewpoint, without changing the teacher's words, and the same chains without \tokens{} are the SFT data of the baselines.

    \item \textbf{Declared viewpoints.} The pose of a declared viewpoint is a position and a heading on the floor plane of the first image's frame, and the view target is $\mathbf{z}_0$'s points re-expressed around it: the origin moves to the position and the forward axis to the heading, while heights are unchanged. When the chain adopts the viewpoint of an input image, the pose is that camera's pose as decoded from $\mathbf{z}_0$. When it stands at a camera but faces elsewhere, the position is the decoded camera's and the heading points to the named object, or follows another camera's decoded heading turned by the stated angle. When it imagines a standpoint, e.g., standing at one object and facing another, the position is the first object's and the heading points to the second. The floor positions of objects come from the annotations of the training scenes, scaled to $\mathbf{z}_0$'s frame, and are the only ground truth in the view target. When it turns relative to where it stands, the previous pose is rotated by the stated angle. The viewpoint is read from the declaring phrase by rules, using only the phrase, the question, and the scene's annotations, never the rest of the chain or the answer, and it serves only as a training target: it never enters the model's input.
\end{itemize}

\subsection{Training Details}
\label{app:training}

This section details the three training stages of \method{}, whose SFT data are described in \cref{app:data}. The parameter settings of all stages are in \cref{tab:hyperparameters}, and every model is trained once. All experiments are conducted on four NVIDIA H100 GPUs.

\begin{itemize}
    \item \textbf{Geometry pre-training.} The \tokens{} of an image set are placed after the images and before the question. Their input embeddings, the readout, and a LoRA~\citep{lora} on every linear layer of the language model are trained together on 16,292 posed image sets from ScanNet and MindCube. The readout is described in \cref{app:readout}. After training, the readout is refit on the frozen model, and this refit readout is the one that SFT keeps frozen.

    \item \textbf{SFT.} SFT starts from the geometry pre-trained model and trains a new LoRA on the reasoning chains, whose viewpoint phrases are each followed by a run of \tokens{}. The loss is the chain's cross-entropy, which leaves out the inserted tokens except the first of every run, plus the geometry loss on $\mathbf{z}_0$ and the view loss of every $\mathbf{z}_k$ against $\mathbf{z}_0$'s own scene seen from the declared viewpoint (\cref{app:data}), both weighted by $\lambda$. Since the first token of every run is predicted, the model learns to write it, and at inference the rest of the run is appended wherever it does.

    \item \textbf{RL.} GRPO~\citep{grpo} starts from the SFT model with a fresh LoRA, on multiple-choice questions from the same sources and from ScanNet training scenes. The reward is the correctness of the final answer under strict extraction, with group-normalized advantages and a KL penalty to the starting model. The \tokens{} inserted during a rollout are context and are not trained.
\end{itemize}

\begin{table*}[!ht]
\centering
\caption{The detailed parameter settings when implementing \method{}.}
\label{tab:hyperparameters}
\footnotesize
\setlength{\tabcolsep}{3.5pt}
\begin{tabular}{lc}
\toprule
\textbf{Parameter} & \textbf{Setting} \\
\midrule
\multicolumn{2}{l}{\textit{All stages}} \\
Scene tokens per view $K_s$ & 2 \\
Camera tokens $K_c$ & 4 \\
LoRA rank & 64 \\
LoRA alpha & 128 \\
LoRA target modules & all linear layers \\
Random seed & 17 \\
Evaluation temperature & 0.0 \\
\midrule
\multicolumn{2}{l}{\textit{Geometry pre-training}} \\
Number of training epochs & 4 \\
Training batch size & 64 \\
Learning rate of the token embeddings & 3e-4 \\
Learning rate of the LoRA & 2e-4 \\
Learning rate of the readout & 1e-3 \\
Learning rate of the layer-mixing weights & 0.05 \\
Confidence weight $\alpha$ & 0.5 \\
\midrule
\multicolumn{2}{l}{\textit{SFT}} \\
Number of training epochs & 2 \\
Training batch size & 64 \\
Learning rate & 2e-4 \\
Warmup ratio & 0.1 \\
Weight of the geometry and view losses $\lambda$ & 0.1 \\
\midrule
\multicolumn{2}{l}{\textit{RL}} \\
Questions per step & 16 \\
Rollouts per question & 8 \\
Training temperature & 1.0 \\
Token limit for responses & 2048 \\
Reward for a correct / wrong answer & 1 / 0 \\
Weight of KL penalty & 0.01 \\
Learning rate & 1e-5 \\
Warmup steps & 4 \\
Gradient clipping & 1.0 \\
\bottomrule
\end{tabular}
\end{table*}

\FloatBarrier
\subsection{Efficiency Setup}
\label{app:efficiency}
\label{app:cost}

\paragraph{Inference.}
We time SpatialClaw and \method{} end to end on the same GPU, one NVIDIA H100 PCIe, with eight questions in flight, the default of SpatialClaw's official code~\citep{spatialclaw}.
The time of a question runs from its start to its final answer, the span that SpatialClaw's runner records.
SpatialClaw runs its official code in its default configuration: Qwen3.5-4B is served by vLLM~\citep{vllm} with thinking enabled at temperature 0.6, and its 3D reconstruction and segmentation tools run on the same GPU.
\method{} decodes greedily as in \cref{tab:main}, and its accuracies are those of \cref{tab:main} on the same questions.
As SpatialClaw takes about 100\,s per question, ViewSpatial is scored on the first 850 of its 5,712 questions in a fixed random order interleaving its five tasks in proportion to their sizes.

\paragraph{Training cost.}
Among the methods trained in our setting, 3DThinker~\citep{threedthinker} is the one that also trains geometry into the model's states, and its supervision is far heavier than ours.
It aligns latent states in the reasoning with the full aggregator features of VGGT~\citep{vggt} for every training image, 1{,}374 tokens of 2{,}048 channels per image (2.8M numbers), which take 192\,GB for our reproduction, through a projector of 36.2M parameters.
Its latent steps feed the last hidden state back as the next input, which standard serving engines do not support, so its rollouts and its inference decode step by step.
\method{} supervises two 3D points per visual token, global and local, and one pose per camera, at most about 1.8K numbers per image and 44\,MB for all 16{,}292 pre-training sets, through a readout of 1.9M parameters, and it runs natively in vLLM.

\FloatBarrier
\section{Additional Experimental Results}
\label{app:results}

This appendix complements \cref{sec:experiments} with the state swap, the per-benchmark token knockout, the design choices behind the two supervised stages, and the geometry along training.
All runs are single runs (seed 17) on Qwen3.5-4B.

\subsection{State Swap}
\label{app:swap}

The swap of \cref{sec:analysis} tests whether the answer depends on the content of the transformed scene rather than on the mere presence of the \tokens{}.
It runs on \method{} after RL over all 1,050 questions of MindCube-Tiny.
The model first answers greedily with its own reasoning.
Its sequence is then cut right after the last run of re-emitted \tokens{}, the run that follows the last viewpoint the reasoning declares, and the model continues greedily from there under one of three conditions, which change the states of that run in every layer and never the text. In the first, the states are left unchanged. In the second, they are replaced by those of $\mathbf{z}_0$ at the same layer and position, i.e., the same scene not re-expressed in the declared frame. In the third, they are moved by random vectors of the same norm as that replacement, a change of the same size that carries no scene.
The last position of the run is left unchanged in all three.
\Cref{tab:app-swap} reports the accuracy and the share of answers that differ from the model's own under each condition.

\begin{table*}[!ht]
\centering
\caption{Swapping the transformed scene at inference (\%, \method{} after RL on Qwen3.5-4B, MindCube-Tiny). After the model has answered, its sequence is cut at the last run of re-emitted \tokens{} and continued with the states of that run unchanged (\emph{None}), replaced in every layer by those of $\mathbf{z}_0$ (\emph{Swap}), or moved by random vectors of the same norm as the swap (\emph{Noise}). \emph{Changed}: answers that differ from the model's own. With the states unchanged, the continuation reproduces 99.71\% of them. $\Delta$: Swap minus Noise in that share, with its 95\% interval from a paired bootstrap over questions (2,000 draws), in bold where the interval excludes zero. The replay decodes with a different inference implementation from that of \cref{tab:main}, so its accuracies with the states unchanged differ from those in \cref{tab:main} by at most 0.80 points.}
\label{tab:app-swap}
\footnotesize
\setlength{\tabcolsep}{3.5pt}
\begin{tabular}{lrrrrrrl}
\toprule
 & & None & \multicolumn{2}{c}{Swap} & \multicolumn{2}{c}{Noise} & \\
\cmidrule(lr){3-3} \cmidrule(lr){4-5} \cmidrule(lr){6-7}
Questions & $n$ & Acc. & Acc. & Changed & Acc. & Changed & $\Delta$ changed [95\% interval] \\
\midrule
All questions & 1,050 & 88.19 & 83.33 & 9.05 & 88.00 & 1.62 & $\mathbf{+7.43}$ [$\mathbf{+5.62}$, $\mathbf{+9.14}$] \\
\midrule
\multicolumn{8}{l}{\textit{By the viewpoint of the last run}} \\
\quad an input image & 863 & 89.92 & 86.56 & 6.49 & 90.03 & 1.27 & $\mathbf{+5.21}$ [$\mathbf{+3.59}$, $\mathbf{+6.95}$] \\
\quad a standpoint no image shows & 187 & 80.21 & 68.45 & 20.86 & 78.61 & 3.21 & $\mathbf{+17.65}$ [$\mathbf{+11.76}$, $\mathbf{+23.53}$] \\
\midrule
\multicolumn{8}{l}{\textit{By setting}} \\
\quad among & 600 & 91.83 & 89.00 & 4.67 & 92.33 & 0.83 & $\mathbf{+3.83}$ [$\mathbf{+2.17}$, $\mathbf{+5.67}$] \\
\quad around & 250 & 89.20 & 88.80 & 4.40 & 88.80 & 1.60 & $\mathbf{+2.80}$ [$\mathbf{+0.40}$, $\mathbf{+5.60}$] \\
\quad rotation & 200 & 76.00 & 59.50 & 28.00 & 74.00 & 4.00 & $\mathbf{+24.00}$ [$\mathbf{+17.50}$, $\mathbf{+30.50}$] \\
\bottomrule
\end{tabular}
\end{table*}

\FloatBarrier
\subsection{Token Knockout}
\label{app:knockout}

\Cref{tab:app-knockout} gives the knockout of \cref{sec:analysis} per benchmark.
The weights are never changed, and only the input at inference differs: all \tokens{} are withheld, so the same weights answer the plain question.
Each knocked-out run is paired with the full model question by question.
After SFT, withholding the \tokens{} costs 2.87 points of the four-benchmark average, with significant losses on MindCube-Tiny, SPAR-Bench, and BLINK.
After RL, it costs 2.37 points, again with significant losses on the same three benchmarks, while the knocked-out model still leads every spatial reasoning model and every Qwen3.5-4B baseline of \cref{tab:main} on each benchmark.

\begin{table*}[!ht]
\centering
\caption{Withholding all \tokens{} at inference, with the weights unchanged (\%, \method{} on Qwen3.5-4B after its SFT stage at the top and after RL at the bottom). $\Delta$: the change against the full model, with 95\% intervals from a paired bootstrap of 2,000 draws that resamples scenes for the rows after SFT, and questions within each task (question groups for MindCube-Tiny) for the rows after RL. Bold marks intervals that exclude zero.}
\label{tab:app-knockout}
\footnotesize
\setlength{\tabcolsep}{3.5pt}
\begin{adjustbox}{max width=\textwidth}
\begin{tabular}{lrrrl}
\toprule
Benchmark & Questions & Full & w/o \tokens{} & $\Delta$ [95\% interval] \\
\midrule
\multicolumn{5}{l}{\textit{After SFT}} \\
MindCube-Tiny & 1,050 & 82.29 & 79.14 & $\mathbf{-3.14}$ [$\mathbf{-5.35}$, $\mathbf{-0.75}$] \\
SPAR-Bench (R\&V) & 3,969 & 64.24 & 62.58 & $\mathbf{-1.66}$ [$\mathbf{-3.08}$, $\mathbf{-0.25}$] \\
BLINK (Spatial) & 400 & 75.00 & 69.00 & $\mathbf{-6.00}$ [$\mathbf{-10.25}$, $\mathbf{-2.00}$] \\
ViewSpatial & 5,712 & 46.17 & 45.47 & $-0.70$ [$-1.75$, $+0.39$] \\
Average & -- & 66.92 & 64.05 & $-2.87$ \\
\midrule
\multicolumn{5}{l}{\textit{After RL}} \\
MindCube-Tiny & 1,050 & 88.00 & 85.81 & $\mathbf{-2.19}$ [$\mathbf{-3.77}$, $\mathbf{-0.69}$] \\
SPAR-Bench (R\&V) & 3,969 & 71.03 & 69.81 & $\mathbf{-1.22}$ [$\mathbf{-2.11}$, $\mathbf{-0.32}$] \\
BLINK (Spatial) & 400 & 78.00 & 72.00 & $\mathbf{-6.00}$ [$\mathbf{-9.50}$, $\mathbf{-2.50}$] \\
ViewSpatial & 5,712 & 54.27 & 54.20 & $-0.07$ [$-1.00$, $+0.82$] \\
Average & -- & 72.83 & 70.46 & $\mathbf{-2.37}$ [$\mathbf{-3.39}$, $\mathbf{-1.34}$] \\
\bottomrule
\end{tabular}
\end{adjustbox}
\end{table*}

\FloatBarrier
\subsection{Design Choices in Geometry Pre-training}
\label{app:design}

\begin{figure*}[!ht]
\centering
\includegraphics[width=\textwidth]{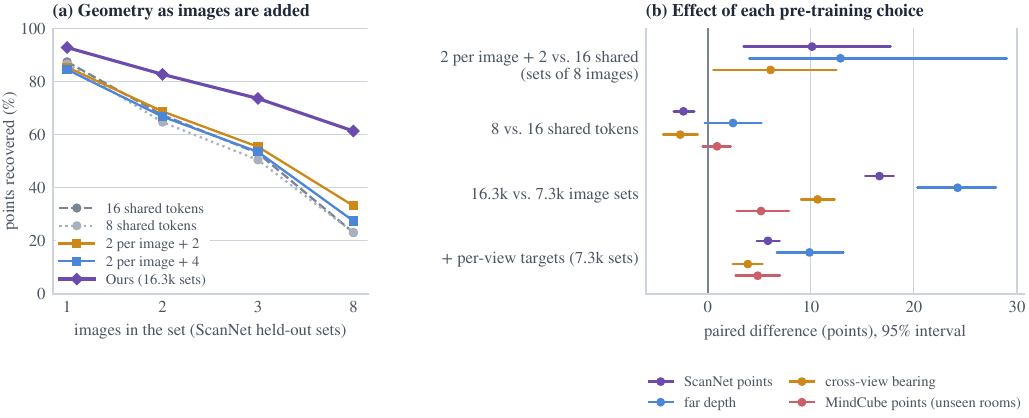}
\caption{Design choices in geometry pre-training, read from the \tokens{} by the refit readout on held-out image sets. (a) Visible points recovered in the first image's frame, by the number of images in a ScanNet set: tokens shared by all images lose the most as images are added. Ours is trained on 16.3k sets with per-view targets, the other layouts on 7.3k. (b) Paired difference of each choice on the same held-out sets, with 95\% intervals from resampling the sets. Metrics are defined in \cref{tab:app-stage-a}.}
\label{fig:app-stage-a}
\end{figure*}

\Cref{fig:app-stage-a} and \cref{tab:app-stage-a} compare the design choices of geometry pre-training by the geometry that the \tokens{} hold on held-out image sets, with differences paired over the same sets and 95\% intervals from resampling them:

\begin{itemize}
    \item \textbf{A separate geometry stage with a trainable backbone.} Geometry is learned in a stage of its own, before the reasoning chains, so that the tokens are fit to the images rather than shaped by the language loss. Pre-training also trains a LoRA on the backbone: with the backbone frozen the tokens cannot relate the views to each other, as the probe of \cref{sec:diagnosis} shows for the baselines.

    \item \textbf{A light readout that can only read the tokens.} The readout is one cross-attention layer without a residual path. Its queries encode only a grid position or derive from the scene tokens of a view, and its keys and values are the \tokens{} alone (\cref{app:readout}). Whatever it decodes must therefore be in the tokens, and a stronger readout would only make that less certain.

    \item \textbf{Tokens per image instead of a fixed set.} A fixed set of 16 tokens is shared by every image of a set, so each image gets fewer of them as images are added, and the number of images a model can take is bounded by what the tokens were trained on. With two tokens per image plus a camera group, every image has its own tokens, carrying the same two ids, and any number of images uses trained embeddings only. \Cref{fig:app-stage-a}a shows where sharing hurts most: on the sets of eight images, two tokens per image recover 10.11 more points than 16 shared tokens (95\% interval 3.47 to 17.73), and on the sets of three images, where they are 8 tokens in all, they still recover more (55.44\% against 53.02\%). It is sharing, not the count, that hurts: halving the shared tokens to 8 costs 2.36 points ($-3.31$ to $-1.32$). Four camera tokens instead of two leave the camera position and heading essentially unchanged and raise the cross-view bearing by 1.05 points.

    \item \textbf{More image sets.} \Cref{fig:app-stage-a}b shows that widening pre-training from 7.3k to 16.3k ScanNet and MindCube sets, without changing the reasoning data, is the largest single gain: 16.66 more points on held-out ScanNet sets (15.28 to 18.05), 34.52 more on the sets of eight images, and 5.16 more on MindCube rooms not seen in training.

    \item \textbf{Redundant per-view targets.} Besides each visual token's point in the first image's frame, the readout also predicts the same point in the frame of that token's own camera. The extra target is redundant, since the first follows from the second and the camera, but it gives the tokens a supervision of depth that does not depend on the camera being right. On 7.3k sets it raises the recovered points by 5.83 (4.75 to 6.97) and far depth by 9.87 points. In the final setting, both point targets are weighted by a learned confidence.

\end{itemize}

\begin{table*}[!ht]
\centering
\caption{Design choices in geometry pre-training, read from the \tokens{} by the refit readout on held-out image sets (\%). \emph{Local}: the per-view point targets, weighted by a learned confidence in the final setting. \emph{Points}: visible points recovered in the first image's frame (within a quarter of the set's mean point distance), on all 416 ScanNet sets and by the number of images in the set; \emph{far depth}: points beyond 1.5 mean distances whose depth is within 20\%; \emph{bearing}, \emph{facing}, \emph{camera}: the later images' relation to the first, their heading within $45^\circ$ and their position within a quarter mean distance (sets with more than one image). MindCube: points on unseen rooms. ScanNet-x: held-out sets of the wider ScanNet split, scored for the 16.3k runs only.}
\label{tab:app-stage-a}
\footnotesize
\setlength{\tabcolsep}{3.5pt}
\resizebox{\textwidth}{!}{\begin{tabular}{lllrrrr|rrrr|r|r}
\toprule
Tokens & Sets & Local & \multicolumn{4}{c}{ScanNet points} & Far & & & & MindCube & ScanNet-x \\ & & & all & 1 img. & 3 img. & 8 img. & depth & Bearing & Facing & Camera & points & points \\
\midrule
\multicolumn{13}{l}{\textit{Geometry trained on 7.3k image sets}} \\
16 (shared) & 7.3k & -- & 55.81 & 87.45 & 53.02 & 23.09 & 25.49 & 72.90 & 88.68 & 55.01 & 73.18 & -- \\
8 (shared) & 7.3k & -- & 53.45 & 86.64 & 50.46 & 23.14 & 27.93 & 70.23 & 88.54 & 49.71 & 74.07 & -- \\
2/img + 2 & 7.3k & -- & 58.23 & 85.45 & 55.44 & 33.20 & 29.40 & 75.54 & 87.39 & 57.02 & 75.41 & -- \\
2/img + 4 & 7.3k & -- & 56.14 & 84.55 & 53.51 & 27.47 & 29.54 & 73.79 & 88.25 & 56.30 & 73.59 & -- \\
4/img + 4 & 7.3k & -- & 58.80 & 86.92 & 56.69 & 29.39 & 31.42 & 74.72 & 88.54 & 61.60 & 74.83 & -- \\
2/img + 4 & 7.3k & points & 61.97 & 87.47 & 59.62 & 36.15 & 39.40 & 77.67 & 89.40 & 56.45 & 78.43 & -- \\
\midrule
\multicolumn{13}{l}{\textit{Geometry trained on 16.3k image sets}} \\
2/img + 2 & 16.3k & -- & 72.19 & 90.75 & 69.94 & 61.58 & 52.09 & 83.40 & 94.13 & 74.79 & 78.67 & 50.28 \\
2/img + 4 & 16.3k & -- & 72.79 & 91.61 & 70.48 & 61.99 & 53.78 & 84.44 & 93.70 & 74.64 & 78.76 & 51.38 \\
\textbf{Ours:} 2/img + 4 & 16.3k & points + conf. & 75.63 & 92.84 & 73.63 & 61.37 & 62.75 & 85.26 & 93.55 & 74.36 & 81.62 & 53.37 \\
\bottomrule
\end{tabular}}
\end{table*}

\FloatBarrier
\subsection{Design Choices in SFT}
\label{app:design-sft}

\begin{figure*}[!ht]
\centering
\includegraphics[width=\textwidth]{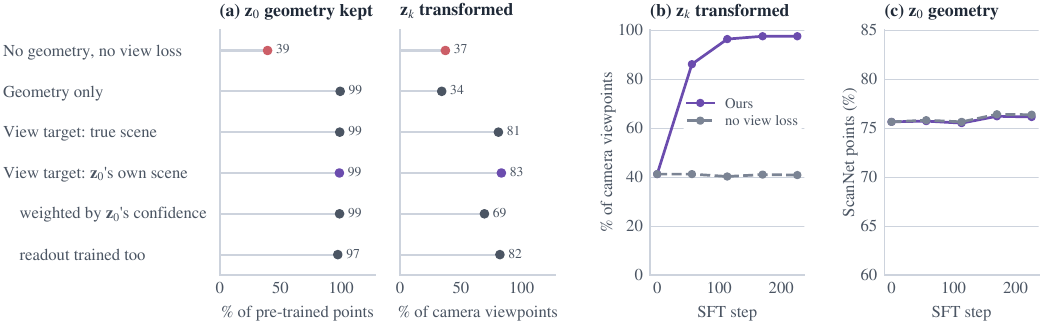}
\caption{Design choices in SFT. (a) Supervision variants in the preliminary setting: the share of the pre-trained ScanNet points that $\mathbf{z}_0$ still holds after SFT, and how often a re-emitted $\mathbf{z}_k$ after a camera's viewpoint decodes to $\mathbf{z}_0$'s scene seen from that camera. The indented rows change one thing in the row ``$\mathbf{z}_0$'s own scene''. Red marks the run without supervision and violet the target we keep. (b, c) The final setting along SFT, with and without the view loss: the view loss transforms $\mathbf{z}_k$ within the first half of training and leaves $\mathbf{z}_0$'s geometry where it was.}
\label{fig:app-stage-b}
\end{figure*}

\Cref{fig:app-stage-b} and \cref{tab:app-stage-b} read two things after SFT: whether $\mathbf{z}_0$ still holds the geometry of pre-training, and how often a re-emitted $\mathbf{z}_k$ after a camera's viewpoint decodes to $\mathbf{z}_0$'s scene seen from that camera, i.e., how often it is transformed.

\paragraph{Supervision of the re-emitted tokens.}
We first compare the supervision variants in a preliminary setting (4,800 chains, 16 shared tokens, one epoch).
Without any geometry or view loss, SFT overwrites the geometry: the ScanNet points read from $\mathbf{z}_0$ fall from 55.81\% to 22.00\%.
The geometry loss alone keeps the geometry but leaves $\mathbf{z}_k$ untransformed (34.24\%).
Supervising $\mathbf{z}_k$ with $\mathbf{z}_0$'s own scene transforms it as well as supervising it with the true scene (83.33\% against 80.91\%), and with $\mathbf{z}_0$ supplying the scene, the loss supervises only how that scene transforms.
Weighting that target by $\mathbf{z}_0$'s confidence transforms it less (69.39\%), and training the readout during SFT adds nothing.

\paragraph{The final setting.}
With two tokens per image, two epochs, and $\lambda{=}0.1$, the geometry even improves during SFT (75.63\% to 76.13\%).
The view loss is what transforms it: $\mathbf{z}_k$ is transformed in 97.49\% of the camera viewpoints with it and in only 40.81\% without it.

\begin{table*}[!ht]
\centering
\caption{Design choices in SFT, on held-out data. $\mathbf{z}_0$ points: the held-out geometry of pre-training read from $\mathbf{z}_0$ by the frozen readout before SFT and after it (\%), where a falling number means SFT overwrote the geometry. $\mathbf{z}_k$ transformed: the share of re-emitted runs after a camera's viewpoint whose decoded scene is closest to $\mathbf{z}_0$'s scene seen from that camera (chance is one over the number of images).}
\label{tab:app-stage-b}
\footnotesize
\setlength{\tabcolsep}{3.5pt}
\begin{adjustbox}{max width=\textwidth}
\begin{tabular}{llrrrrr}
\toprule
Variant & View loss & \multicolumn{2}{c}{$\mathbf{z}_0$ points ScanNet} & \multicolumn{2}{c}{$\mathbf{z}_0$ points MindCube} & $\mathbf{z}_k$ transformed \\ & & before & after & before & after & \\
\midrule
\multicolumn{7}{l}{\textit{Preliminary setting: 4,800 chains, 16 shared tokens, 1 epoch}} \\
No geometry, no view loss & -- & 55.81 & 22.00 & 73.54 & 40.45 & 37.27 \\
Geometry only ($\mathbf{z}_k$ free) & -- & 55.81 & 55.35 & 73.54 & 74.36 & 34.24 \\
View target: true scene & \checkmark & 55.81 & 55.15 & 73.54 & 73.97 & 80.91 \\
View target: $\mathbf{z}_0$'s own scene & \checkmark & 55.81 & 54.99 & 73.54 & 74.29 & 83.33 \\
\quad weighted by $\mathbf{z}_0$'s confidence & \checkmark & 55.81 & 55.06 & 73.54 & 74.22 & 69.39 \\
\quad readout trained too & \checkmark & 55.81 & 54.25 & 73.54 & 74.00 & 82.12 \\
\midrule
\multicolumn{7}{l}{\textit{Final setting: 2 tokens per image, 2 epochs}} \\
\textbf{Ours} & \checkmark & 75.63 & 76.13 & 82.51 & 83.26 & 97.49 \\
\quad without the view loss ($\mathbf{z}_k$ free) & -- & 75.63 & 76.36 & 82.51 & 83.63 & 40.81 \\
\bottomrule
\end{tabular}
\end{adjustbox}
\end{table*}

\FloatBarrier
\subsection{Geometry along Training}
\label{app:dynamics}

\begin{figure*}[!ht]
\centering
\includegraphics[width=\textwidth]{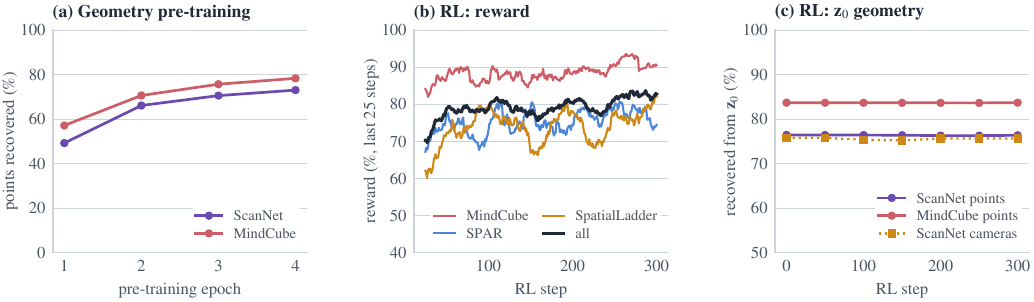}
\caption{The geometry through training. (a) Geometry pre-training, epoch by epoch (readout as trained, before the refit): the points recovered on held-out ScanNet and MindCube sets. (b) Reward by question source, averaged over the last 25 steps, along an RL run from SFT. (c) The geometry read from $\mathbf{z}_0$ along the same run, which has no geometry loss: points on held-out ScanNet and MindCube sets and the cameras' positions stay unchanged.}
\label{fig:app-dynamics}
\end{figure*}

\Cref{fig:app-dynamics} follows the geometry held by the \tokens{} through the three training stages.

\paragraph{Four epochs of geometry pre-training.}
In \cref{fig:app-dynamics}a, the points recovered on held-out ScanNet sets rise from 49.22\% after one epoch to 73.02\% after four, the last still adding 2.46 points.

\paragraph{No geometry loss in RL.}
RL optimizes the answer's reward only.
\Cref{fig:app-dynamics}b and c follow an RL run from SFT in which the geometry was read along training: the training reward rises from 70.28\% to 82.75\% over 300 steps (first and last 25 steps), while the geometry read from $\mathbf{z}_0$ does not move, with the recovered ScanNet points changing by $-0.07$ points, the MindCube points by less than 0.01, and the cameras' positions by $-0.14$.
Under the KL penalty to the SFT model and the low learning rate, the geometry stays in place without any geometry term.
The RL model of \cref{tab:main} behaves alike: on the held-out sets, its $\mathbf{z}_0$ holds nearly as many of the ScanNet points as after SFT, and its re-emitted $\mathbf{z}_k$ are transformed nearly as often.

\FloatBarrier
\section{A Benchmark for Spatial Capability beyond Answer Accuracy}
\label{app:capability-bench}
\begin{figure}[!ht]
\centering
\includegraphics[width=\linewidth]{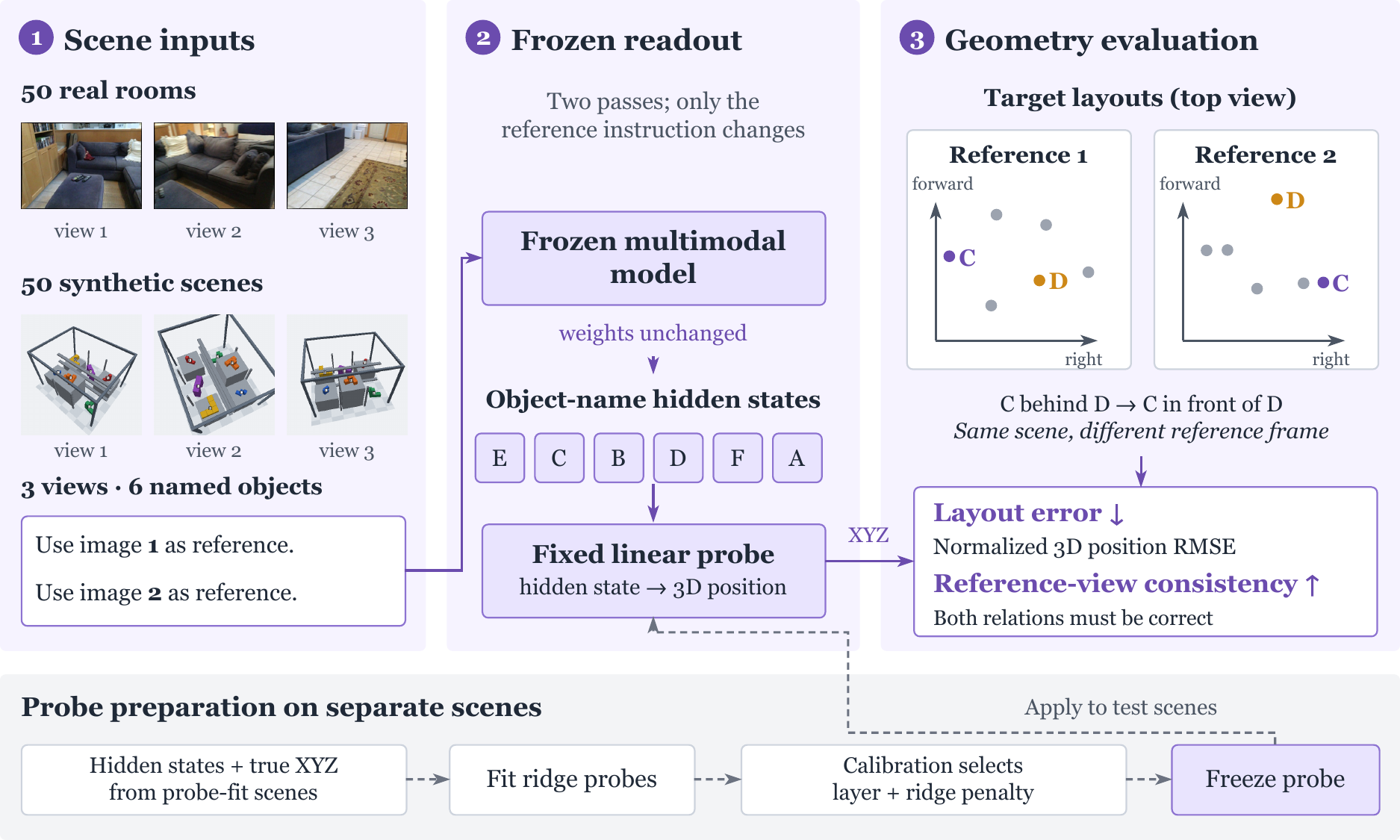}
\caption{Protocol of the benchmark. The same three views and six object names are presented under two reference-view instructions; a frozen model produces the states of the object names, and a fixed linear probe decodes their camera-relative positions. The top views show the true layouts of one synthetic scene, in which objects C and D reverse their front--back order between the two references. The probe is fitted and selected on scenes disjoint from the test set.}
\label{fig:capability-bench-protocol}
\end{figure}

We construct a compact benchmark that measures, independently of the answers a model generates, how accurately object positions can be read from its representations and whether they follow a change of reference view.
\Cref{fig:capability-bench-protocol} summarizes the protocol: the model stays frozen, and a linear probe reads the position of each object from the hidden state of its name.

\paragraph{Scenes.}
The test set contains 100 scenes, 50 real rooms and 50 synthetic scenes, each with six objects and three views.
The real rooms use the RGB views released with SpatialLadder~\citep{spatialladder}, with object centers and camera frames from the ScanNet~\citep{scannet} annotations.
The synthetic scenes hold compound objects on platforms and shelves, with partitions, height differences, and partial occlusion, and their geometry and cameras are recorded by the renderer.
Objects are identified by name in the real rooms and by the labels A--F in the synthetic scenes.

\paragraph{Paired inputs.}
The central comparison presents the same ordered images and object list twice, changing only the instruction from ``Use image 1 as the reference view'' to ``Use image 2 as the reference view'', and examines the object pairs whose front--back relation reverses between the two camera frames.
Further pairs keep the geometry and reorder the images or the object mentions; synthetic pairs also recolor the objects, relabel them, or move one of them.

\paragraph{Readout.}
Following the object-token readout of S-Space~\citep{sspace}, we take the hidden state at the last subword of each object name and fit a linear ridge probe to the object's 3D position in the right--up--forward frame of the reference camera:
\begin{equation}
    \widehat{\mathbf p}^{(r)}_i = W\,\widetilde{\mathbf h}^{(r)}_{\ell,i} + \mathbf b,
    \label{eq:bench-readout}
\end{equation}
where $r$ is the reference view and $\widetilde{\mathbf h}$ is the state at layer $\ell$, standardized on the probe-fit set.
One probe is fitted per model and domain on 96 real rooms or 300 synthetic scenes; a separate set of 20 real rooms or 50 synthetic scenes selects the layer among $3, 7, \dots, 31$ and the ridge penalty among $0.1, 1, 10, 100$ by mean absolute coordinate error, and the selected probe is frozen for the test scenes.
Physical rooms, including rescans, are disjoint across probe fitting, calibration, and testing.
All models receive the same images and instruction, and the input of \method{} retains its initial \tokens{}.

\paragraph{Metrics.}
\emph{Layout error} compares the decoded configuration with the true one in the first reference frame.
Each configuration is centered and divided by its root-mean-square object spread, and for the normalized configurations $\widehat P$ and $P$ of $n$ objects,
\begin{equation}
    E_{\mathrm{layout}} = \sqrt{\frac{1}{n}\sum_{i=1}^{n}
        \left\|\widehat{\mathbf p}_i-\mathbf p_i\right\|_2^2}.
    \label{eq:bench-layout}
\end{equation}
\emph{Reference-view consistency} (RVC) is the share of object pairs whose front--back relation is decoded correctly in both reference frames, among the pairs whose true relation reverses and whose coordinate gap exceeds $0.08$ of the normalized spread in both frames.
Scores are averaged over the eligible pairs of a scene and then over scenes.

\paragraph{Models and controls.}
We evaluate five Qwen3.5-4B models: Base, SFT, GRPO, and \method{} after SFT and after RL.
The SFT baseline is an independent reproduction of the recipe in \cref{app:training} (LoRA rank 64, 7,500 teacher-written chains, two epochs).
Two controls check the readout: blank images under the fixed probe, and a probe refitted on labels permuted across training scenes.

\paragraph{The readout follows the reference view.}
\Cref{tab:capability-bench-main} reports the results on the 100 test scenes.
SFT raises RVC over Base from 19.72\% to 26.64\% on real rooms and from 18.76\% to 30.89\% on synthetic scenes, whereas \method{} reaches 35.18\% and 44.38\% after SFT and 38.10\% and 47.26\% after RL, 11.46 and 16.37 points above SFT.
Its layout error is lower as well, 0.839 against 0.959 for SFT on real rooms and 0.791 against 0.817 on synthetic scenes.
The readout also rests on the images and on the learned correspondence between states and geometry: for the RL model, blank images and shuffled labels each bring RVC below every model in \cref{tab:capability-bench-main} on both domains.
The advantage of \method{} thus extends from its answers to the states of ordinary object tokens under a neutral instruction.

\begin{table}[!ht]
\centering
\caption{Spatial representations on the 100 test scenes, 50 per domain. $E_{\mathrm{layout}}$: the layout error of the normalized configuration; RVC: reference-view consistency (\%), which requires the front--back relation to be correct in both reference frames. Bold marks the best model in each column.}
\label{tab:capability-bench-main}
\footnotesize
\setlength{\tabcolsep}{9pt}
\begin{tabular}{lrrrr}
\toprule
 & \multicolumn{2}{c}{Real rooms} & \multicolumn{2}{c}{Synthetic scenes} \\
\cmidrule(lr){2-3}\cmidrule(lr){4-5}
Model & $E_{\mathrm{layout}}\!\downarrow$ & RVC $\uparrow$ & $E_{\mathrm{layout}}\!\downarrow$ & RVC $\uparrow$ \\
\midrule
Base & 1.001 & 19.72 & 0.839 & 18.76 \\
+ SFT & 0.959 & 26.64 & 0.817 & 30.89 \\
+ GRPO & 0.949 & 24.97 & 0.838 & 22.12 \\
Ours (SFT) & 0.844 & 35.18 & 0.800 & 44.38 \\
Ours (RL) & \textbf{0.839} & \textbf{38.10} & \textbf{0.791} & \textbf{47.26} \\
\bottomrule
\end{tabular}
\end{table}

\paragraph{The relations hold under input and scene changes.}
\Cref{tab:capability-bench-interventions} scores paired relations under changes to the input and to the scene; a relation counts when it is decoded correctly in both observations, averaged over the eligible relations on the three axes of a scene.
Both models of \method{} exceed all three baselines in every column, most clearly on the relations that an object move changes (29.20\% and 27.31\% against at most 25.13\%).

\begin{table}[!ht]
\centering
\caption{Paired spatial relations under input and scene changes (\%). Real rooms: the images or the object mentions are reordered under the same reference camera. Synthetic scenes: the objects are recolored or relabeled A--F, or one object is moved, with the relations it changes and those between the unmoved objects scored separately. Bold marks the best model in each column.}
\label{tab:capability-bench-interventions}
\footnotesize
\setlength{\tabcolsep}{5pt}
\begin{tabular}{lrrrrrr}
\toprule
 & \multicolumn{2}{c}{Real rooms} & \multicolumn{4}{c}{Synthetic scenes} \\
\cmidrule(lr){2-3}\cmidrule(lr){4-7}
Model & Images & Mentions & Colors & Labels & Moved & Unmoved \\
\midrule
Base & 63.70 & 66.22 & 64.91 & 56.18 & 24.42 & 69.17 \\
+ SFT & 65.05 & 68.31 & 66.18 & 57.98 & 25.13 & 70.59 \\
+ GRPO & 64.64 & 69.94 & 67.03 & 54.22 & 24.08 & 70.15 \\
Ours (SFT) & \textbf{67.01} & 72.11 & 67.76 & \textbf{60.20} & \textbf{29.20} & 72.07 \\
Ours (RL) & 66.61 & \textbf{72.15} & \textbf{68.86} & 59.90 & 27.31 & \textbf{72.28} \\
\bottomrule
\end{tabular}
\end{table}

\FloatBarrier
\section{\method{} as a Spatial Expert for General Models}
\label{app:spatial-expert}
\begin{figure}[!ht]
\centering
\includegraphics[width=\linewidth]{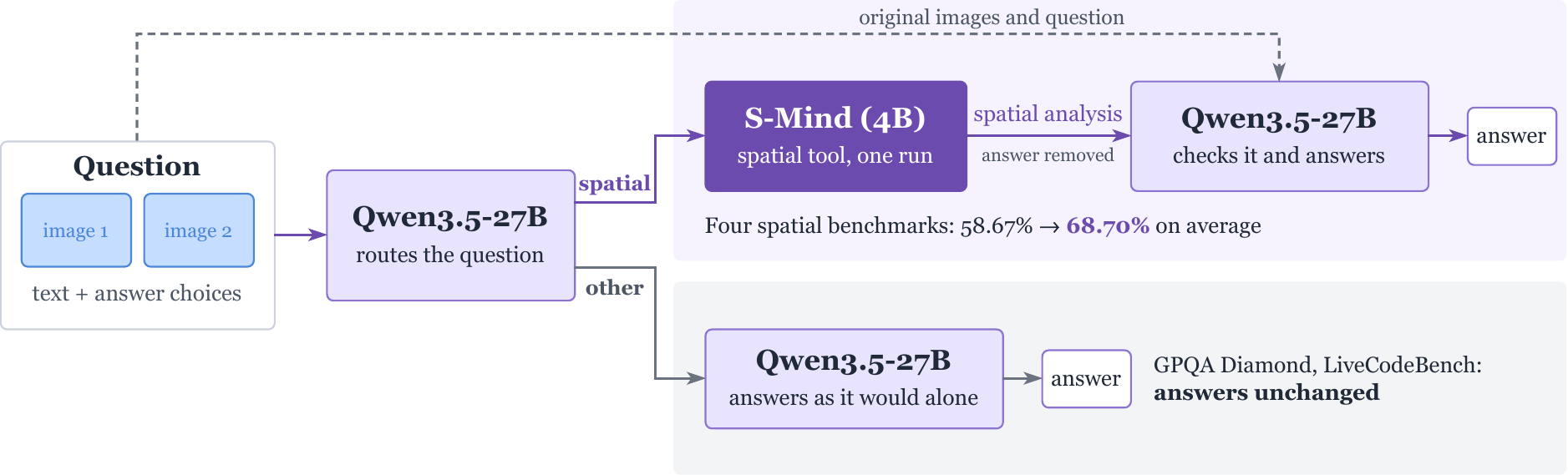}
\caption{\method{} as a spatial tool for a general model. Qwen3.5-27B routes each question: a spatial question goes to \method{}, whose analysis returns to the 27B model with its answer selection removed, and the 27B model answers from it and from the original inputs; any other question it answers as it would alone.}
\label{fig:expert-pipeline}
\end{figure}

We further examine whether \method{} can serve a larger general model through language: as \cref{fig:expert-pipeline} outlines, the 4B RL model acts as a spatial tool for Qwen3.5-27B, with both models frozen.

\paragraph{Pipeline.}
Qwen3.5-27B first routes the question in a separate pass, given its text, answer choices, and number of images.
For a visual spatial question, \method{} receives the question, the choices, and the ordered images, runs once, and returns its spatial analysis with the final answer selection removed.
The 27B model then reads this analysis alongside the original inputs, checks the reference viewpoint and the object identities, and produces the answer itself; if the analysis is unusable, it answers from the original inputs alone.
Any other question is answered by the 27B model with its original prompt and decoding settings, outside the routing conversation.

\paragraph{The spatial analysis transfers to the larger model.}
We compare Qwen3.5-27B alone and with \method{} on 64 questions from each of the four benchmarks, sampled by task before inference and aggregated as in \cref{app:benchmarks}, with thinking disabled, greedy decoding, and at most 4{,}096 new tokens.
The router calls the tool on all 256 questions.
As \cref{tab:expert-spatial} shows, the average rises from 58.67\% to 68.70\%, with gains on all four benchmarks and the largest on MindCube-Tiny, from 59.59\% to 81.30\%, where reasoning across views is central.

\begin{table}[!ht]
\centering
\caption{\method{} as a spatial tool for Qwen3.5-27B: accuracy (\%) on 64 questions per benchmark. The 27B model produces every final answer.}
\label{tab:expert-spatial}
\footnotesize
\setlength{\tabcolsep}{6pt}
\begin{tabular}{lrrrrr}
\toprule
 & MindCube-Tiny & SPAR-Bench & BLINK & ViewSpatial & Avg. \\
\midrule
Qwen3.5-27B & 59.59 & 50.30 & 70.12 & 54.66 & 58.67 \\
\quad +\,\method{} analysis & \textbf{81.30} & \textbf{59.09} & \textbf{73.26} & \textbf{61.14} & \textbf{68.70} \\
\bottomrule
\end{tabular}
\end{table}

\paragraph{General tasks are unaffected.}
On the 198 questions of GPQA Diamond\footnote{\url{https://huggingface.co/datasets/Idavidrein/gpqa}} and the 175 code-generation problems of the incremental v6 split of LiveCodeBench\footnote{\url{https://github.com/LiveCodeBench/LiveCodeBench}}, the router sends every question to the 27B model itself.
Its answers, generated independently with thinking disabled at temperature zero, match those of the standalone model question by question on both benchmarks.

\FloatBarrier
\section{Case Study}
\label{app:qualitative}
\definecolor{camC1}{HTML}{F2B75B}\definecolor{camC2}{HTML}{62C6E3}\definecolor{camC3}{HTML}{EE86A8}\definecolor{camC4}{HTML}{9FD27C}
\newcommand{\casechip}[2]{\tikz[baseline=(c.base)]\node[fill=#1!40, draw=#1!80!black, rounded corners=1.5pt, inner xsep=1.6pt, inner ysep=1pt, font=\scriptsize](c){#2};}
\newcommand{\casecap}[1]{{\scriptsize #1\par}\vspace{2pt}}
\newcommand{\casenote}[1]{\par\vspace{1pt}{\scriptsize #1\par}}
\newlength{\casew}
\newcommand{\casepanel}[3]{\setlength{\casew}{#1}\begin{tikzpicture}\node[anchor=south west, inner sep=0](p){\includegraphics[width=\casew]{#2}};\begin{scope}[x={(p.south east)}, y={(p.north west)}]#3\end{scope}\end{tikzpicture}}
\newcommand{\caselab}[5]{\fill[#4, draw=white, line width=0.4pt] (#1,#2) circle (1.2pt); \node[anchor=#3, outer sep=1.6pt, inner xsep=1.4pt, inner ysep=0.9pt, font=\tiny, text=white, fill=black, fill opacity=0.72, text opacity=1, draw=#4, draw opacity=1, line width=0.5pt, rounded corners=1.2pt] at (#1,#2) {#5};}

\begin{figure*}[!ht]
\centering
\footnotesize
\definecolor{caseO1}{HTML}{8D73D1}\definecolor{caseO2}{HTML}{C47F10}\definecolor{caseO3}{HTML}{239C85}
\begin{minipage}[t]{0.58\linewidth}\vspace{0pt}
\includegraphics[width=\linewidth]{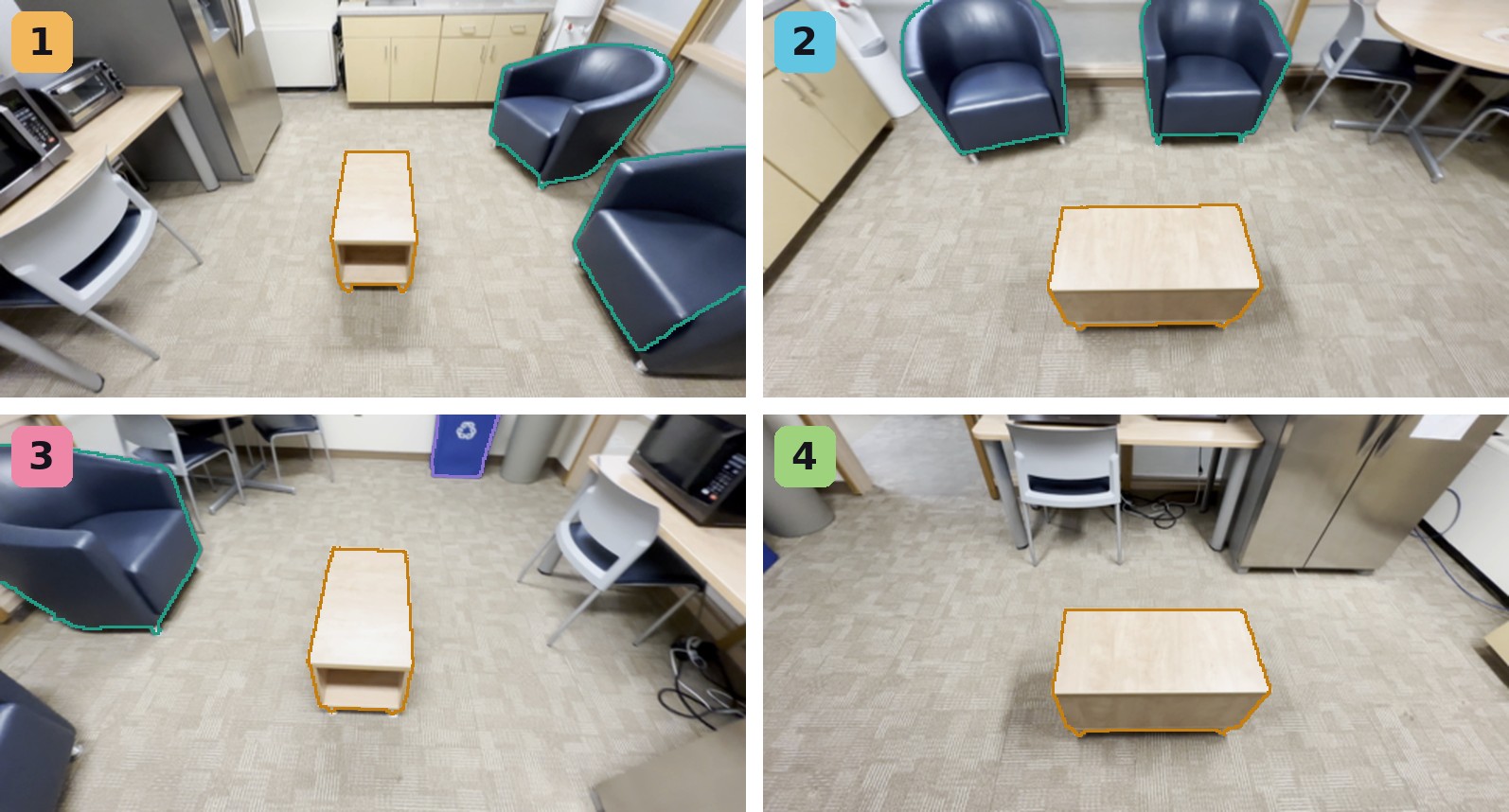}
\end{minipage}\hfill
\begin{minipage}[t]{0.395\linewidth}\vspace{0pt}\raggedright
\textbf{Question.} Based on these four images (image 1, 2, 3, and 4) showing the coffee table from different viewpoints (front, left, back, and right), with each camera aligned with room walls and partially capturing the surroundings: From the viewpoint presented in image 2, what is to the right of the coffee table?\par\smallskip
(A) Cream-colored cabinet\\
(B) Table and white chair\\
(C) Two blue leather armchairs\\
\textbf{(D) Blue bin}
\end{minipage}\par\smallskip
{\raggedright\textbf{Answers.} \method{}: \textbf{D}\,\ding{51} \quad Qwen3.5-4B: B\,\ding{55} \quad +\,SFT: B\,\ding{55} \quad +\,GRPO: B\,\ding{55}\par}\smallskip
{\raggedright\textbf{\method{}'s reasoning.} From the viewpoint of image 3, \casechip{camC3}{$\mathbf{z}_1$} I see the blue bin beyond the coffee table near the table and white chair.
From the viewpoint of image 2, \casechip{camC2}{$\mathbf{z}_2$} I face the two blue leather armchairs with the coffee table ahead, and the blue bin side of the room is to my right.
So the blue bin is to the right of the coffee table.\par}\medskip
\begin{minipage}[t]{0.325\linewidth}\vspace{0pt}\centering
\casecap{$\mathbf{z}_0$: all four views fused}
\casepanel{\linewidth}{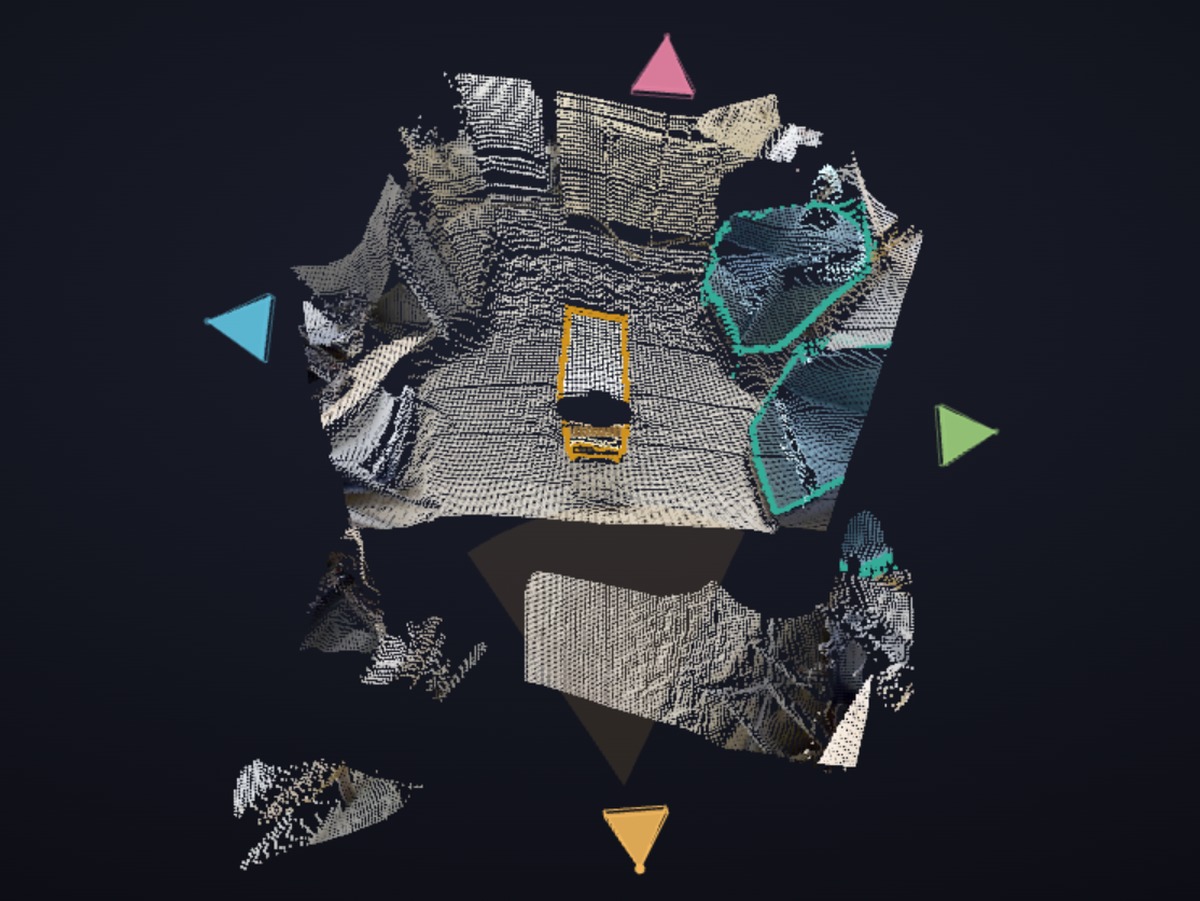}{\caselab{0.430}{0.209}{south}{caseO1}{blue bin}\caselab{0.458}{0.512}{north east}{caseO2}{coffee table}\caselab{0.664}{0.762}{south}{caseO3}{armchairs}}
\casenote{headings $+89^\circ$\,\ding{51}, $-176^\circ$\,\ding{51}, $-88^\circ$\,\ding{51}}
\end{minipage}\hfill\begin{minipage}[t]{0.325\linewidth}\vspace{0pt}\centering
\casecap{$\mathbf{z}_1$: from image 3's viewpoint}
\casepanel{\linewidth}{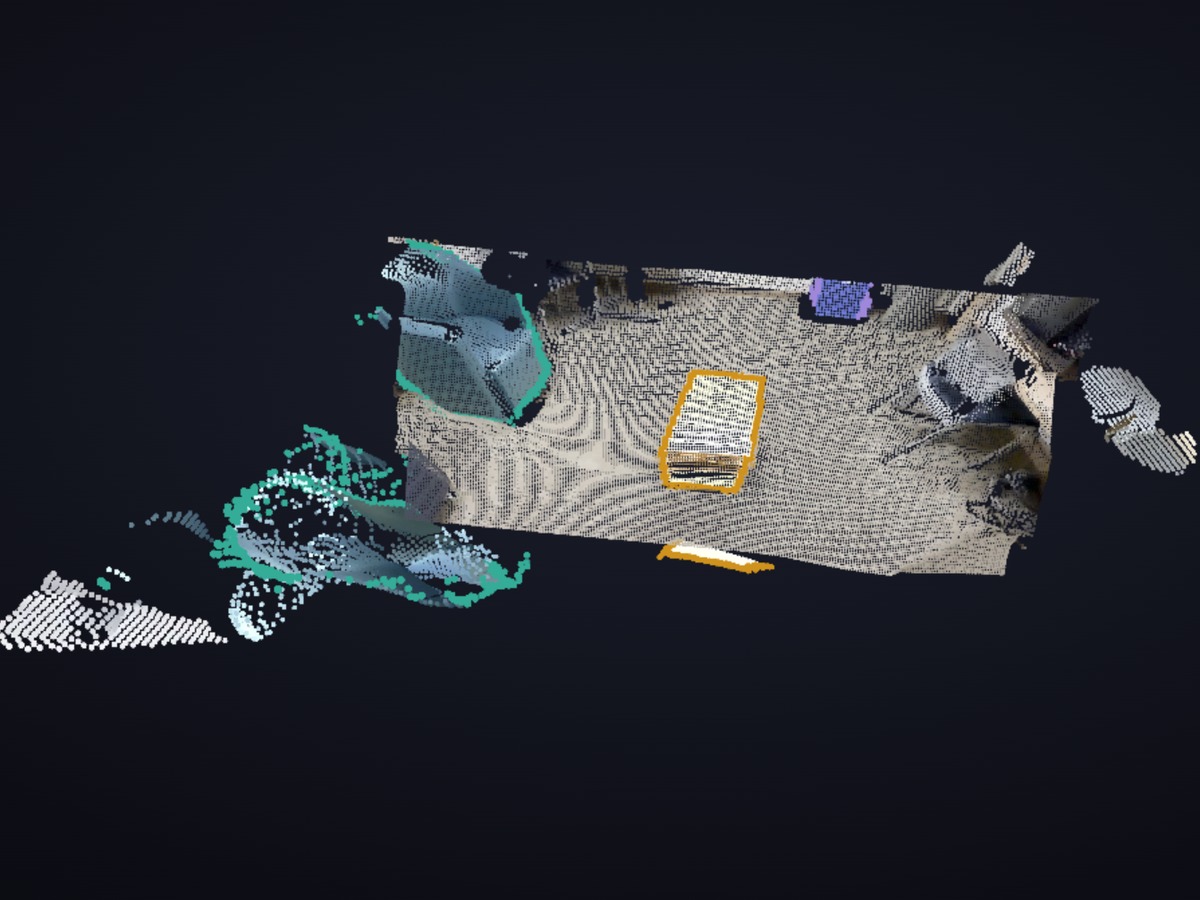}{\caselab{0.668}{0.691}{south}{caseO1}{blue bin}\caselab{0.583}{0.596}{south}{caseO2}{coffee table}\caselab{0.384}{0.739}{south}{caseO3}{armchairs}}
\casenote{match error 0.16 (others $\geq$0.82)}
\end{minipage}\hfill\begin{minipage}[t]{0.325\linewidth}\vspace{0pt}\centering
\casecap{$\mathbf{z}_2$: from image 2's viewpoint}
\casepanel{\linewidth}{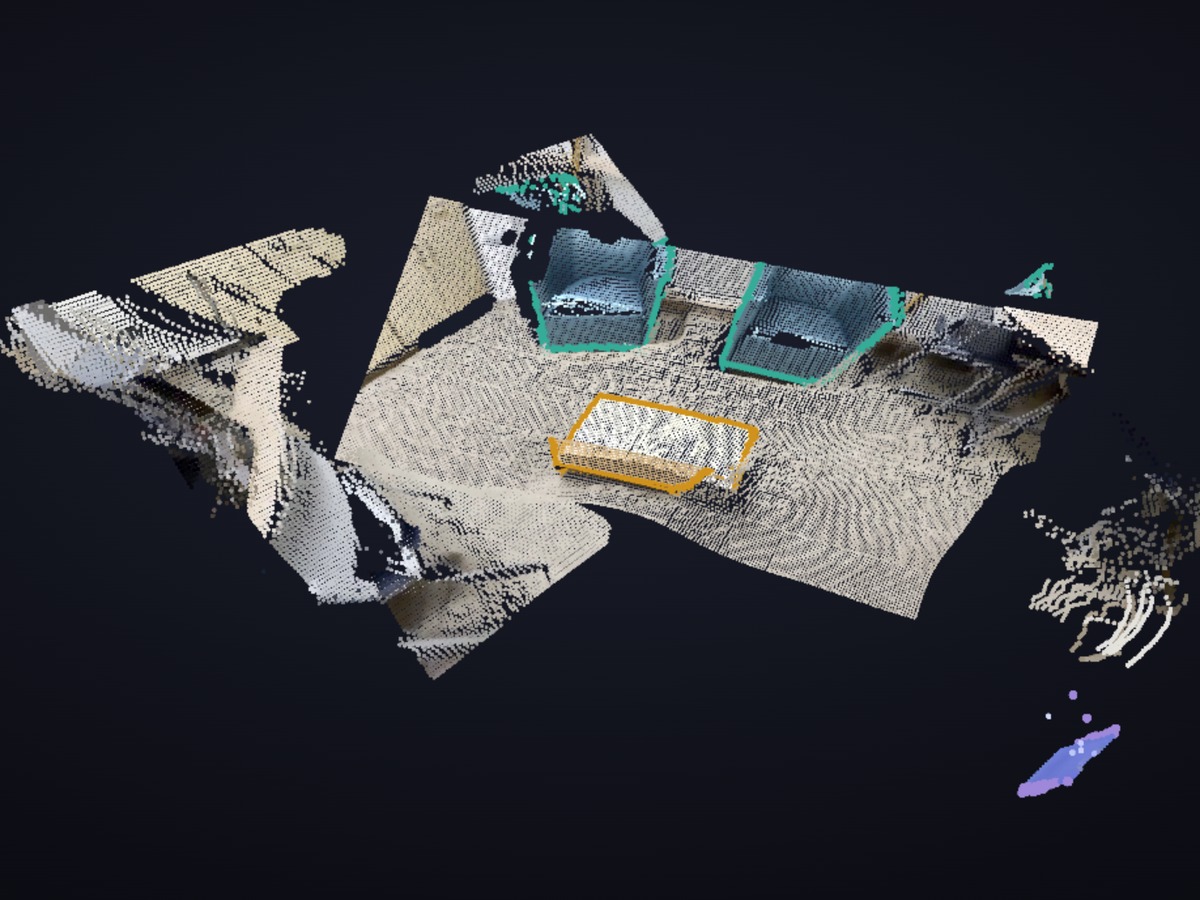}{\caselab{0.899}{0.207}{south}{caseO1}{blue bin}\caselab{0.532}{0.466}{north}{caseO2}{coffee table}\caselab{0.588}{0.747}{south}{caseO3}{armchairs}}
\casenote{match error 0.23 (others $\geq$0.77)}
\end{minipage}\par\medskip
\begin{minipage}[t]{0.325\linewidth}\vspace{0pt}\centering
\casecap{Qwen3.5-4B: $\mathbf{z}_0$ from the probe}
\casepanel{\linewidth}{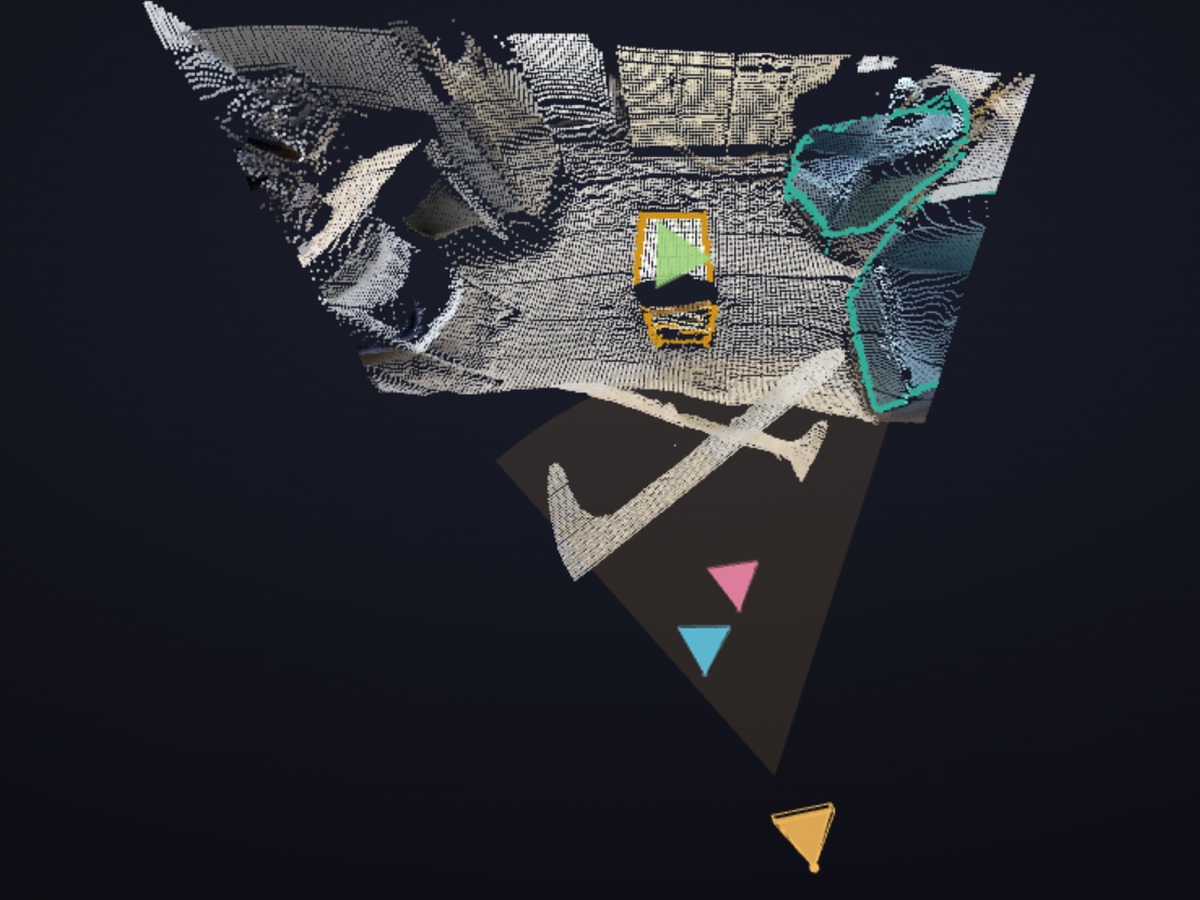}{\caselab{0.521}{0.651}{north east}{caseO2}{coffee table}\caselab{0.742}{0.864}{south}{caseO3}{armchairs}}
\casenote{headings $+12^\circ$\,\ding{55}, $-53^\circ$\,\ding{55}, $-73^\circ$\,\ding{51}}
\end{minipage}\hfill\begin{minipage}[t]{0.325\linewidth}\vspace{0pt}\centering
\casecap{+\,SFT: $\mathbf{z}_0$ from the probe}
\casepanel{\linewidth}{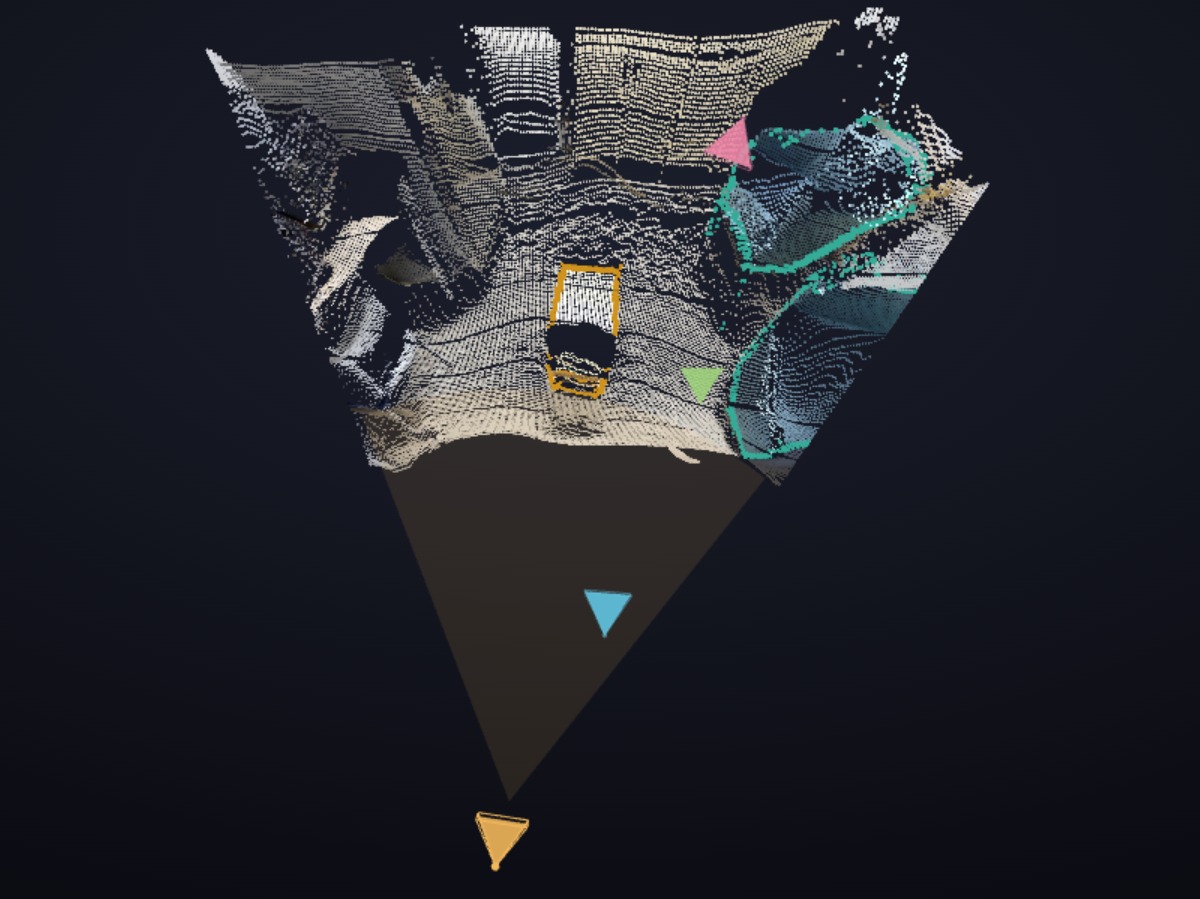}{\caselab{0.484}{0.688}{south}{caseO2}{coffee table}\caselab{0.686}{0.527}{north}{caseO3}{armchairs}}
\casenote{headings $+75^\circ$\,\ding{51}, $-119^\circ$\,\ding{55}, $-87^\circ$\,\ding{51}}
\end{minipage}\hfill\begin{minipage}[t]{0.325\linewidth}\vspace{0pt}\centering
\casecap{+\,GRPO: $\mathbf{z}_0$ from the probe}
\casepanel{\linewidth}{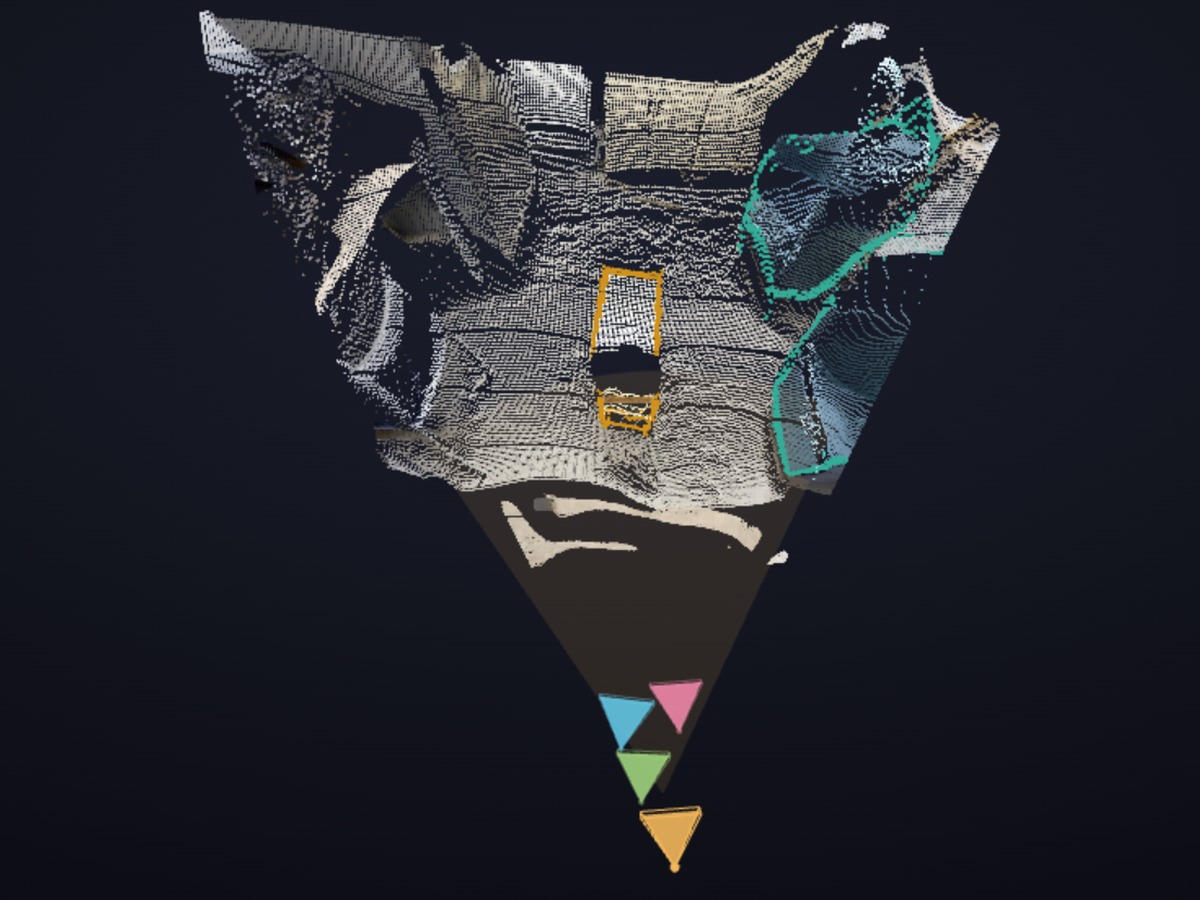}{\caselab{0.487}{0.616}{east}{caseO2}{coffee table}\caselab{0.699}{0.846}{south}{caseO3}{armchairs}}
\casenote{headings $+54^\circ$\,\ding{51}, $+50^\circ$\,\ding{55}, $-66^\circ$\,\ding{51}}
\end{minipage}
\caption{\textbf{Four views around a coffee table.} The question asks what lies to the right of the coffee table as seen from image 2. Top: \method{}'s $\mathbf{z}_0$ and its scenes at the runs it re-emits, decoded by its frozen readout; bottom: the $\mathbf{z}_0$ that the geometry probe reads from each baseline. Each $\mathbf{z}_0$ is seen from straight above, each camera as a wedge in its photo's color opening the way it faces and image 1's field of view shaded; each $\mathbf{z}_k$ from behind its viewer.}
\label{fig:case-coffee}
\end{figure*}

\begin{figure*}[!ht]
\centering
\footnotesize
\definecolor{caseO1}{HTML}{8D73D1}\definecolor{caseO2}{HTML}{C47F10}\definecolor{caseO3}{HTML}{239C85}
\begin{minipage}[t]{0.58\linewidth}\vspace{0pt}
\includegraphics[width=\linewidth]{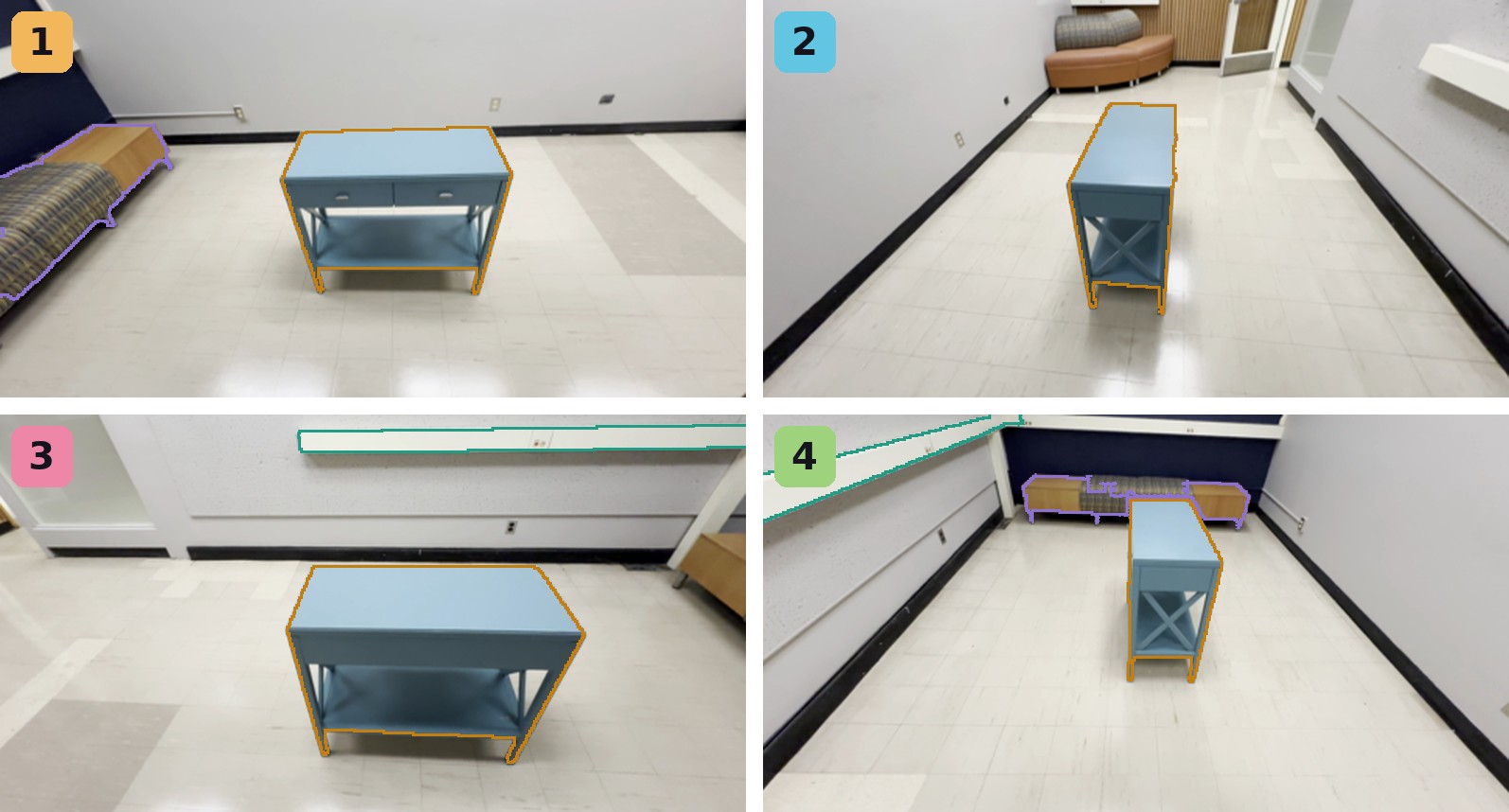}
\end{minipage}\hfill
\begin{minipage}[t]{0.395\linewidth}\vspace{0pt}\raggedright
\textbf{Question.} Based on these four images (image 1, 2, 3, and 4) showing the blue table from different viewpoints (front, left, back, and right), with each camera aligned with room walls and partially capturing the surroundings: From the viewpoint presented in image 3, what is to the right of the blue table?\par\smallskip
(A) Plain white wall\\
(B) Sofa\\
(C) Wall with an air conditioning unit\\
\textbf{(D) Seating bench}
\end{minipage}\par\smallskip
{\raggedright\textbf{Answers.} \method{}: \textbf{D}\,\ding{51} \quad Qwen3.5-4B: A\,\ding{55} \quad +\,SFT: B\,\ding{55} \quad +\,GRPO: B\,\ding{55}\par}\smallskip
{\raggedright\textbf{\method{}'s reasoning.} From the viewpoint of image 4, \casechip{camC4}{$\mathbf{z}_1$} I see the seating bench beyond the blue table against the far wall.
From the viewpoint of image 3, \casechip{camC3}{$\mathbf{z}_2$} I face the table with the wall with an air conditioning unit ahead, and the seating bench lies to my right toward the far side of the room.
So the seating bench is to the right of the blue table.\par}\medskip
\begin{minipage}[t]{0.325\linewidth}\vspace{0pt}\centering
\casecap{$\mathbf{z}_0$: all four views fused}
\casepanel{\linewidth}{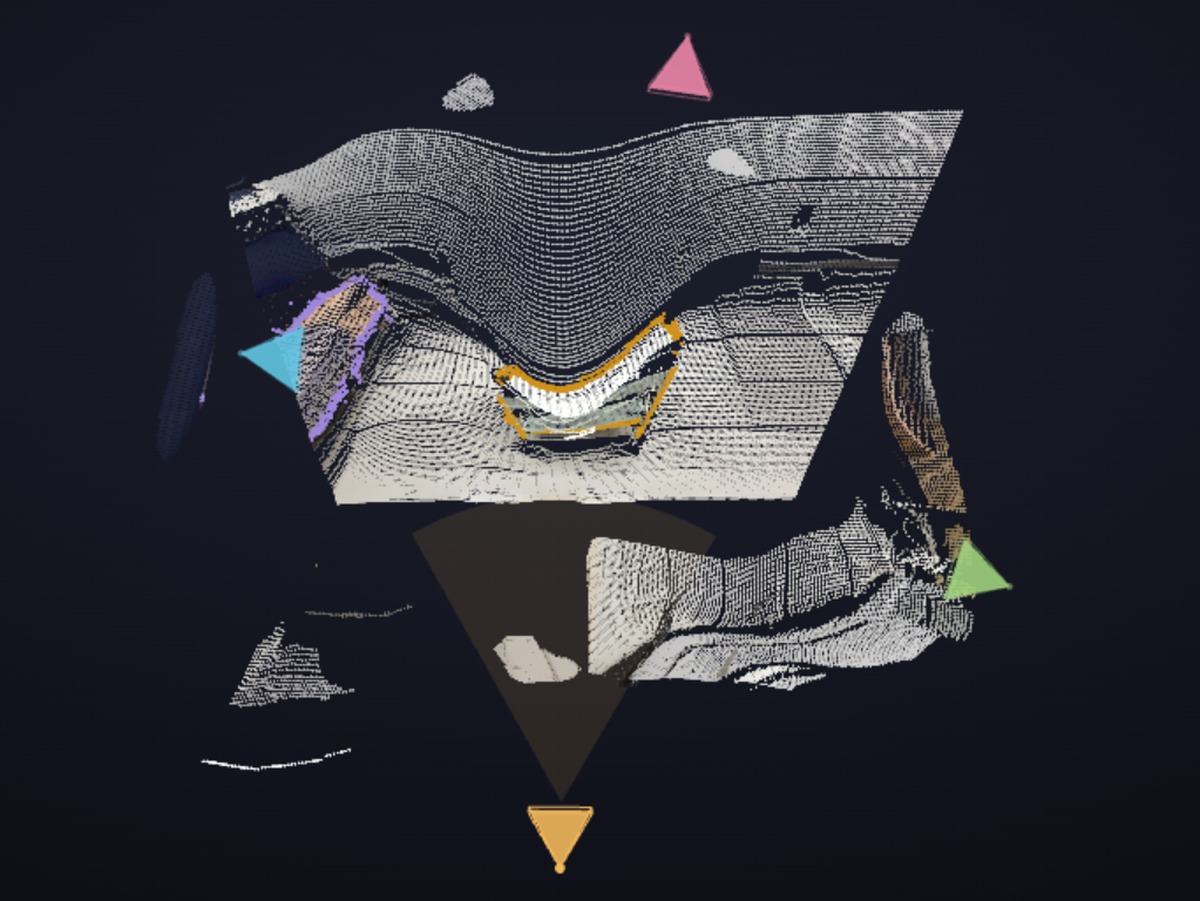}{\caselab{0.256}{0.697}{south}{caseO1}{seating bench}\caselab{0.471}{0.619}{south}{caseO2}{blue table}\caselab{0.319}{0.231}{south}{caseO3}{air conditioner}}
\casenote{headings $+86^\circ$\,\ding{51}, $+180^\circ$\,\ding{51}, $-78^\circ$\,\ding{51}}
\end{minipage}\hfill\begin{minipage}[t]{0.325\linewidth}\vspace{0pt}\centering
\casecap{$\mathbf{z}_1$: from image 4's viewpoint}
\casepanel{\linewidth}{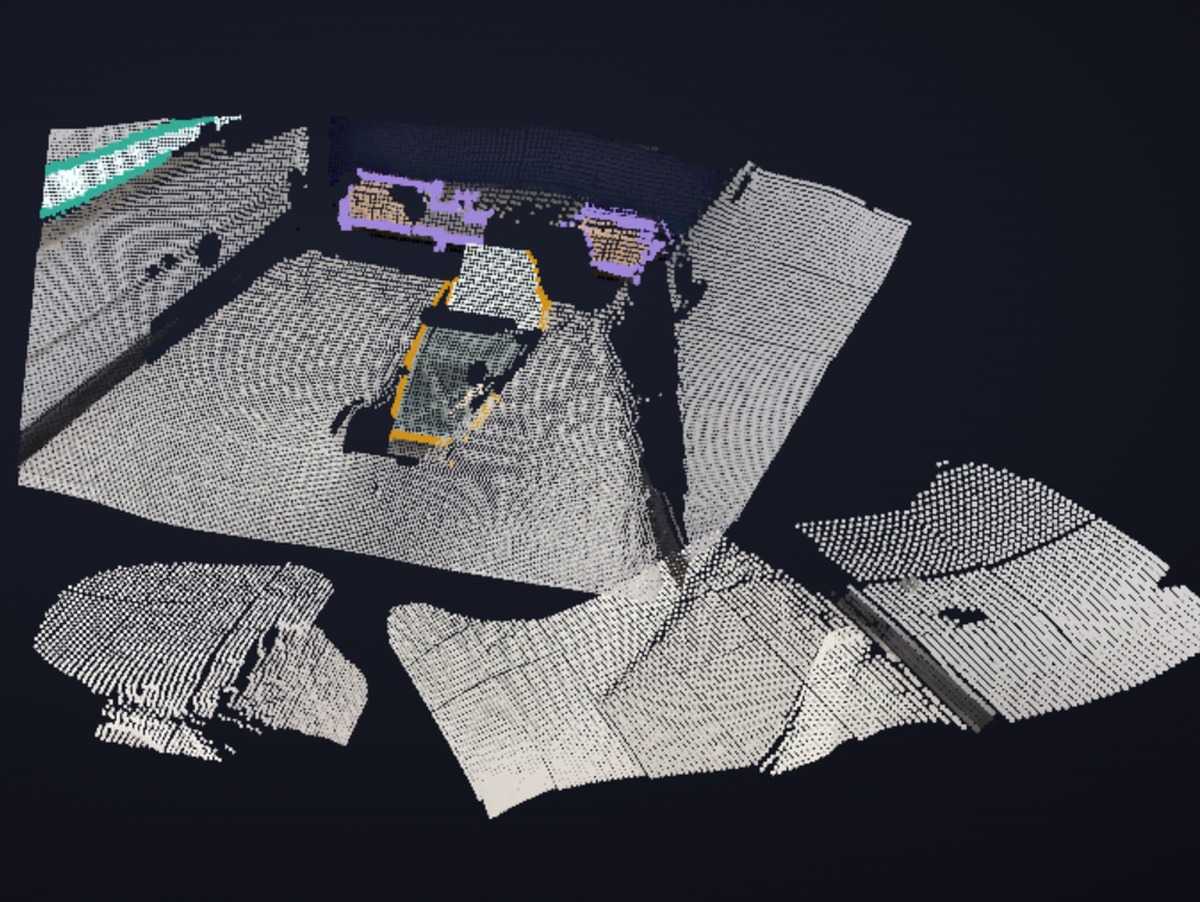}{\caselab{0.387}{0.792}{south}{caseO1}{seating bench}\caselab{0.365}{0.544}{north}{caseO2}{blue table}\caselab{0.192}{0.884}{south west}{caseO3}{air conditioner}}
\casenote{match error 0.18 (others $\geq$0.68)}
\end{minipage}\hfill\begin{minipage}[t]{0.325\linewidth}\vspace{0pt}\centering
\casecap{$\mathbf{z}_2$: from image 3's viewpoint}
\casepanel{\linewidth}{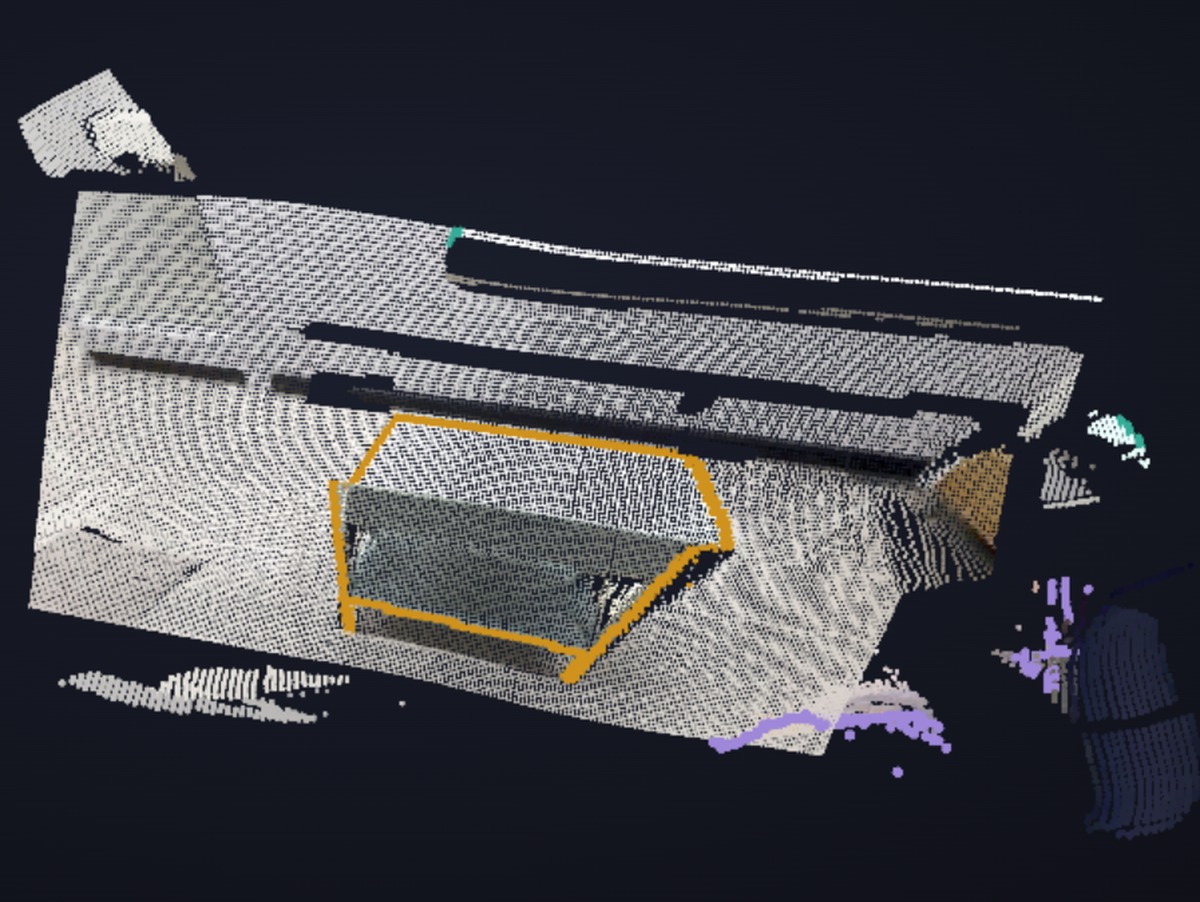}{\caselab{0.683}{0.154}{north}{caseO1}{seating bench}\caselab{0.426}{0.506}{south}{caseO2}{blue table}\caselab{0.630}{0.742}{south}{caseO3}{air conditioner}}
\casenote{match error 0.20 (others $\geq$0.75)}
\end{minipage}\par\medskip
\begin{minipage}[t]{0.325\linewidth}\vspace{0pt}\centering
\casecap{Qwen3.5-4B: $\mathbf{z}_0$ from the probe}
\casepanel{\linewidth}{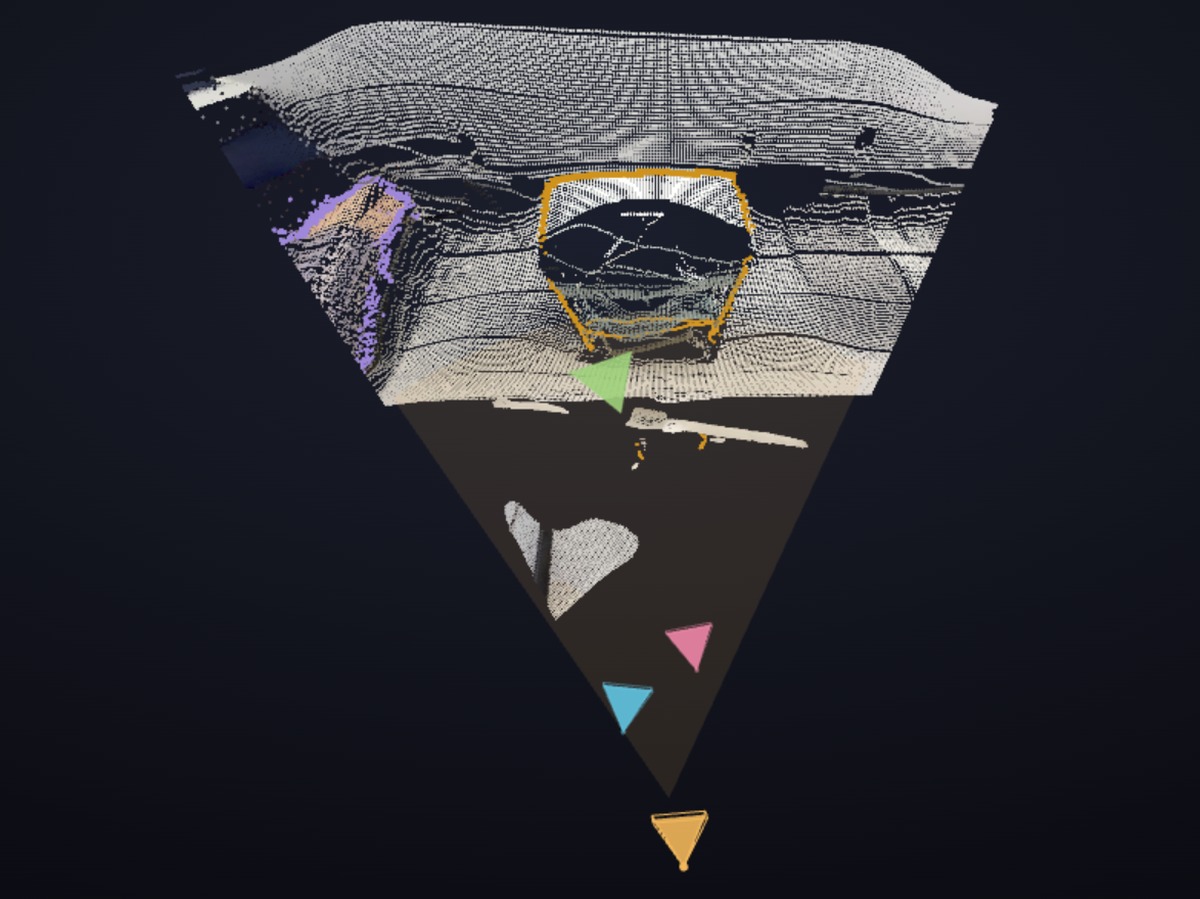}{\caselab{0.267}{0.680}{north}{caseO1}{seating bench}\caselab{0.518}{0.804}{south}{caseO2}{blue table}}
\casenote{headings $+9^\circ$\,\ding{55}, $-45^\circ$\,\ding{55}, $-72^\circ$\,\ding{51}}
\end{minipage}\hfill\begin{minipage}[t]{0.325\linewidth}\vspace{0pt}\centering
\casecap{+\,SFT: $\mathbf{z}_0$ from the probe}
\casepanel{\linewidth}{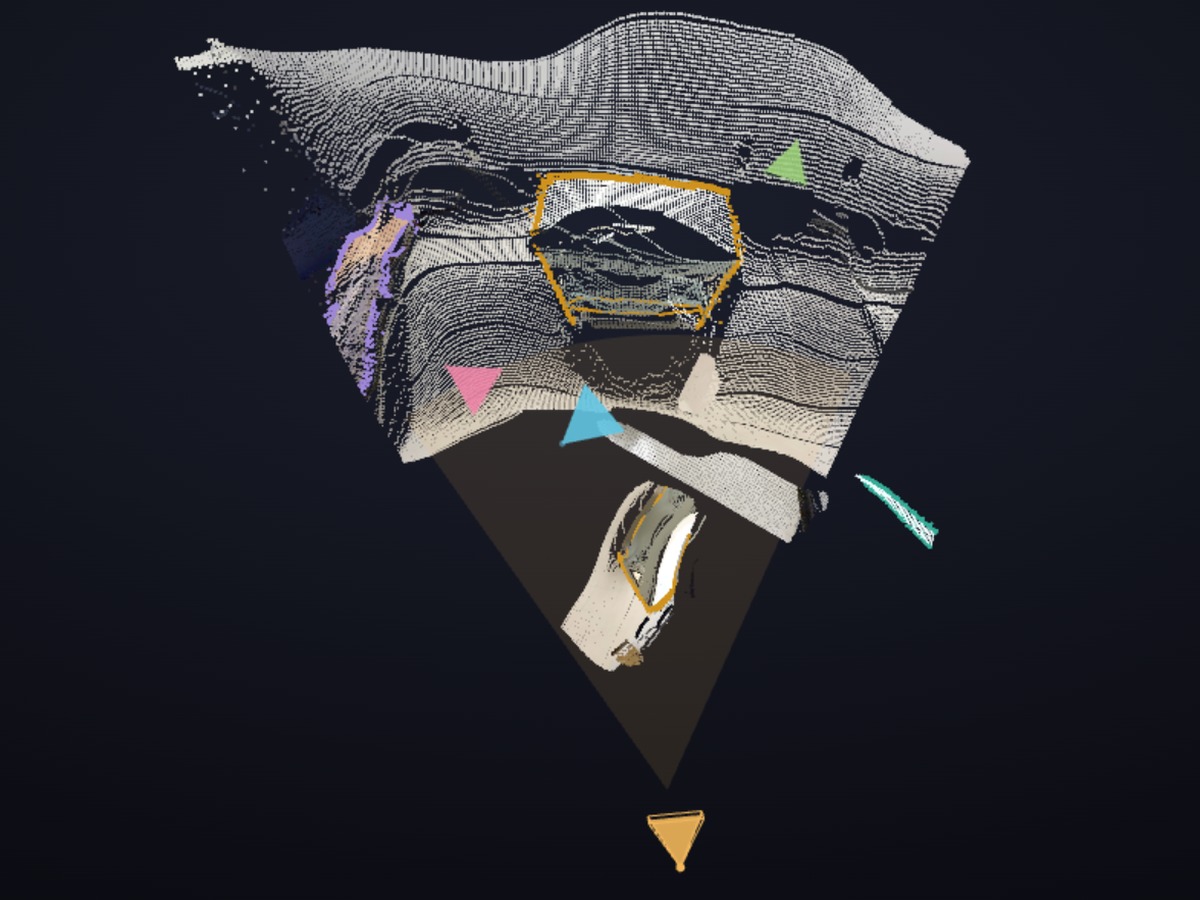}{\caselab{0.290}{0.749}{south}{caseO1}{seating bench}\caselab{0.554}{0.346}{north}{caseO2}{blue table}\caselab{0.724}{0.483}{south}{caseO3}{air conditioner}}
\casenote{headings $+59^\circ$\,\ding{51}, $-89^\circ$\,\ding{55}, $-81^\circ$\,\ding{51}}
\end{minipage}\hfill\begin{minipage}[t]{0.325\linewidth}\vspace{0pt}\centering
\casecap{+\,GRPO: $\mathbf{z}_0$ from the probe}
\casepanel{\linewidth}{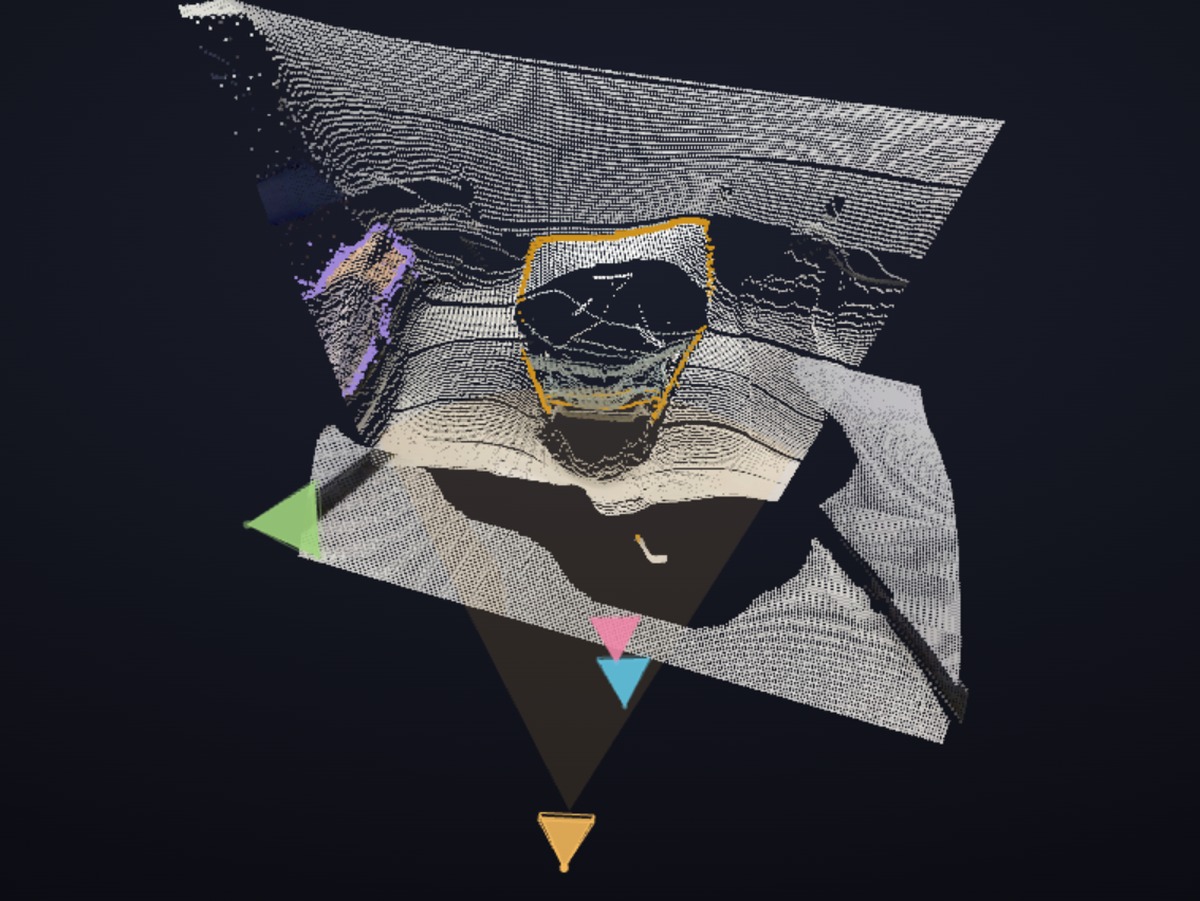}{\caselab{0.275}{0.765}{south}{caseO1}{seating bench}\caselab{0.492}{0.536}{north}{caseO2}{blue table}}
\casenote{headings $+36^\circ$\,\ding{55}, $+26^\circ$\,\ding{55}, $-4^\circ$\,\ding{55}}
\end{minipage}
\caption{\textbf{Four views around a blue table.} The question asks what lies to the right of the blue table as seen from image 3. Top: \method{}'s $\mathbf{z}_0$ and its scenes at the runs it re-emits, decoded by its frozen readout; bottom: the $\mathbf{z}_0$ that the geometry probe reads from each baseline. Each $\mathbf{z}_0$ is seen from straight above, each camera as a wedge in its photo's color opening the way it faces and image 1's field of view shaded; each $\mathbf{z}_k$ from behind its viewer.}
\label{fig:case-bench}
\end{figure*}

\Cref{fig:case-coffee,fig:case-bench} show two questions of MindCube-Tiny from two different rooms that \method{} answers correctly while Qwen3.5-4B, SFT, and GRPO all answer wrongly; each fuses four views around a table and asks from another image's viewpoint.
The scenes of \method{} are decoded by its frozen readout from the \tokens{} of its own reasoning, $\mathbf{z}_0$ after the images and $\mathbf{z}_k$ at the run after its $k$-th viewpoint phrase; the baselines have no \tokens{}, so their $\mathbf{z}_0$ is read by the geometry probe of \cref{sec:diagnosis} from probe tokens placed after their images.
Each scene is drawn as on the project page, in dots: every $\mathbf{z}_0$ from straight above, with image 1's camera at the bottom facing up, its field of view shaded, and each camera as a wedge in its photo's color that opens the way the camera faces; every $\mathbf{z}_k$ from behind and above its viewer. Where the photos overlap, each surface is drawn from one photo only, the viewer's first, so that the copies of an object decoded from different photos do not hide one another.
Three objects per question are outlined in the same color in the photos and in every scene, with masks from SAM~2~\citep{sam2} prompted with boxes we drew; each name stands beside the decoded points of the object's patches.
Under each scene is what we measure on it: the headings at which $\mathbf{z}_0$ places the later cameras (\ding{51}/\ding{55}: within $45^\circ$ of what the question's setup implies or not), and for each $\mathbf{z}_k$ the median distance of its points from $\mathbf{z}_0$ re-expressed at the declared viewpoint, against $\mathbf{z}_0$ re-expressed at any other camera.

In both cases, every run that \method{} re-emits holds the scene re-expressed at its declared viewpoint, with the named objects around the viewer where the photos show them.
The $\mathbf{z}_0$ of \method{} places every camera as the setup implies, whereas the probe places at least one camera of every baseline wrongly; the baselines' scenes pile the photos up in front of image 1, within its field of view, where objects from opposite sides of the room fall together.

\makeatletter\setlength{\@fptop}{0pt}\makeatother

\end{document}